\documentclass{article}

\usepackage{arxiv}

\usepackage[utf8]{inputenc} 
\usepackage[T1]{fontenc}    
\usepackage{hyperref}       
\usepackage{url}            
\usepackage{booktabs}       
\usepackage{amsfonts}       
\usepackage{nicefrac}       
\usepackage{microtype}      
\usepackage{lipsum}
\usepackage{graphicx}
\usepackage{hyperref}
\usepackage{epstopdf, epsfig}
\usepackage{bm}
\usepackage{amsthm}
\usepackage{amsmath}
\usepackage{amssymb}
\usepackage[utf8]{inputenc}
\usepackage[T1]{fontenc}
\usepackage{mathptmx}
\usepackage[numbers]{natbib}
\usepackage{eucal}
\usepackage{color}
\usepackage{subfigure}
\usepackage[ruled,linesnumbered]{algorithm2e}
\usepackage{natbib}
\usepackage{caption}
\SetKwComment{Comment}{$\triangleright$\ }{}

\newcommand{\vect}[1]{\bm{#1}}

\newcommand{\matr}[1]{\mathbf{#1}}

\title{A deep learning based surrogate model for stochastic simulators}

\author{
  Akshay Thakur \\
  Department of Applied Mechanics\\
  Indian Institute of Technology Delhi\\
  Hauz Khas - 110016, New Delhi, India \\
  \texttt{akshaythakur1482@gmail.com} \\
   \And
 Souvik Chakraborty \\
  Department of Applied Mechanics\\
  School of Artificial Intelligence\\
  Indian Institute of Technology Delhi\\
  Hauz Khas - 110016, New Delhi, India \\
  \texttt{souvik@am.iitd.ac.in} \\
}

\begin{document}
\maketitle

\begin{abstract}
We propose a deep learning-based surrogate model for stochastic simulators. The basic idea is to use generative neural network to approximate the stochastic response. The challenge with such a framework resides in designing the network architecture and selecting loss-function suitable for stochastic response. While we utilize a simple feed-forward neural network, we propose to use conditional maximum mean discrepancy (CMMD) as the loss-function. CMMD exploits the property of reproducing kernel Hilbert space and allows capturing discrepancy between the between the target and the neural network predicted distributions. The proposed approach is mathematically rigorous, in the sense that it makes no assumptions about the probability density function of the response. Performance of the proposed approach is illustrated using four benchmark problems selected from the literature. Results obtained indicate the excellent performance of the proposed approach.   
\end{abstract}


\section{Introduction}\label{S:1}
Computational models, numerical simulations or more simply, simulators, regardless of the different names attached, have been pervasively used in numerous fields, including nearly every applied and fundamental mathematical and physical science, to study phenomena ranging from simple to the most complex. Simulators can be deterministic or stochastic. For a deterministic simulator, the value of the output stays the same every time the model is evaluated for a fixed set of input parameters. On the other hand, stochastic simulators, due to associated stochasticity induced due to incomplete knowledge or some source of randomness intrinsic to the system, for fixed set of input parameters produce an output which is a random variable following an unknown conditional distributions. In other words, every time the model is executed the output is different. Stochastic simulators find wide-spread usage in complex system modeling in many fields such as epidemiology \cite{BRITTON201024, allen2017primer}, biology \cite{saarinen2008stochastic, kloeden1992stochastic}, finance \cite{shreve2004stochastic}, and radio-astronomy \cite{kloeden1992stochastic}, to name a few. 


It is computationally expensive to conduct analysis of stochastic systems, particularly when the most straightforward and traditional methods for analysis like Monte Carlo (MC) methods are used, and the situation does not improve even with the usage of several improved procedures such as the Quasi-MC method \cite{caflisch1998monte}, the Latin Hypercube Sampling \cite{loh1996latin}, and the Markov Chain-MC method \cite{brooks1998markov}. The high computational cost arises from the need to repeatedly evaluate the model with the same input parameters in order to obtain good statistics for successful and complete characterization of the response distribution. The repeated model evaluations are also known as replications and different sets of input values for which these replications are performed are collectively called the experimental design. In addition, if the single run of a stochastic model is already expensive, obtaining a large number of realizations becomes an even more time-consuming and difficult task. These problems become quite evident and aggravated when conducting statistical studies such as uncertainty quantification, sensitivity analysis, and optimization because of the requirement of large number of replications at all points of the experimental design. Therefore, it is much more sensible and convenient to replace the actual stochastic model with surrogate models, which mimic the input-output mapping and are computationally inexpensive.

It is not possible to directly use standard surrogate modeling methods for emulation of stochastic models because of associated randomness with the latter. This inability makes surrogate creation for stochastic models significantly more challenging. The field of development of surrogate models for stochastic simulators is still in its younger stages; however, in the recent years it has seen some advancement. Broadly, the non-intrusive methods for stochastic model emulation can be classified into three categories - the \textit{random field approach}, the \textit{statistical approach}, and the \textit{replication based approach}.
The  \textit{random field approach} uses random fields as input parameters and the stochastic simulator is approximated by using Karhunen-Loève expansion and Polynomial Chaos (PC) expansions \cite{blatman2011adaptive} or some other surrogate modeling technique \cite{chakraborty2017polynomial}. However, there is a need for fixing a random seed inside the simulator, and thus, this approach works only if the generation of data is done in a particular manner.

The second category of methods, \textit{replication based approach}, is a two step approach. In the first step, replications are performed over the experimental design and parameters for a general parametric distribution are estimated so that it approximates the distribution of the stochastic model's response. In the second step, the obtained parameters are then emulated  by using standard surrogate modeling techniques such as PC expansions \cite{blatman2011adaptive} and Gaussian Processes \cite{nayek2019gaussian,browne2016stochastic,chakraborty2017moment}. Zhu and Sudret \cite{zhu2020replication} found that the quality of created surrogate model depends upon the number of replication and the accuracy with which the response distribution is approximated in the first step. To circumvent this, a joint modeling method was proposed which adds an optimization step to the existing approach. The resulting surrogate model was found to be accurate and requiring  relatively less replications. However, the extra optimization makes the approach computationally expensive. Also, in general, for the parametric replication based approach, the shape of the output distribution is restricted due to prior hypothesis. Although this restriction is removed by non-parametric replication based distribution estimators \cite{moutoussamy2015emulators}, the potential for applicability of these estimators is reduced by the requirement of around $10^4$ replications for each point of the experimental design to build a surrogate model.

Finally, in the third category of methods, \textit{statistical approach}, there is no requirement for random seed fixation or replications. The generalized additive models \cite{hastie2017generalized} and  the generalized linear models \cite{mccullagh2019generalized} can be used for successful and efficacious estimation of the response distribution belonging to exponential family. In case the response distribution follows some arbitrary shape, it is possible to use non-parametric methods such as projection estimators \cite{efromovich2010dimension} and kernel density estimators \cite{hall2004cross, fan2018local}. But it is simultaneously true that the non-parametric approaches become quickly intractable as the input dimensionality increases \cite{tsybakov2009}. Recently, Zhu and Sudret \cite{zhu2021emulation} introduced a statistical approach where they employed response distribution estimation with GLD together with the a modification of feasible generalized least-squares algorithm and maximum conditional likelihood estimations to eliminate the need for replications, and thus, developed a general approach capable of accurate surrogate model construction for stochastic simulators.

In recent times, the incorporation of deep learning techniques \cite{chakraborty2021transfer,kumar2021grade,kumar2021state} have helped in enhancing the capabilities of modeling complex systems in a variety of applied and fundamental scientific disciplines. Also, the recent years have also seen a large number of studies implement deep learning methods for surrogate modeling of various physical and engineering systems. However, so far, no such implementation have been done for emulation of stochastic systems. In this paper, we
propose a flexible deep learning based non-intrusive framework for construction of surrogate models for stochastic simulators. The proposed framework blends feed forward neural network (FNN) with conditional maximum mean discrepancy (CMMD) loss-function. Compared to existing surrogate models for stochastic simulators, the proposed approach has following advantages:
\begin{itemize}
  \item Unlike random field-based approach, there is no requirement for data to be generated in a specific way.
  \item We note that most of the studies of stochastic surrogate modeling have restrictions on the response distribution shapes that could be emulated, and those which do not have such restrictions require large number of repeated model runs for successful emulation. Our framework places no shape restrictions and thus, is capable of emulating response distributions regardless of their shape.
  \item The propounded approach enjoys the advantages of expressiveness and parametric nature of neural networks in combination with the benefits of the non-parametric nature of the employed kernel.
\end{itemize}

The remainder of the paper is organised as follows. Section \ref{S:2} provide details on the problem statement. Section \ref{S:3} briefly covers the preliminary knowledge required to understand the mathematical underpinnings of the proposed frame wok and involves a brief review of the CGMMNs and generative moment matching networks (GMMN). Section \ref{S:4} provides details on the proposed framework. In Section \ref{S:5}, four examples are presented to demonstrate the performance of the proposed framework. Finally, the concluding remarks are provided in Section \ref{S:6}.

\section{Problem statement}\label{S:2}
 A classical simulator $\mathcal{W}_{d}(\bm{x})$, which most often is a deterministic computational model, maps the given set of input parameters  $\vect{x}=\left(x_{1}, x_{2}, \ldots, x_{N}\right) \in \mathcal{M}_{\vect{X}} \subset \mathbb{R}^{N}$ to a response $y \in \mathbb{R}$, which means that if same input parameters are supplied the resulting output from the model will also be the same. However, for stochastic simulators, uncertainty, which arises from randomness intrinsic to the system or from incomplete knowledge, is always present. This uncertainty results in different response $y$ even when the input parameters $\vect{x}$ are constant. One way to model this uncertainty is by introducing a set of random variables $\vect{z}(\omega)$, which on being grouped with $\vect{x}$, help in characterising the stochastic response as a joint PDF $f_{Y(\vect{x},\vect{z}(\omega))}$.
Based on these preliminaries, we consider a stochastic simulator $\mathcal{W}_{s}$, which, conventionally, can be expressed as
\begin{align}
\mathcal{W}_{s}: \mathcal{M}_{\vect{X}} \times \Omega &\mapsto \mathbb{R}\nonumber \\ 
(\vect{x}, \omega) &\mapsto \mathcal{W}_{s}(\vect{x}, \omega),
\label{equation1}
\end{align}

\noindent where the input vector is denoted by $\vect{x}\in\mathcal{M}_{\vect{X}}$, $\mathcal{M}_{\vect{X}}$ is the input space, and the event space of the probability space $\{\Omega,\Sigma, \mathbb{P}\}$, which also manifests the intrinsic randomness.

Stochastic simulators are expensive when compared to deterministic simulations, as one has to perform considerable repeated model run for a fixed set of the input parameters to identify the response distribution and to accomplish tasks such as uncertainty quantification and optimization. The aim of this paper is to develop a generative deep learning-based framework for construction of surrogate models for stochastic simulators which addresses some of limitations of the previously developed approaches and provides some additional benefits.
    
\section{Preliminary}\label{S:3}
\subsection{Reproducing kernel Hilbert Space and kernel embedding}\label{S:11}
We start by a brief overview of embedding in reproducing kernel Hilbert space (RKHS). If there is an RKHS $\mathcal{K}$ of real valued functions on $\Omega$, $f :\Omega\rightarrow\mathbb{R}$, and its inner product is denoted by $\langle\cdot,\cdot\rangle_{\mathcal{K}}$, then there exists a reproducing kernel $k\in\mathcal{K}$ such that the inner product satisfies the reproducing property for every $f\in\mathcal{K}$
\begin{equation}
f(x)=\langle f, k(\cdot, \vect{x})\rangle_{\mathcal{K}}=\sum \alpha_{i} k\left(x, x_{i}\right).
\label{equa1}
\end{equation}
This is also referred as the `kernel trick'. The kernel trick allows to create an infinite dimensional feature map of $\vect{x}$. It is also key to note that there exists a unique RKHS $\mathcal{K}$ for each positive definite kernel such that Eq. \eqref{equa1} satisfies every function in $\mathcal{K}$. Also, embedding can be performed  by taking the expectation of a distribution on its feature map as given by Eq.  \eqref{equa2}
\begin{equation}
\mu_{\vect{X}}=\mathbb{E}_{\vect{X}}[\phi(\vect{X})]=\int_{\Omega} \phi(\vect{X}) d P(\vect{X}),
\label{equa2}
\end{equation}
where $\mathbb{E}$ is the expectation operator. Furthermore, if $\mathbb{E}_{X}[k(X, X)] \leq \infty$, it can be guaranteed that $\mu_{X}$ is an element of RKHS.

\subsection{Conditional Distributions and kernel embedding}\label{S:13}
For a conditional distribution $P(Y \mid \vect{X})$, the kernel embedding is given by the following equation
\begin{equation}
\mu_{Y \mid \vect{x}} =\mathbb{E}_{Y \mid \vect{x}}[\phi(Y)]= \int_{\Omega} \phi(y) d P(y \mid \vect{x}).
\label{equa7}
\end{equation}
The operator $C_{Y \mid \vect{X}}$ is conventionally used to represent a conditional distribution's embedding. Also, it should be noted that the embeddings of  a conditional distribution are not single elements in the RKHS, rather a family of points, one point for every fixed value of $\vect{X}$, is swept out in the RKHS.  A single RKHS element, $\mu_{Y \mid \vect{x}} \in \mathcal{G}$, can only be obtained by fixing $\vect{X = x}$, where $\vect{x}$ is a particular value in $\vect{X}$ . This means that $C_{Y \mid \vect{X}}$ can be seen as an operator mapping from $\mathcal{K}$ to $\mathcal{H}$, but for the definition to be complete the operator also has to satisfy the following properties
\begin{equation}
   \mu_{Y \mid \vect{x}}=C_{Y \mid \vect{X}} \phi(\vect{x}),\;\;
 \mathbb{E}_{Y \mid \vect{x}}[g(Y) \mid \vect{x}]=\left\langle g, \mu_{Y \mid \vect{x}}\right\rangle_{\mathcal{H}},
\label{equa8}  
\end{equation}
where  $g$ is an element of RKHS $\mathcal{H}$. Further, \citet{song2009hilbert}  used the cross-covariance operator $(C_{\vect{X} Y}: \mathcal{H} \rightarrow \mathcal{K})$ , which is a generalization of the covariance matrix, given by
\begin{equation}
  C_{X Y}=\mathbb{E}_{\vect{X} Y}[\phi(\vect{X}) \otimes \phi(Y)]-\mu_{\vect{X}} \otimes \mu_{Y},
\label{equa9}  
\end{equation}
where $\otimes$ denotes the tensor product, to show that if both the properties given in Eq.  \eqref{equa8} are satisfied and $\mathbb{E}_{Y \mid \vect{X}}[g(Y) \mid \vect{X}] \in \mathcal{K}$, the operator $C_{Y \mid \vect{X}}$ exists and is given by
\begin{equation}
  C_{Y \mid \vect{X}} =\mathcal{C}_{Y \vect{X}} \mathcal{C}_{\vect{X} \vect{X}}^{-1}.
\label{equa10}  
\end{equation}
Given a training dataset containing $N$ pairs of examples sampled from $P(\vect{X}, Y)$, $\mathcal{D}_{\vect{X} Y}=\left\{\left(\vect{x}_{i}, \boldsymbol{y}_{i}\right)\right\}_{i=1}^{N}$, the conditional embedding operator can be estimated with the following equation
\begin{equation}
 \widehat{C}_{Y \mid \vect{X}}=\Phi(\matr{K}+\lambda \matr{I})^{-1} \boldsymbol{\Upsilon}^{\top},
\label{equa11}  
\end{equation}
 where  $\lambda$ acts as a method of regularization, $\boldsymbol{\Phi}=\left(\phi\left(y_{1}\right), \ldots, \phi\left(y_{N}\right)\right)$, and  $\boldsymbol{\Upsilon}=(\phi\left(\vect{x}_{1}\right), \ldots, \phi\left(\vect{x}_{N}\right))$, $\matr{K}=\boldsymbol{\Upsilon}^{\top} \boldsymbol{\Upsilon}$. Also, $\widehat{C}_{Y \mid \vect{X}}$ asymptotically satisfies the both the properties in Eq.  \eqref{equa8} and is also an element in the $\mathcal{K} \otimes \mathcal{H}$ tensor product space.
\subsection{Maximum mean discrepancy}\label{S:12}
Assume we have two sets of samples $\vect{X} = \bigl\{x_{i}\bigr\}_{i=1}^N$ and $\vect{Y} = \bigl\{y_{j}\bigr\}_{j=1}^M$, which belong to probability distributions $P_{\vect{X}}$ and $P_{\vect{Y}}$, respectively. Now, if we were to answer the question whether the generating distribution $P_{\vect{X}} = P_{\vect{Y}}$, then the frequentist estimation would be done through Maximum Mean Discrepancy \cite{gretton2012kernel} (also known as kernel two sample test). The underlying idea is that, if, on comparison, the statistics of the two given sets of samples turn out to be the same, then the generating distributions could be considered to be identical. The following equation (Eq.  \eqref{equa1}) describes the MMD measure for computing the difference of statistics
\begin{equation}
\operatorname{MMD}[\mathcal{F},P_{\vect{X}},P_{\vect{Y}}] = \sup_{f\, \in\,\mathcal{F}}\left(\mathbb{E}_{\vect{X}}\left[f(\vect{X})\right] - \mathbb{E}_{\vect{Y}}\left[f(\vect{Y})\right]  \right),
\label{equa3}
\end{equation}
where $\mathcal{F}$ is a class of functions. It was shown by \cite{gretton2012kernel} that if $\mathcal{F}$ is specified to be an RKHS $\mathcal{K}$, then Eq.  \eqref{equa3} can be solved in closed form, i.e., the two samples can be distinguished, and MMD objective function could be represented with following equation
\begin{equation}
\operatorname{MMD}(\mathcal{F}, p, q)=\left\|\mu_{p}-\mu_{q}\right\|_{\mathcal{K}}^{2},
\label{equa4}
\end{equation}
where $\mu_{p},\mu_{q} \in \mathcal{K}$. However, in practice, the MMD objective in Eq.  \eqref{equa5}, which computes the squared difference between the kernel mean embeddings, is used
\begin{equation}
\widehat{\mathcal{L}}_{\mathrm{MMD}}^{2}=\left\|\frac{1}{N} \sum_{i=1}^{N} \phi\left(x_{i}\right)-\frac{1}{M} \sum_{j=1}^{M} \phi\left(y_{i}\right)\right\|_{\mathcal{K}}^{2},
\label{equa5}
\end{equation}
where, choosing $\phi$ to be identity function results in matching of sample mean. Also, higher order moments can also be matched by selection of some suitable $\phi$. Further, according to \cite{gretton2012kernel}, 
kernel trick could be applied to Eq.  \eqref{equa5} to obtain the following equation
\begin{equation}
\widehat{\mathcal{L}}_{\mathrm{MMD}}^{2}=E\left[k\left(\vect{X}, \vect{X}^{\prime}\right)-2 k(\vect{X}, \vect{Y})+k\left(\vect{Y}, \vect{Y}^{\prime}\right)\right]
\label{equa6}
\end{equation}
For further details on successful implementation of the GMMNs with MMD, the reader is directed to \cite{li2015generative,dziugaite2015training}.

 \subsection{Conditional maximum  mean discrepancy}\label{S:14}
 
 Although we might require a substantial amount of training data, it is still possible to check if two conditional distributions $P_{Y \mid \vect{X}}$ and $P_{Z \mid \vect{X}}$ are equal, for every $\vect{X=x}$ in a finite domain of $\vect{X}$ by using MMD separately for each $X$. But, it no longer remains possible to do so if $\vect{X}$ is continuous. However, \cite{ren2016conditional} proposed a criterion, conditional maximum  mean discrepancy, which circumvents these issues.
 
 By virtue of properties in Eqs.  \eqref{equa8} and \eqref{equa4}, \citet{ren2016conditional} proved that if $\mathcal{K}$ is a universal RKHS, $k(\cdot, \cdot)$ is kernel associated with the RKHS, $\mathbb{E}_{Y \mid \vect{X}}[g(Y) \mid \vect{X}] \in$ $\mathcal{K}$, $ \mathbb{E}_{Z \mid \vect{X}}[g(Z) \mid \vect{X}] \in \mathcal{K}$, and $C_{Y \mid \vect{X}}, C_{Z \mid \vect{X}} \in \mathcal{K} \otimes \mathcal{H}$, then $P_{Y \mid \vect{X}}=P_{Z \mid \vect{X}}$, meaning we have $P_{Y \mid \vect{x}}=P_{Z \mid \vect{x}}$ for every particular value of $\vect{x}$, given the embedding of conditional distributions $C_{Y \mid \vect{X}}=C_{Z \mid \vect{X}}$.
 One can also defined the following objective for CMMD involving Hilbert-Schmidt norm between conditional embedding operators to measure the difference between two conditional distributions \cite{ren2016conditional}
 \begin{equation}
\mathcal{L}_{\mathrm{CMMD}}^{2}=\left\|C_{Y \mid \vect{X}}-C_{Z \mid \vect{X}}\right\|_{\mathcal{K} \otimes \mathcal{H}}^{2}.
\label{equa12}  
\end{equation}
For two given sample training datasets $\mathcal{D}^{s}=\left\{\left(\vect{x}_{i}, y_{i}\right)\right\}_{i=1}^{N}$ and $\mathcal{D}^{d}=\left\{\left(\vect{x}_{i}, y_{i}\right)\right\}_{i=1}^{M}$, if we replace the difference measure by empirical estimates of conditional embedding operators and define $\widetilde{\matr{K}}=\matr{K}+\lambda \matr{I}$, then the difference between the two conditional distributions can be computed in terms of gram matrices by using kernel trick in the following way
\begin{align}
\widehat{\mathcal{L}}_{\mathrm{CMMD}}^{2} &=\left\|\Phi_{d}\left(\matr{K}_{d}+\lambda \matr{I}\right)^{-1} \boldsymbol{\Upsilon}_{d}^{\top}-\Phi_{s}\left(\matr{K}_{s}+\lambda \matr{I}\right)^{-1} \boldsymbol{\Upsilon}_{s}^{\top}\right\|_{\mathcal{K} \otimes \mathcal{H}}^{2}\nonumber  \\
&=\operatorname{Tr}\left(\matr{K}_{d} \widetilde{\matr{K}}_{d}^{-1} \matr{L}_{d} \widetilde{\matr{K}}_{d}^{-1}\right)+\operatorname{Tr}\left(\matr{K}_{s} \widetilde{\matr{K}}_{s}^{-1} \matr{L}_{s} \widetilde{\matr{K}}_{s}^{-1}\right)-2 \cdot \operatorname{Tr}\left(\matr{K}_{s d} \widetilde{\matr{K}}_{d}^{-1} \matr{L}_{d s} \widetilde{\matr{K}}_{s}^{-1}\right),\label{equa13} 
\end{align}
where the two sample datasets are represented with the subscripts $s$ and $d$ and the implicitly formed feature matrices for dataset $\mathcal{D}^{d}$ are given by $\boldsymbol{\Upsilon}_{d}=\left(\phi\left(\vect{x}_{1}^{d}\right), \ldots, \phi\left(\vect{x}_{N}^{d}\right)\right)$ and $\boldsymbol{\Phi}_{d}=\left(\phi\left(y_{1}^{d}\right), \ldots, \phi\left(y_{N}^{d}\right)\right)$. Also, for dataset $\mathcal{D}^{s}$, $\boldsymbol{\Upsilon}_{s}$ and $\boldsymbol{\Phi}_{s}$ are defined in a similar fashion. Further, the gram matrices for the input variables are defined as $\matr{K}_{d}=\boldsymbol{\Upsilon}_{d}^{\top} \boldsymbol{\Upsilon}_{d}$ and $\matr{K}_{s}=\boldsymbol{\Upsilon}_{s}^{\top} \boldsymbol{\Upsilon}_{s}$ , whereas the gram matrices for output variables are defined as $\matr{L}_{d}=\boldsymbol{\Phi}_{d}^{\top} \boldsymbol{\Phi}_{d}$ and $\matr{L}_{s}=\boldsymbol{\Phi}_{s}^{\top} \boldsymbol{\Phi}_{s}$. Also, the definition of gram matrices between the input and the output variable dataset are given as $\matr{K}_{s d}=\boldsymbol{\Upsilon}_{s}^{\top} \boldsymbol{\Upsilon}_{d}$ and $\matr{L}_{d s}=\boldsymbol{\Phi}_{d}^{\top} \boldsymbol{\Phi}_{s}$.

Both MMD and CMMD can be used as loss-functions in deep learning frameworks. For instance, \cite{li2015generative, dziugaite2015training} used MMD as a loss-function in their study. However, in the context of regression, we are generally interested in computing $p(y \mid \vect{x})$ as compared to $p(\vect{x}, y)$. Notwithstanding the fact that $p(y \mid \vect{x})$ can still be computed using $p(\vect{x}, y)$, it is often very tedious for finite domain of input variables and impossible for a continuous variables. Also, there are much less non-essential assumptions associated with a CMMD-based model $p(y \mid \vect{x})$. Furthermore, a CMMD-based model is able to outperform a MMD-based model and requires a fewer training examples \cite{lafferty2001conditional, ng2002discriminative}. Therefore, CMMD is a better option as far as the development of surrogate models for stochastic simulators is considered.

\textbf{Remarks:} It is to be noted that the existence of Hilbert-Schmidt norm and thus, a well-defined CMMD objective for the condition mean embedding operator$C_{Y \mid \vect{X}} \in$ $\mathcal{K} \otimes \mathcal{H}$ are assumed. However, this assumption might not always be true. Nevertheless, for all practical purposes, the utilized empirical estimator $\Phi(\matr{K}+\lambda \matr{I})^{-1} \boldsymbol{\Upsilon}^{\top}$ provides a valid approximation \cite{song2009hilbert}. 

\section{Proposed framework}\label{S:4}
\begin{figure}
\includegraphics[width=1\textwidth]{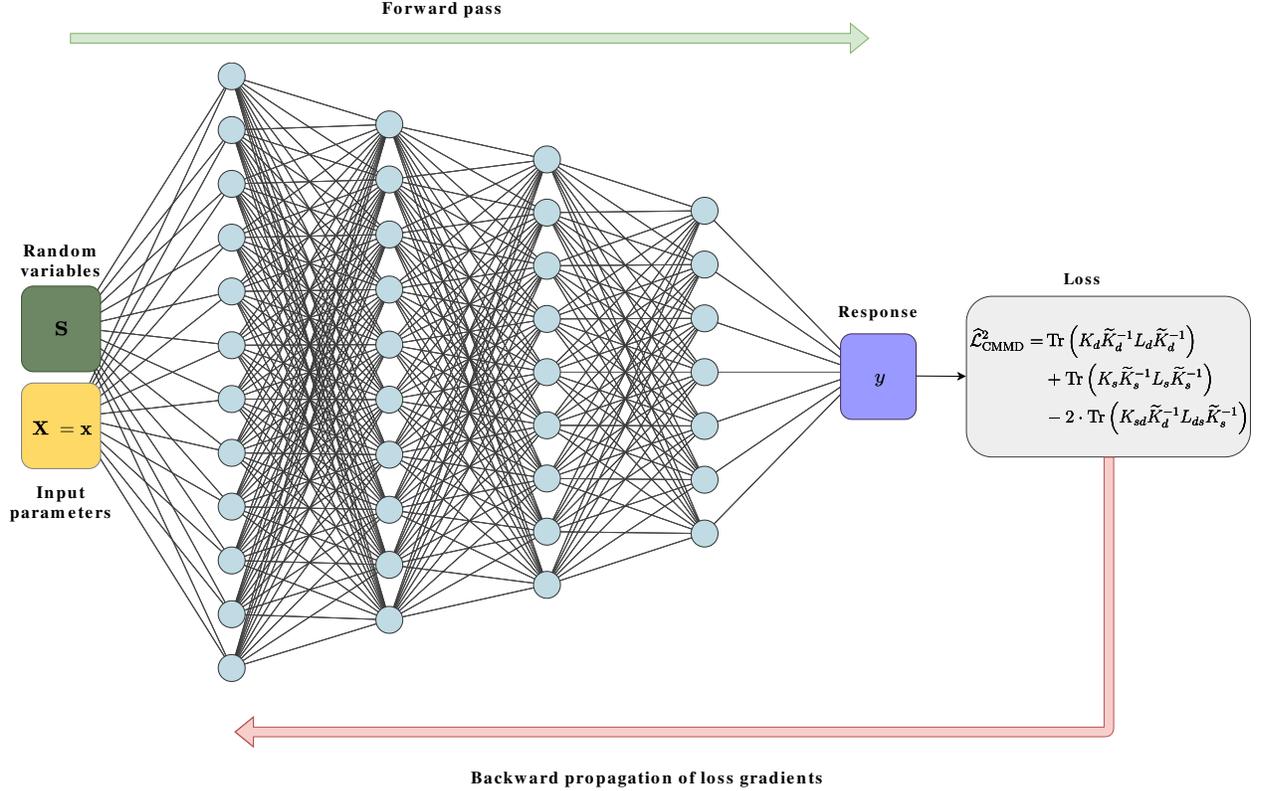}
\caption{Schematic representation of the proposed deep learning framework based on CGMMNs.$\vect{X = x}$ are the set of input parameters which are concatenated with the set of random variable $\vect{S}$ to form the input to the five-layer neural network. The loss function for the network is the CMMD objective and after calculation of the loss, the gradients of the same are back-propagated through the network.}
\label{fig:0}
\end{figure}

Having discussed MMD and CMMD, we proceed towards discussing the proposed deep learning-based surrogate model for stochastic simulators. Considering $\vect{X} \in \mathbb{R}^{N}$ and $y$ to be input and output variables, we are interested in developing a surrogate model that can approximate $P_{Y \mid \vect{X}}$. To that end, we use an FNN having five layers. Given the advantages of CMMD, we propose to use the same as a loss-function. We use rectified linear unit (ReLu) activation function for the hidden layers. As for the last (output) layer, the activation of the output layer is decided on a case to case basis. It should be noted though that any choice could be made for hidden layers' activation function as well, provided it is nonlinear. A schematic representation of the proposed framework is presented in Fig. \ref{fig:0}.   
The functioning of the network is contingent on the fact that for distribution $\mathcal{Q}$, which resides on the sample space $\mathbb{J}$, and any regular enough continuous distribution $\mathcal{T}$ residing on $\mathbb{L}$, there exists a function $K$, which maps $\mathbb{L}$ to $\mathbb{J}$, i.e., $K:\mathbb{L}\mapsto\mathbb{J}$, in such a way that $K(\vect{x}) \sim \mathcal{Q}$ and $\vect{x} \sim \mathcal{T}$. A deep neural network (DNN) makes it possible to find this function $K$. But a successful search would demand that a dependence of $K$ on the input variables is established. Therefore, to establish this dependence, the input of the DNN, as shown in Fig. \ref{fig:0}, is created by concatenating the given set of input parameters, $\vect{x}$, with a set of additional random variables sampled from a continuous distribution denoted by $\vect{S} \in \mathbb{R}^{M}$. For the current study, $\vect{S}$, the set of random variables, is sampled from standard normal distribution $\mathcal{N}(0, 1)$. The concatenated inputs are then passed through the neural network and training is performed by optimizing the square-root of CMMD objective given in Eq.  \eqref{equa13}. The obtained output is the predicted response $y$, which should be a sample from the conditional distribution $P_{Y \mid \vect{X}}$, and can also be represented with the following equation
\begin{equation}
 y=f(\vect{x}, \vect{S} ; \boldsymbol{\theta}),
\label{equa14}
\end{equation}
where $\boldsymbol{\theta}$ is used to denote the network parameters and $f:(\vect{x}, \vect{S}) \mapsto y$ is the mapping function uncovered by the neural network.

\subsection{Kernel and automatic kernel parameter learning}\label{S:41}
It is possible to create an explicit feature map, which consists of infinite number of terms and has the ability to extend the moment matching between two distributions to all orders, by using Taylor expansion \cite{li2015generative}. However, the preceding statement is only possible if the kernel is universal. Therefore, for the current study we choose the universal squared exponential kernel \cite{duvenaud2014automatic}, which is given by the following equation
\begin{equation}
 \matr{k}_{\mathrm{SE}}\left(\matr{x},\matr{x^{\prime}}\right)=\sigma^{2} \exp \left(-\frac{\left(\matr{x}-\matr{x^{\prime}}\right)^{2}}{2 \ell^{2}}\right),
\label{equa15}  
\end{equation}
where the parameters $\ell$  and $\sigma$ determine the spread and maximum value of the kernel, respectively. Further, we automate the learning of these parameters by making them trainable variables in our deep learning framework and including them in the optimization of the CMMD objective via gradient based optimization algorithms involving gradient calculation with back-propagation.  
\subsection{Algorithm}\label{S:42}
\begin{algorithm}[htbp!]
\caption{Mini-batch gradient descent for CGMMN based surrogate modeling of stochastic simulators}\label{alg:one}
\textbf{Input:} Input and response dataset  $\mathcal{D} = \left\{\left(\vect{x}_{i}, y_{i}^{t}\right)\right\}_{i=1}^{N}$ generated from stochastic simulations\\
Initialize network parameters and hyper-parameters $\vect{\theta}, \alpha, \lambda,\ell,\sigma,and\;M$\\
Divide $\mathcal{D}$ into $N_{m}$ mini batches of size $b = N/N_{m}$ randomly\\
 \For{$e= 1$ to  $number\; of\; epochs$}    
        { 
        	Regenerate $\vect{S}$, a set of standard normal variables of size $M$, every $k$ epochs\\
        	 \For{$n_{m} = 1$ to $N_{m}$}    
                { 
                	Draw a mini batch $\mathcal{A}$ from $\mathcal{D}$\\
                	 \For{$l= 1$ to $b$}    
                        { 
        	              $y_{l}\leftarrow f(\vect{x}_{l}, \vect{S}_{l} ; \boldsymbol{\theta})$
                        }
        	        Store all $(\vect{x}_{l},y_{l})$ in dataset $\mathcal{A}^{\prime}$\\
        	        Calculate the gradient $\frac{\partial \hat{\mathcal{L}}_{\mathrm{CMMD}}}{\partial \theta}$ on mini batch datasets $\mathcal{A}$ and $\mathcal{A}^{\prime}$.\\
        	        Update $\boldsymbol{\theta}$ using some gradient-based Adam optimizer
        	        
                }
                
        }
\textbf{Ouput:} Learned parameter $\boldsymbol{\theta}$
\end{algorithm}

\section{Results and discussion}
\label{S:5}
In this section, four benchmark problems are presented to demonstrate the performance of the proposed approach. The pliability of the propounded approach is assessed by careful selection of benchmark problems, such that the response PDFs, especially for the last two problems, exhibit rich set of shapes. In addition, we present case studies by varying the number of points of the experimental design, number of replications, and the number of additional input random variables. Hellinger distance is used as a metric for quantification of error for these case studies. Formally, the Hellinger distance between two continuous PDFs $r$ and $t$ is given by the following equation
\begin{equation}
H(r(y), t(y)) =\frac{1}{\sqrt{2}}\left\|\sqrt{r(y)}-\sqrt{t(y)}\right\|_{2}.
\end{equation}
The accuracy of the proposed approach across the experimental design is evaluated by plotting the statistics of predicted response against the statistics of true or reference response for 4000 sets of input parameter $\vect{X}$ for every case. Furthermore, we also present the statistics of error metric, Hellinger distance, calculated for the 4000 sets of input parameter $\vect{X}$ in  a tabular form for each problem.   
\subsection{Example 1: A one-dimensional simulator}\label{ss:1}
As the first example, we consider the following analytical case put forth by Zhu and Sudret \cite{zhu2020replication}
\begin{equation}
Y(X,\omega) = \sin{\bigg(\frac{2\pi}{3}X + \frac{\pi}{6}\bigg)}\cdot(Z_{1}(\omega)\cdot Z_{2}(\omega))^{\cos{X}}.
\label{eqn1}
\end{equation}
In Eq.  \eqref{eqn1}, $X\sim\mathcal{U}(0,1)$ and the latent variables $Z_{1}(\omega)\sim\mathcal{LN}(0,0.25)$ and $Z_{2}(\omega)\sim\mathcal{LN}(0,0.25)$. This definition leads to  $Y(x,\omega)$ following a lognormal distribution $\mathcal{LN}(\mu(x),\,\sigma(x))$, where $\mu(x) = \log\Big(\sin{\Big(\frac{2\pi}{3}x + \frac{\pi}{6}\Big)}\Big)$ and $\sigma(x) = \sqrt{\frac{3}{8}}cos(x)$ \cite{zhu2020replication}. The objective here is to develop a surrogate model for Eq.  \eqref{eqn1} using the proposed framework. To that end, the vector $(X,\vect{S})^{T}$ is used as an input for the neural network, where $\vect{S}$ is a vector containing $M$ standard normal random variables, i.e., $(s_{1},s_{2},...,s_{M})$. Also, Swish activation function is used for the output layer, the batch size is set to 300, and the number of epochs required for successful training is 300. Throughout this study, we use an investigation method where we sample $N$ times through our input parameters' range to obtain $N$ sets of input parameters, and then perform $R$ number of replications for each input parameter set to generate our complete training dataset. Also, we fix the number of the standard normal input random variables, $N_{z}$, by performing convergence studies.

\begin{figure}[htbp!]
    \centering
    \subfigure[]{\includegraphics[width=0.475\textwidth]{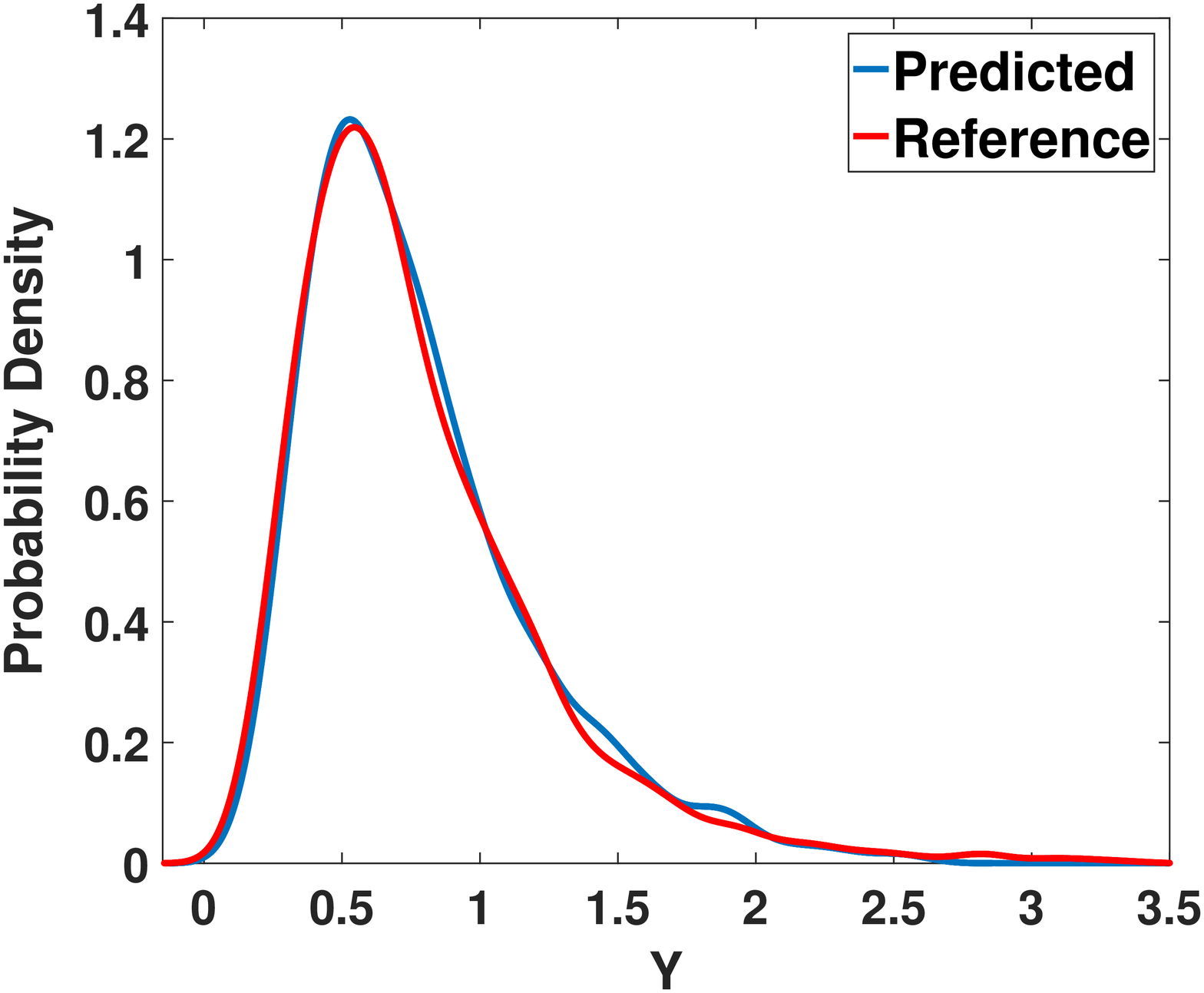}} 
    \subfigure[]{\includegraphics[width=0.475\textwidth]{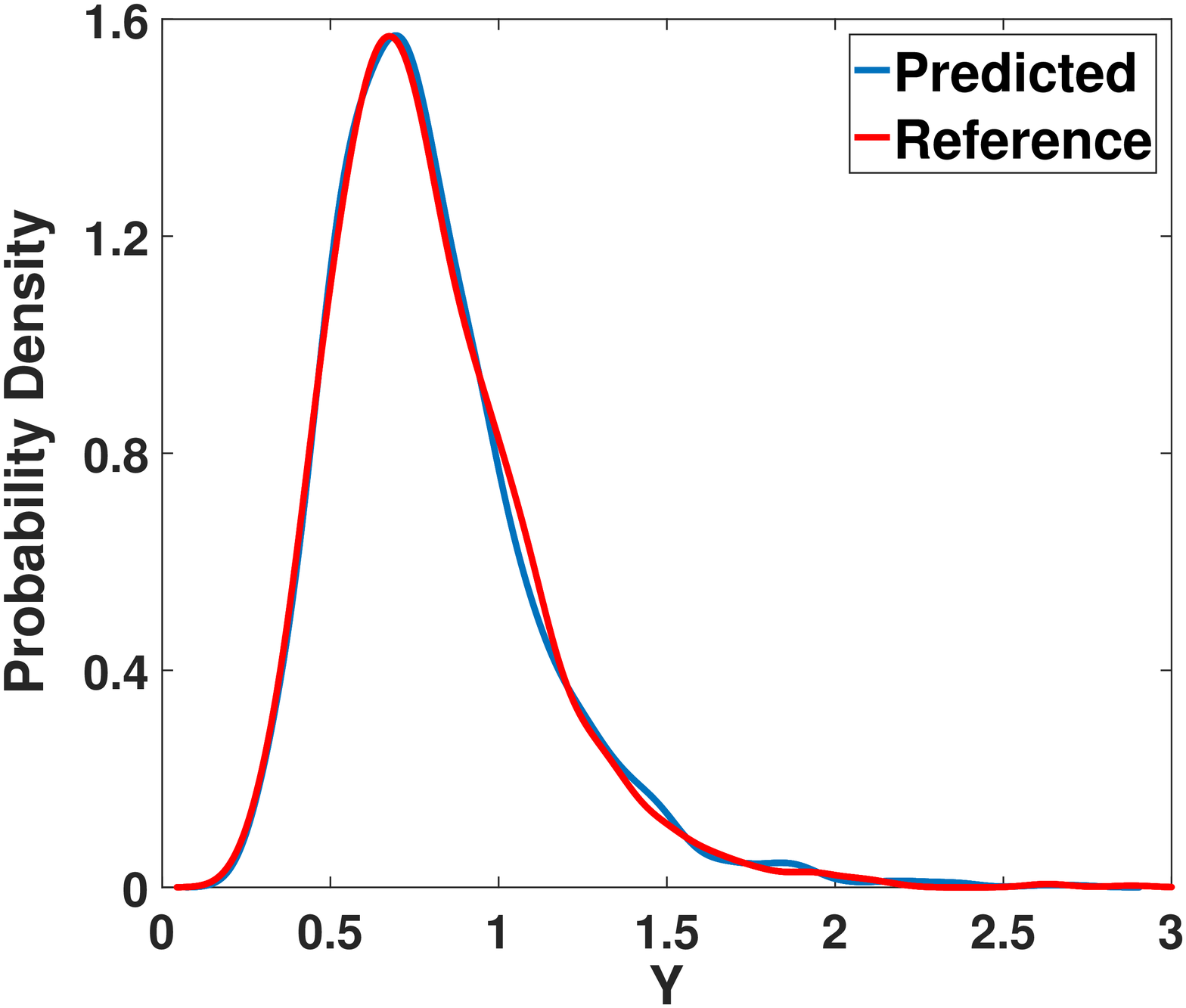}}
    \caption{ \textbf{(a)} PDF prediction for $x = 0.1$   \textbf{(b)} PDF prediction for $x = 0.85$}
    \label{fig:1}
\end{figure}

\begin{figure}[htbp!]
    \centering
    \subfigure[]{\label{subfig:lab11}\includegraphics[width=0.475\textwidth]{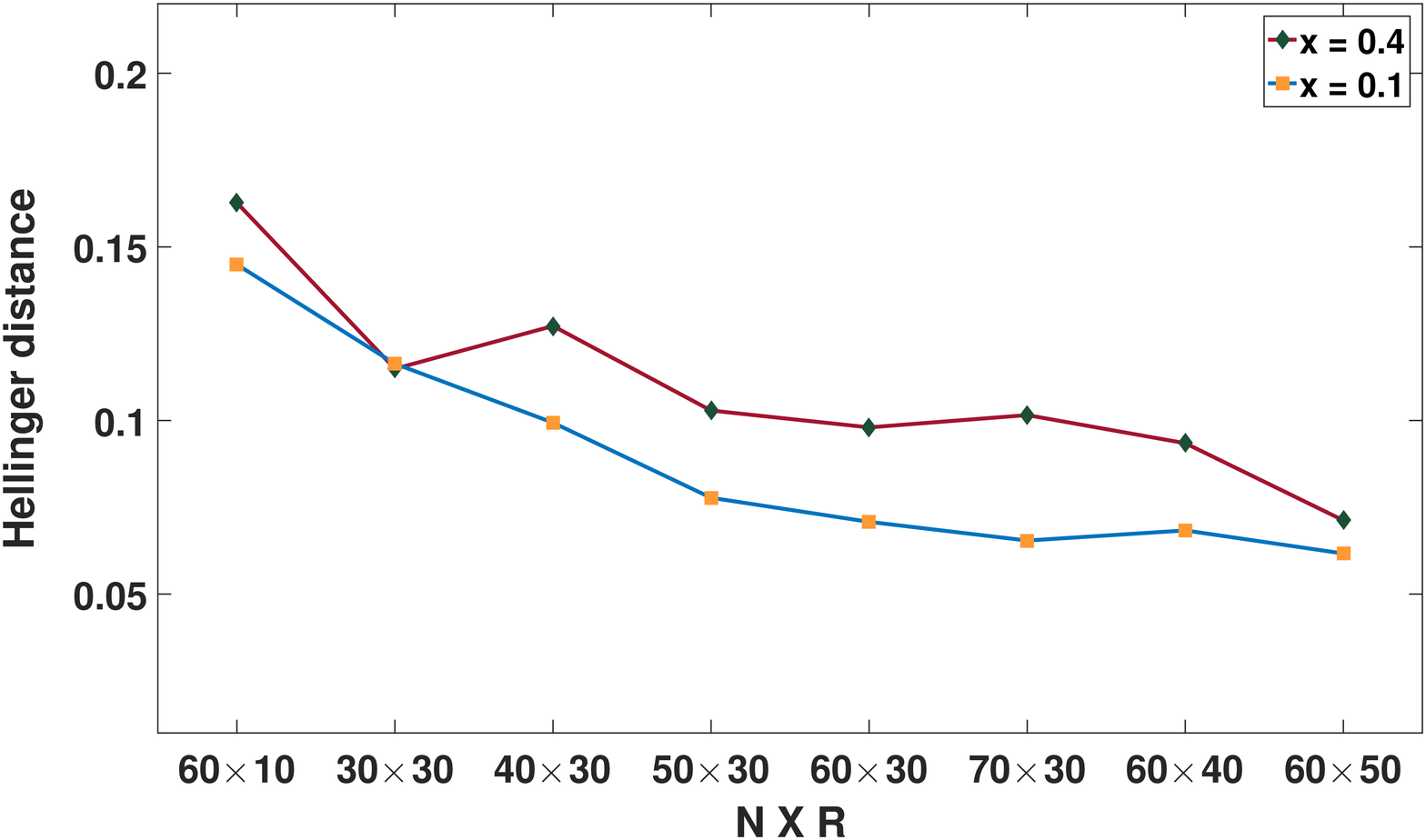}} 
    \subfigure[]{\label{subfig:lab12}\includegraphics[width=0.475\textwidth]{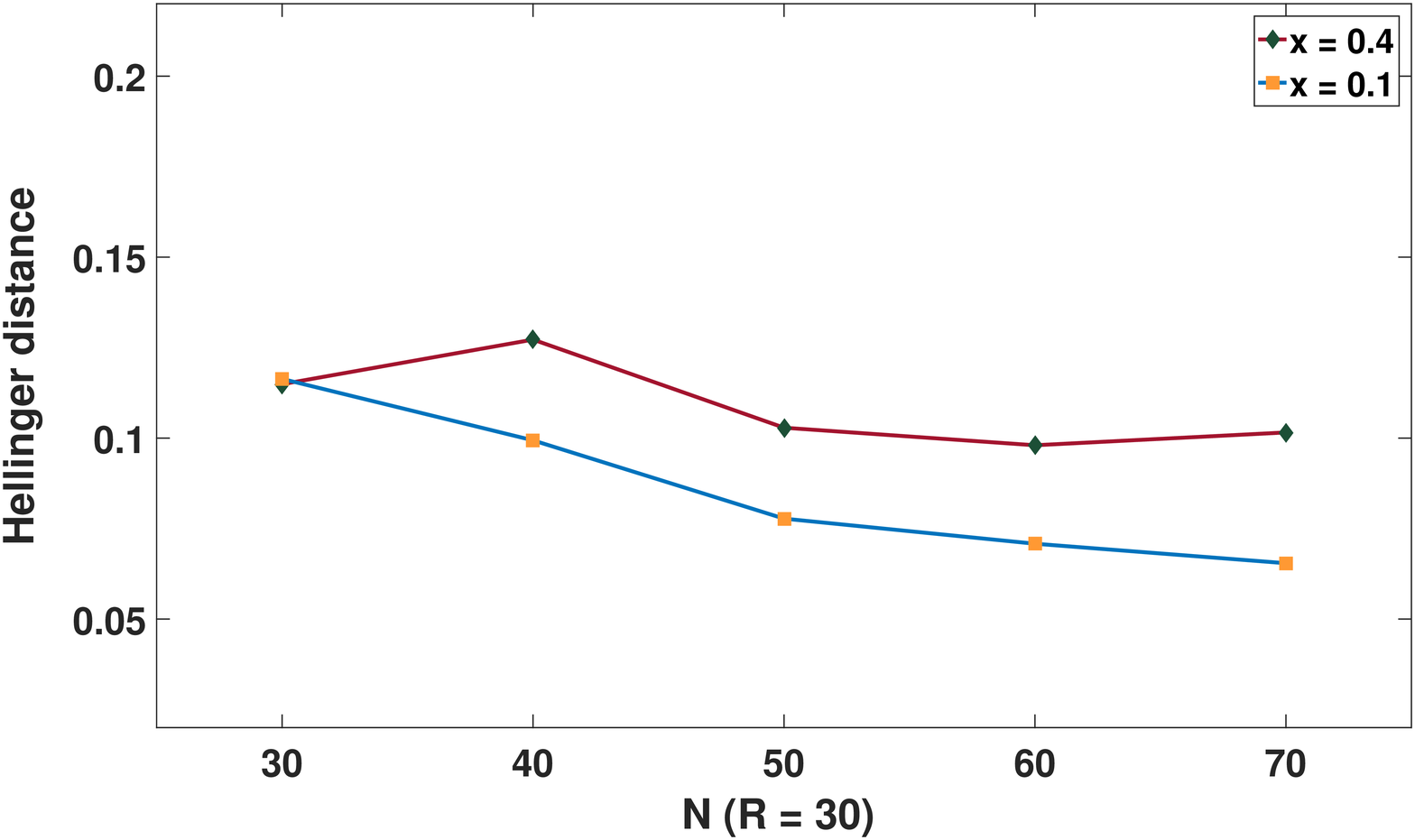}} 
    \subfigure[]{\label{subfig:lab13}\includegraphics[width=0.475\textwidth]{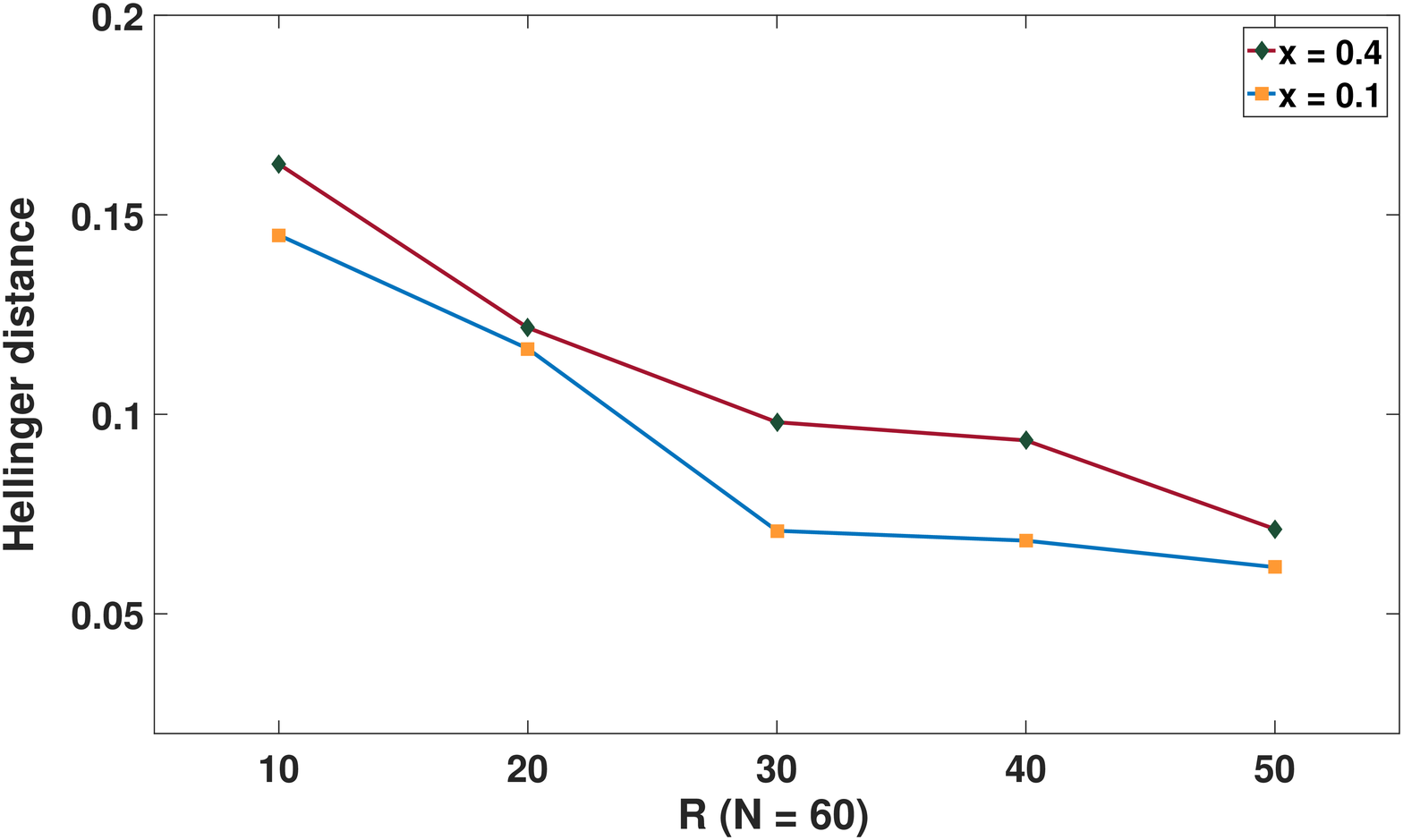}}
    \caption{ Hellinger distance between predicted PDF and true response PDF for  $x = 0.1$ and $x = 0.4$ with \textbf{(a)} variation in $N$ and $R$  for $N_{z} = 21$  \textbf{(b)} variation in $N$ for $R = 30$ and $N_{z} = 21$ \textbf{(c)} variation in $R$ for $N = 60$ and $N_{z} = 21$.}
    \label{fig:2n1}
\end{figure}

\begin{figure}[htbp!]
    \centering
    {\label{subfig:lab14}\includegraphics[width=0.5\textwidth]{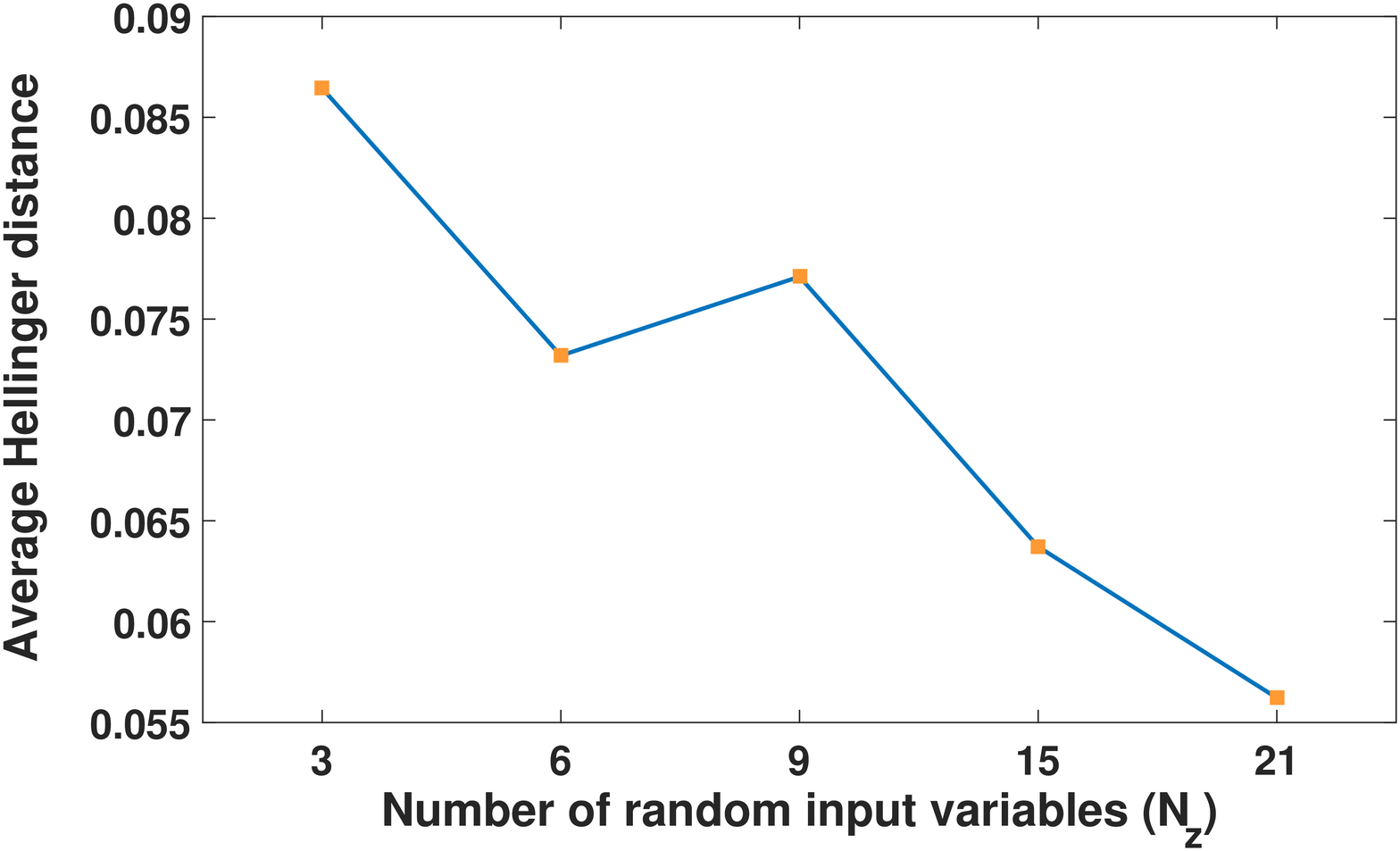}}
    \caption{ Average of Hellinger distance between predicted PDF and true response PDF for  $x = 0.1$, $x = 0.4$, and $x = 0.85$ with variation in the number of standard normal random input variables, $N_{z}$, for $N = 60$ and $R = 50$.}
    \label{fig:2n2}
\end{figure}

Fig. \ref{fig:1} shows the PDF predictions by the deep neural network for $x = 0.1$ and $x = 0.85$ with $N = 60$, $R = 50$, and $N_{z} = 21$. Observation reveal that the proposed approach is able to accurately predict the PDF of response Y at the two $x$-values.
 Fig. \ref{fig:2n1} shows the influence of $N$ and $R$ on the performance of proposed approach for two different input parameter values with Hellinger distance between the true and predicted PDF of their response as the error metric. It could be clearly seen from Fig. \ref{subfig:lab11} that as $N\times R$ increases and $N_{z}$ is kept the same, the error metric follows a generally downward trend. Also, Fig. \ref{subfig:lab12}, shows that for fixed $R$ and $N_{z}$, as number of points of the experimental design, $N$, are increased, the error again follows a generally downward trend. This indicates that the proposed framework is performing as expected. Similar trends are observed for increase in $R$ with a fixed $N$ and $N_{z}$ in Fig. \ref{subfig:lab13}. However, perhaps more importantly, it is also observed that difference between the errors for different values of $N$ and $R$ is not substantial, indicating the potential use of the proposed approach for scenarios where limited replications are available. Finally, for increase in $N_{z}$ with fixed $N$ and $R$, we compute the mean Hellinger distances, and again, we see a generally downward trend in the error as presented in Fig. \ref{fig:2n2}. 

\begin{figure}
    \centering
    \subfigure[]{\includegraphics[width=0.45\textwidth]{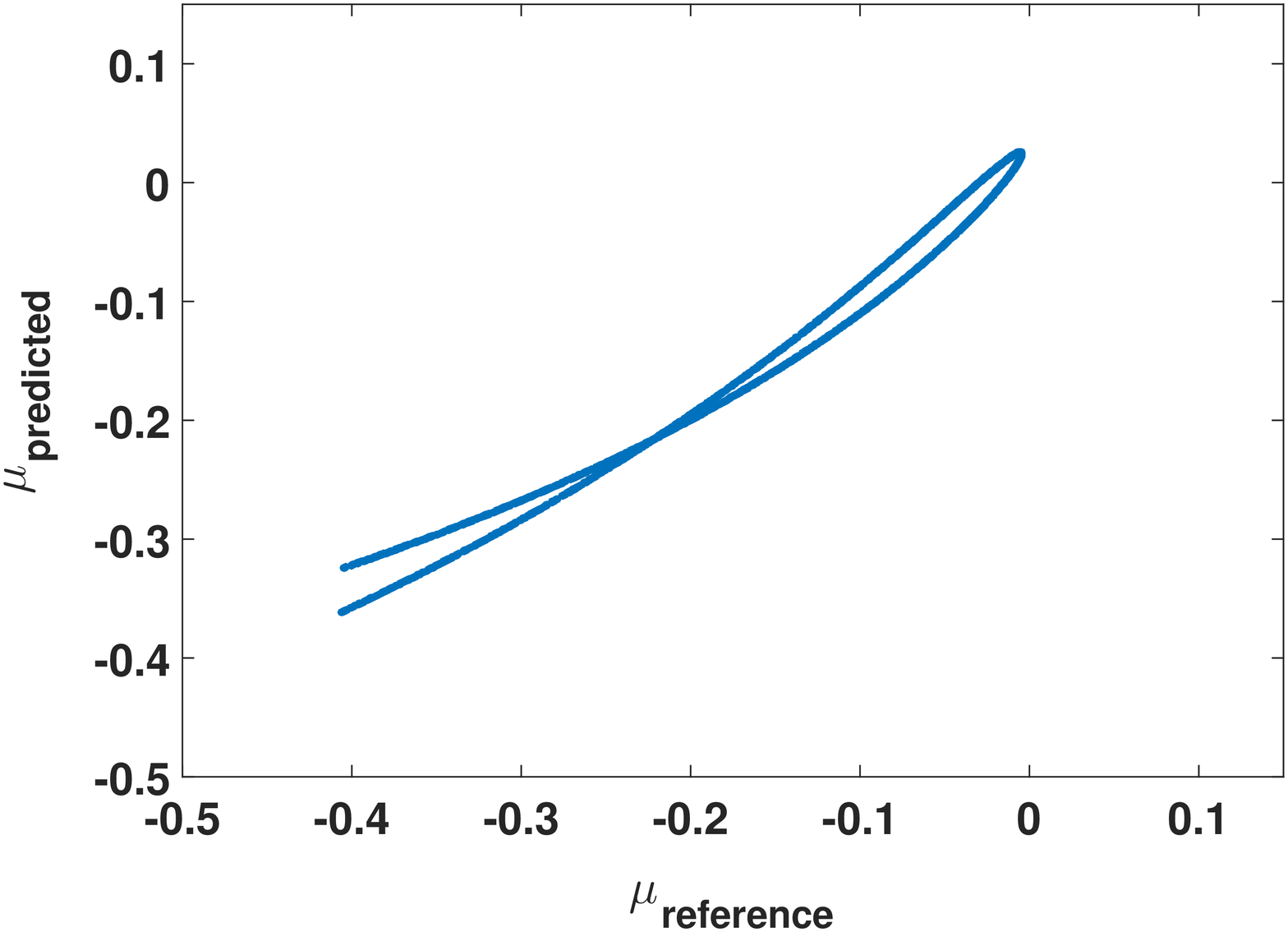}} 
    \subfigure[]{\includegraphics[width=0.45\textwidth]{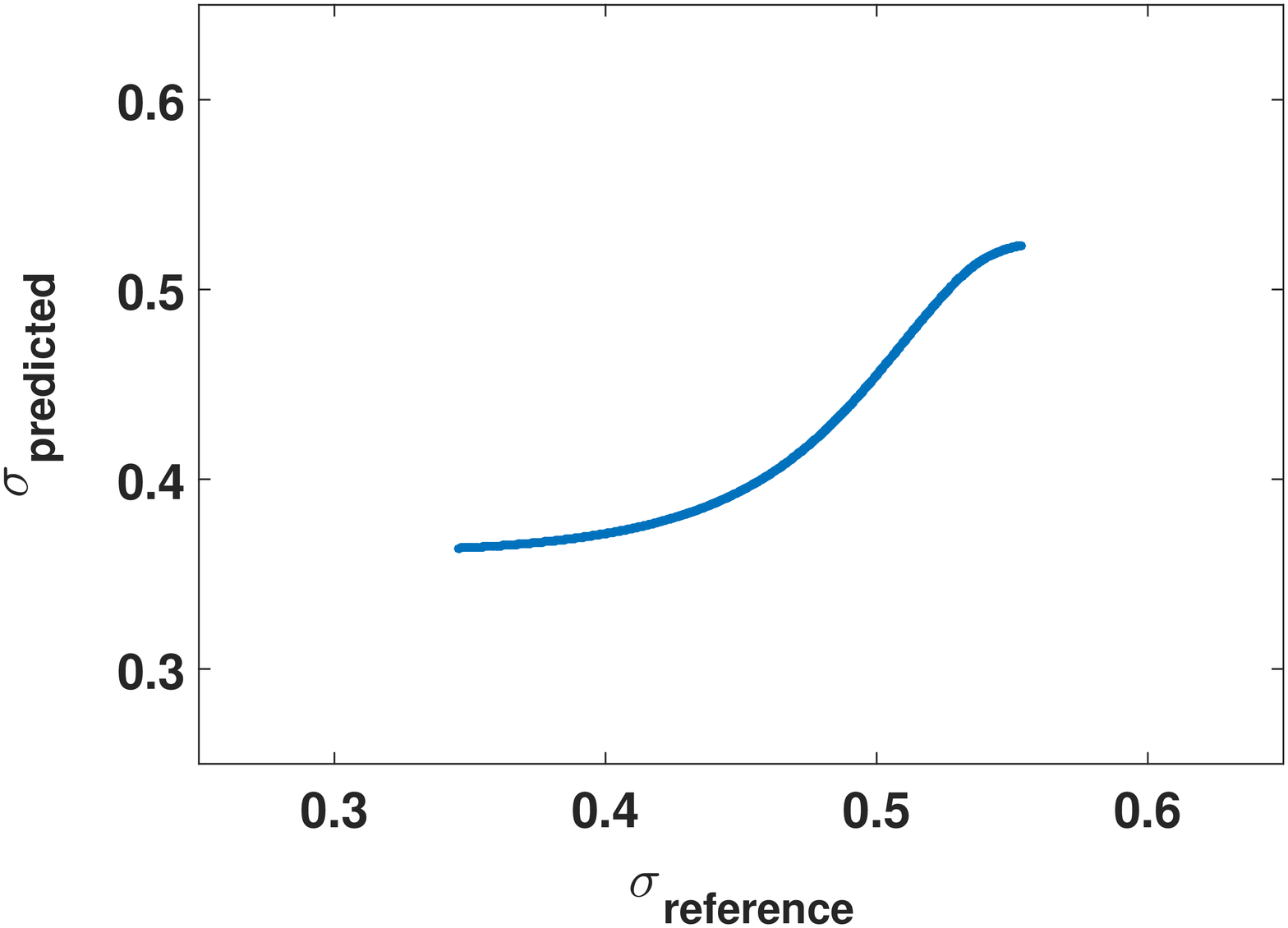}} 
    \subfigure[]{\includegraphics[width=0.45\textwidth]{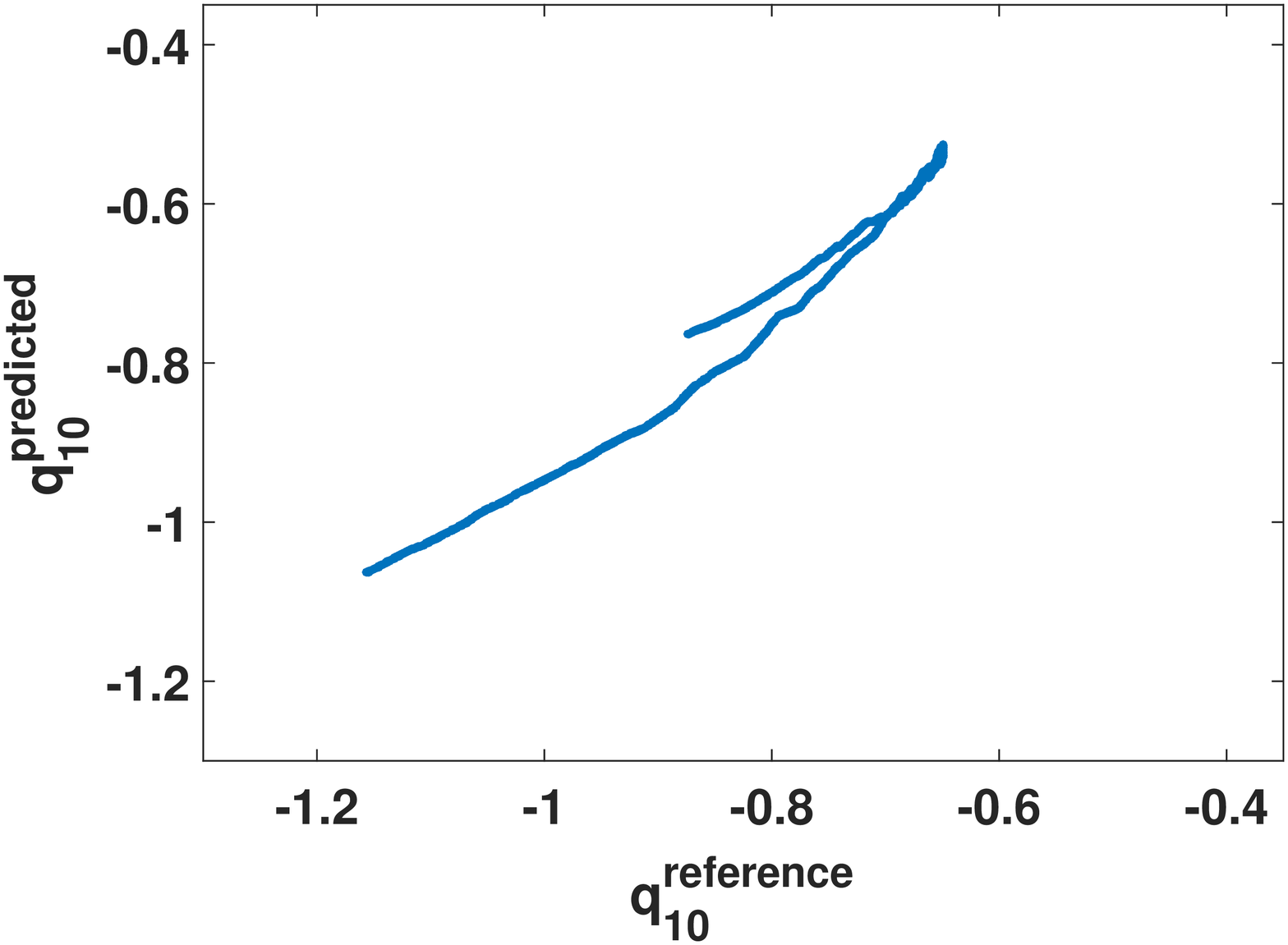}}
    \subfigure[]{\includegraphics[width=0.45\textwidth]{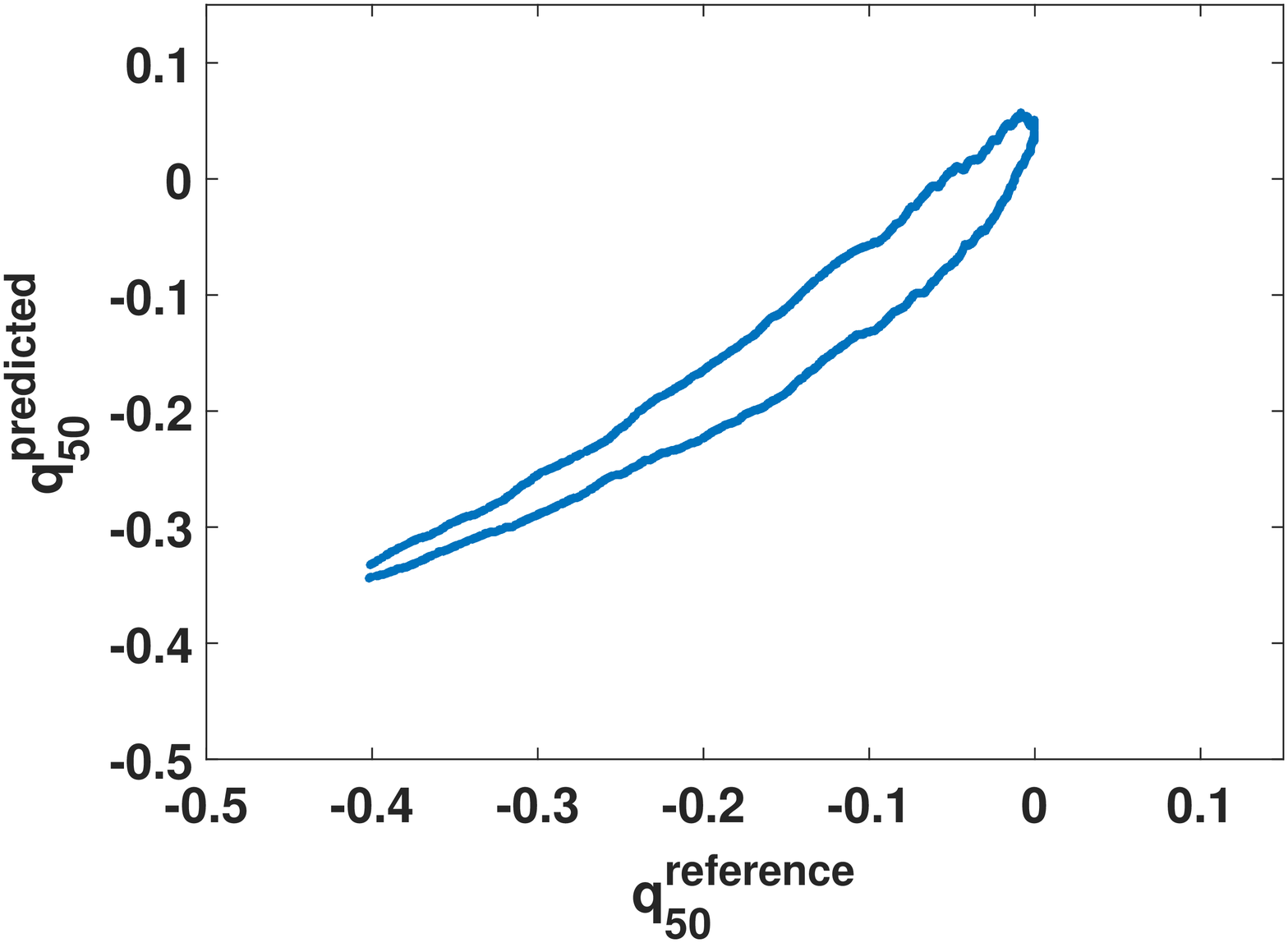}}
    \subfigure[]{\includegraphics[width=0.45\textwidth]{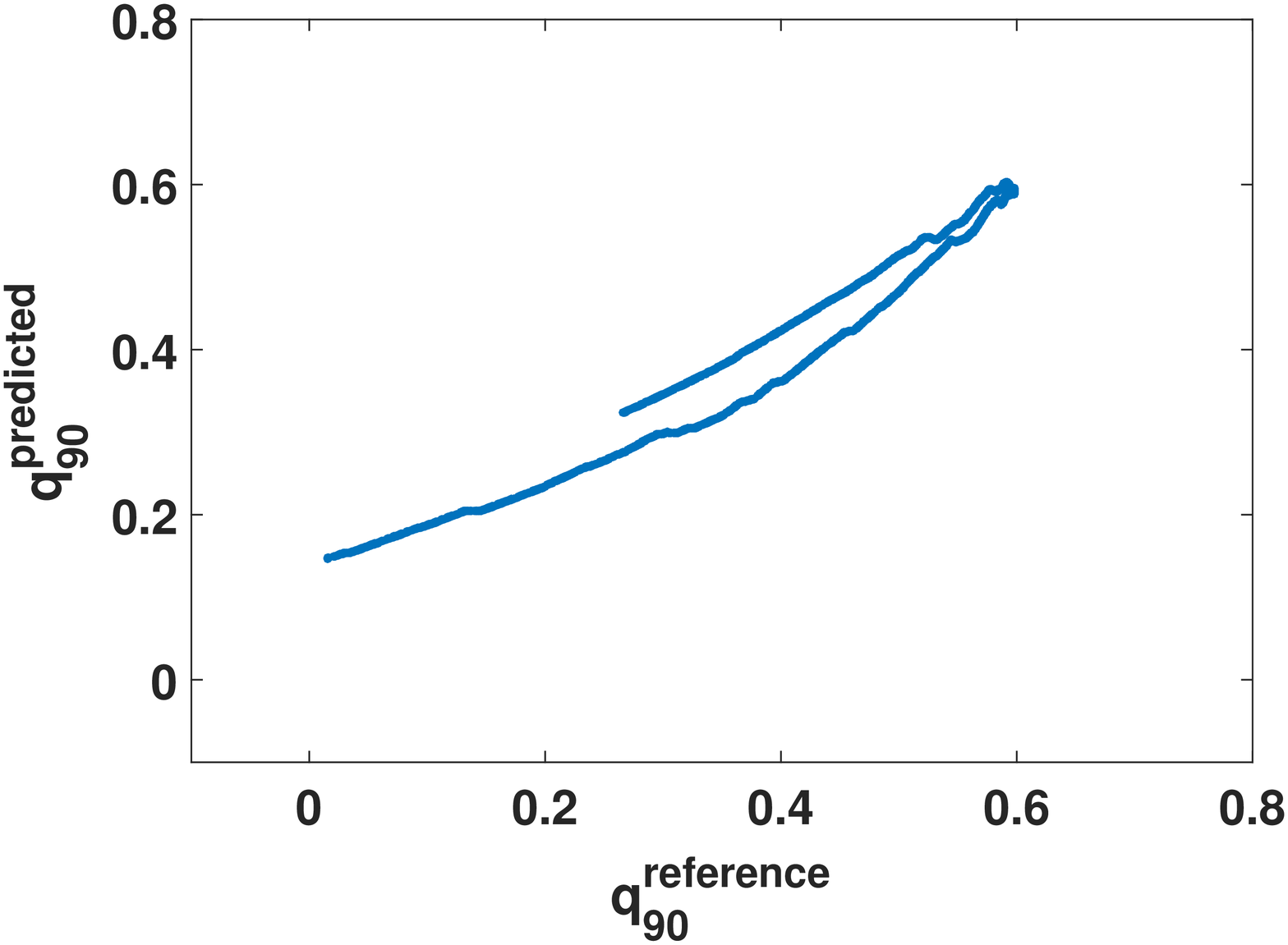}}
    \caption{ Comparison of predicted response distribution with true response distribution for 4000 $X$-values with  the help of \textbf{(a)} mean  \textbf{(b)} standard deviation \textbf{(c)} 10\% quantile   \textbf{(d)} 50\% quantile \textbf{(e)} 90\% quantile.}
    \label{fig:3}
\end{figure}

The accuracy of the developed model across the entire input parameter range is evaluated by comparison of model predicted response distribution statistics for 4000 values of input parameter, $X$, with true response distribution statistics for $N = 60$, $R = 50$, and $N_{z} = 21$. The results for mean, standard deviation, 10\%, 50\%, and 90\% quantiles are provided in the Fig. \ref{fig:3}.
Finally, the Hellinger distance between the predicted response distributions and true response distribution is computed for same values of input vector $X$ to provide a quantitative value of the error across the range. The various error statistics are provided in the Table \ref{tab:1}
\begin{table}[htbp!]
    \centering
    \begin{tabular}{||c | c | c | c | c||} 
    \hline
    Mean & Standard deviation & 10\% quantile  & 50\% quantile  & 90\% quantile  \\ [0.5ex] 
    \hline\hline
    0.08542 & 0.03725 & 0.05819 & 0.07623 & 0.11042  \\ 
    \hline
    \end{tabular}
    \caption{Hellinger distance based error statistics for predicted and true response distributions at 4000 $X$-values.}
    \label{tab:1}
\end{table}

\subsection{Example 2: A two-dimensional simulator}\label{ss:2}
As the second analytical example, we consider the Black-Scholes stochastic differential equation used for modeling stock prices. The equation is defined as follows:
\begin{equation}
dS_{t} = x_{1}S_{t}dt + x_{2} S_{t} dW_{t},
\label{eqn2}
\end{equation}
where $S_{t}$ is the stock price, $x_{1}$ is the drift which determines the return expected on a stock in a time interval $dt$, $x_{2}$ is the volatility of stock which determines variability of return around $x_{1}$, and $W_{t}$ is the standard Wiener process, which imparts the stochastic nature. The solution of Eq.  \eqref{eqn2}  is a stochastic process, which depends upon $x_{1}$ and $x_{2}$, while the Eq.  \eqref{eqn2} itself is defined with respect to time.

We follow the case definition provided by \cite{zhu2021emulation}, where the QoI is $Y(x_{1},x_{2}) = S_{1}(x_{1},x_{2})$, denoting the value of stock in an year and the initial condition $S_{0}(x_{1},x_{2})$ is taken to be equal to 1. Also, the parameters $x_{1}\sim\mathcal{U}(0,0.1)$ and $x_{2}\sim\mathcal{U}(0,0.4)$ have uniform distributions. The solution to the Black-Scholes SDE could be found by employing Ito's Lemma. The solution $Y(x_{1},x_{2}, \omega)$ adheres to a lognormal distribution and is defined by the following equation
\begin{equation}
Y{(x_{1},x_{2},\omega)} = \exp{\bigg(x_{1}-\frac{x_{2}^{2}}{2} + x_{2}\cdot z_{1}(\omega)\bigg)},
\label{eqn3}
\end{equation}
where $z_{1}(\omega)$ is the latent variable which follows $\mathcal{N}(0,1)$. The objective here is to develop a surrogate model for the stochastic response. To that end, the vector $(\vect{X},\vect{S})^{T}$ serves as the input to the neural network, where $\vect{X}$ is $(x_{1},x_{2})$ and $\vect{S}$ is a vector consisting of $M$ standard normal random variables, i.e., $(s_{1},s_{2},...,s_{M})$. Also, linear activation function is used for the output layer, the batch size is set to 300, and the number of epochs required for successful training is 250.
Evidently, the presence of a closed form solution as a lognormal distribution eliminates the need of simulating the entire process, and therefore, Eq.  \eqref{eqn3} is used for generating the samples.

\begin{figure}[htbp!]
    \centering
    \subfigure[]{\includegraphics[width=0.45\textwidth]{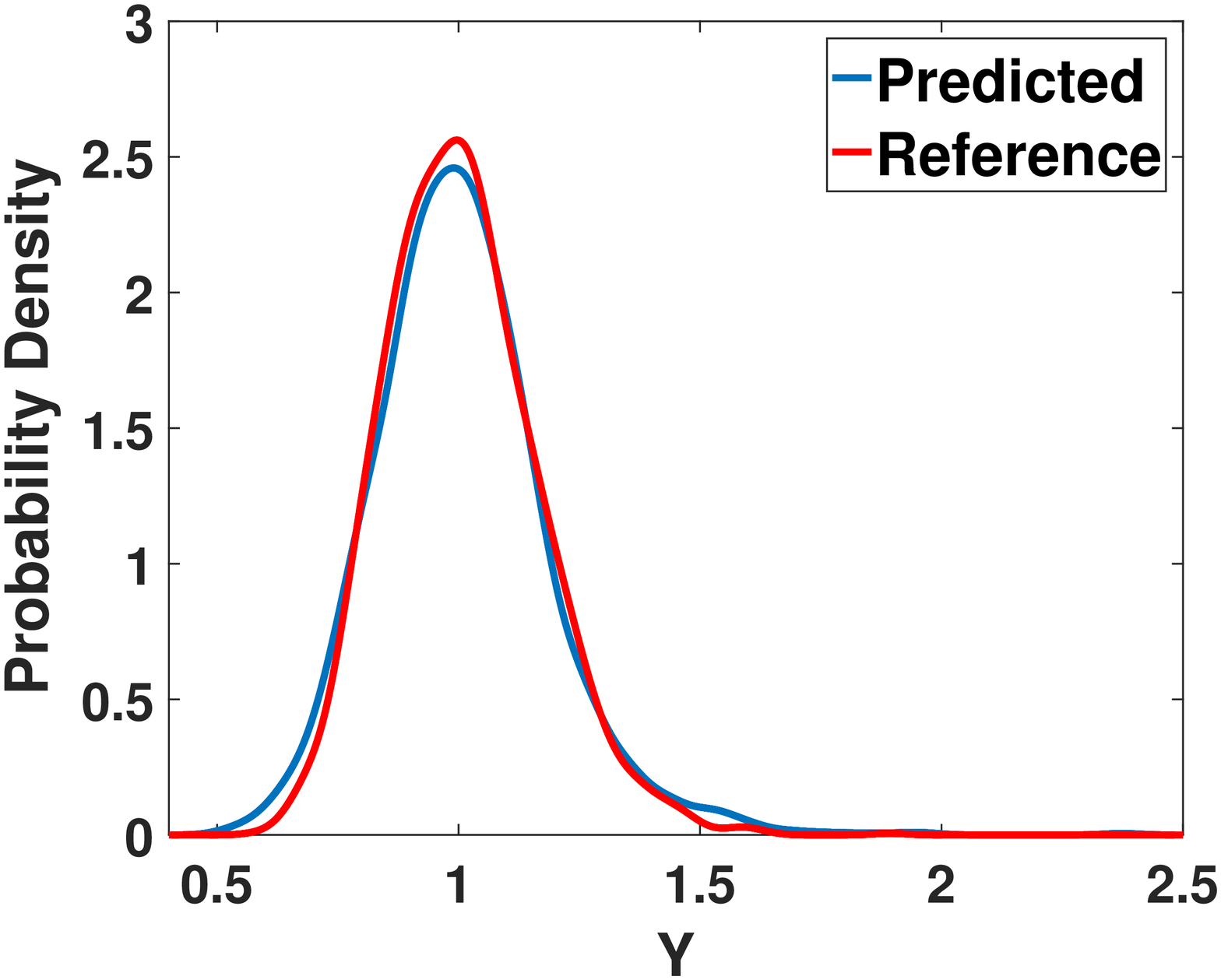}} 
    \subfigure[]{\includegraphics[width=0.45\textwidth]{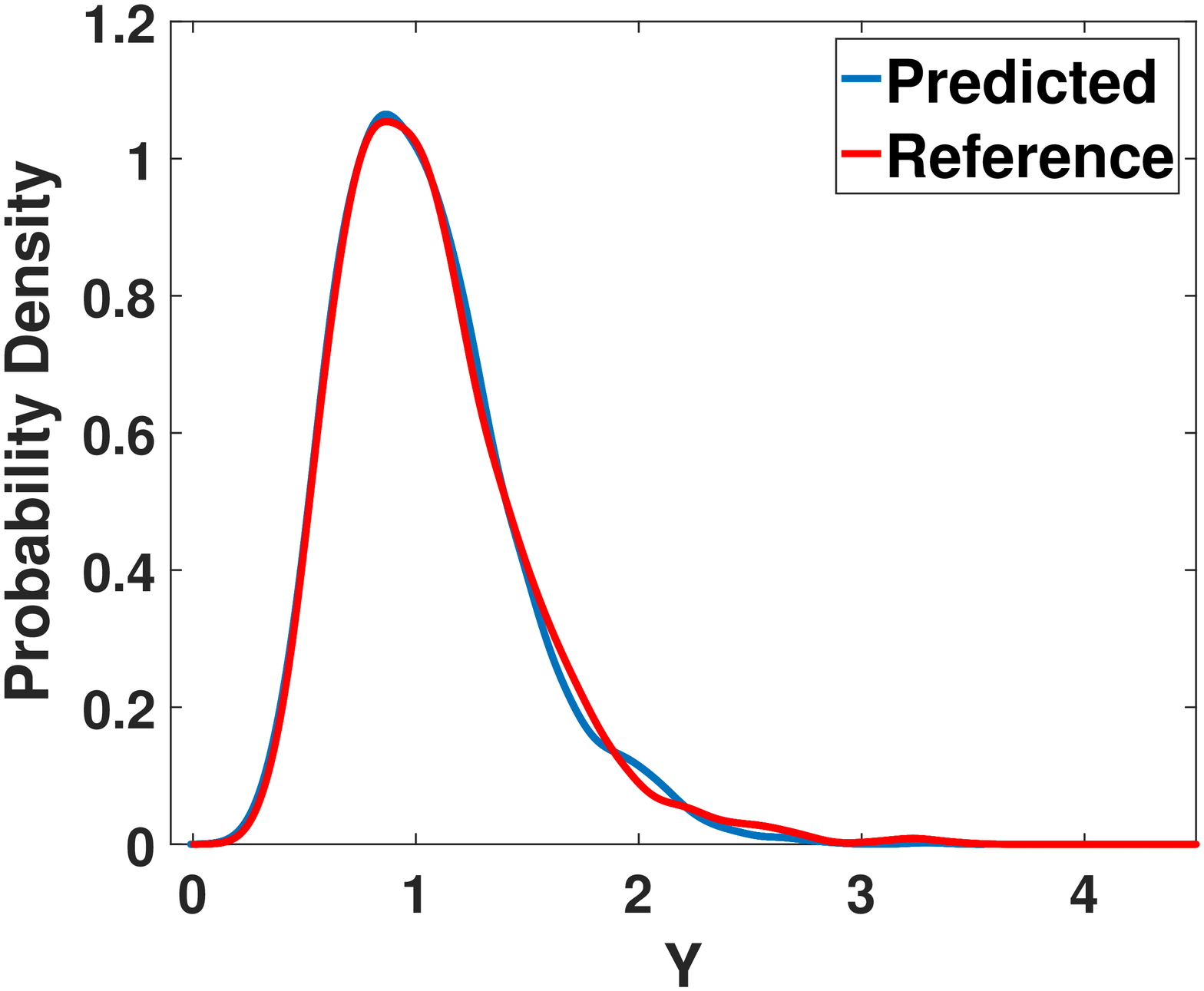}}
    \caption{ \textbf{(a)} PDF prediction for $x_{1} = 0.08$ and $x_{2} = 0.375$   \textbf{(b)} PDF prediction for $x_{1} = 0.01$ and $x_{2} = 0.15$}
    \label{fig:4}
\end{figure}

\begin{figure}[htbp!]
    \centering
    \subfigure[]{\label{subfig:lab5n1}\includegraphics[width=0.475\textwidth]{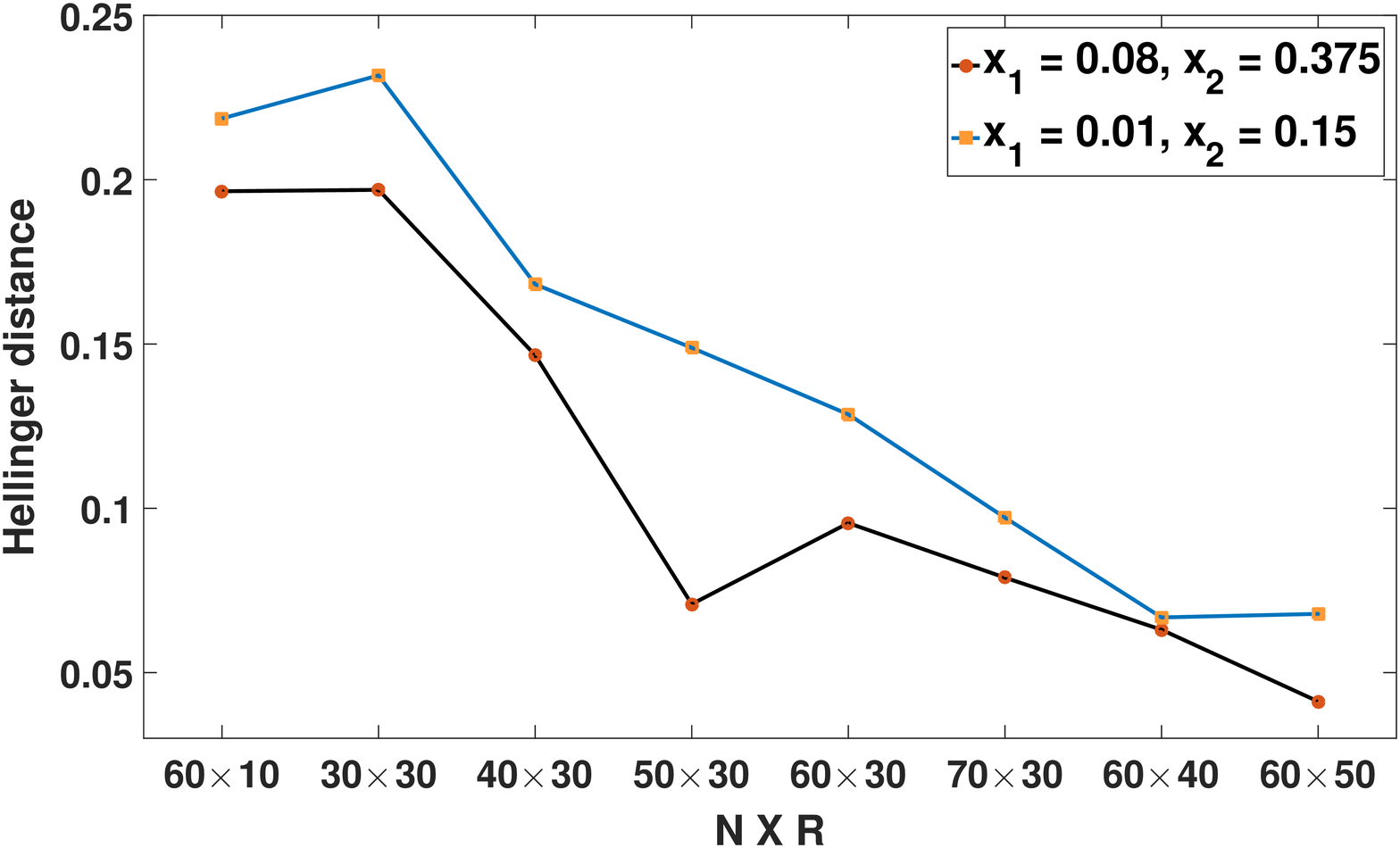}}
    \subfigure[]{\label{subfig:lab5n2}\includegraphics[width=0.475\textwidth]{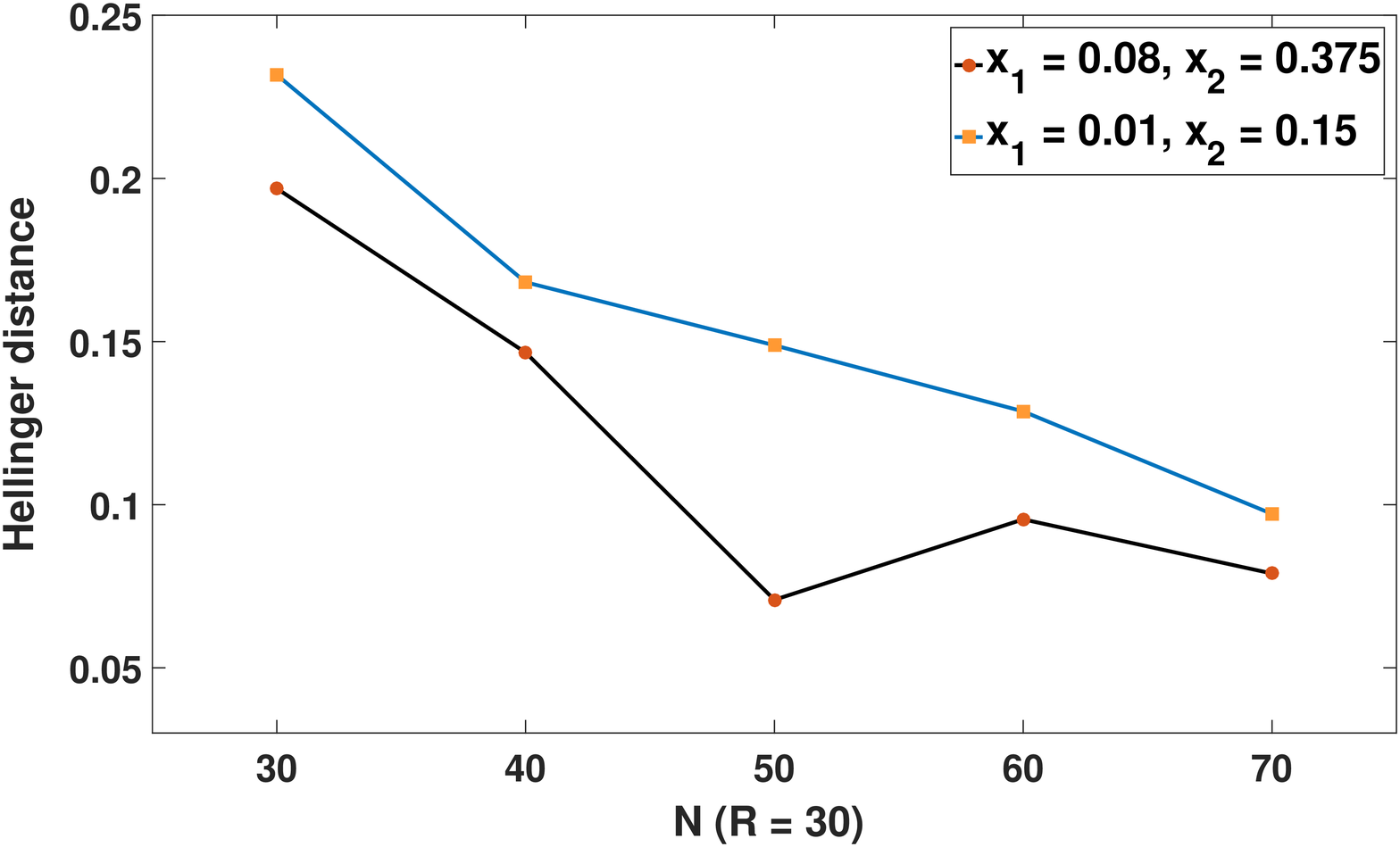}} 
    \subfigure[]{\label{subfig:lab5n3}\includegraphics[width=0.475\textwidth]{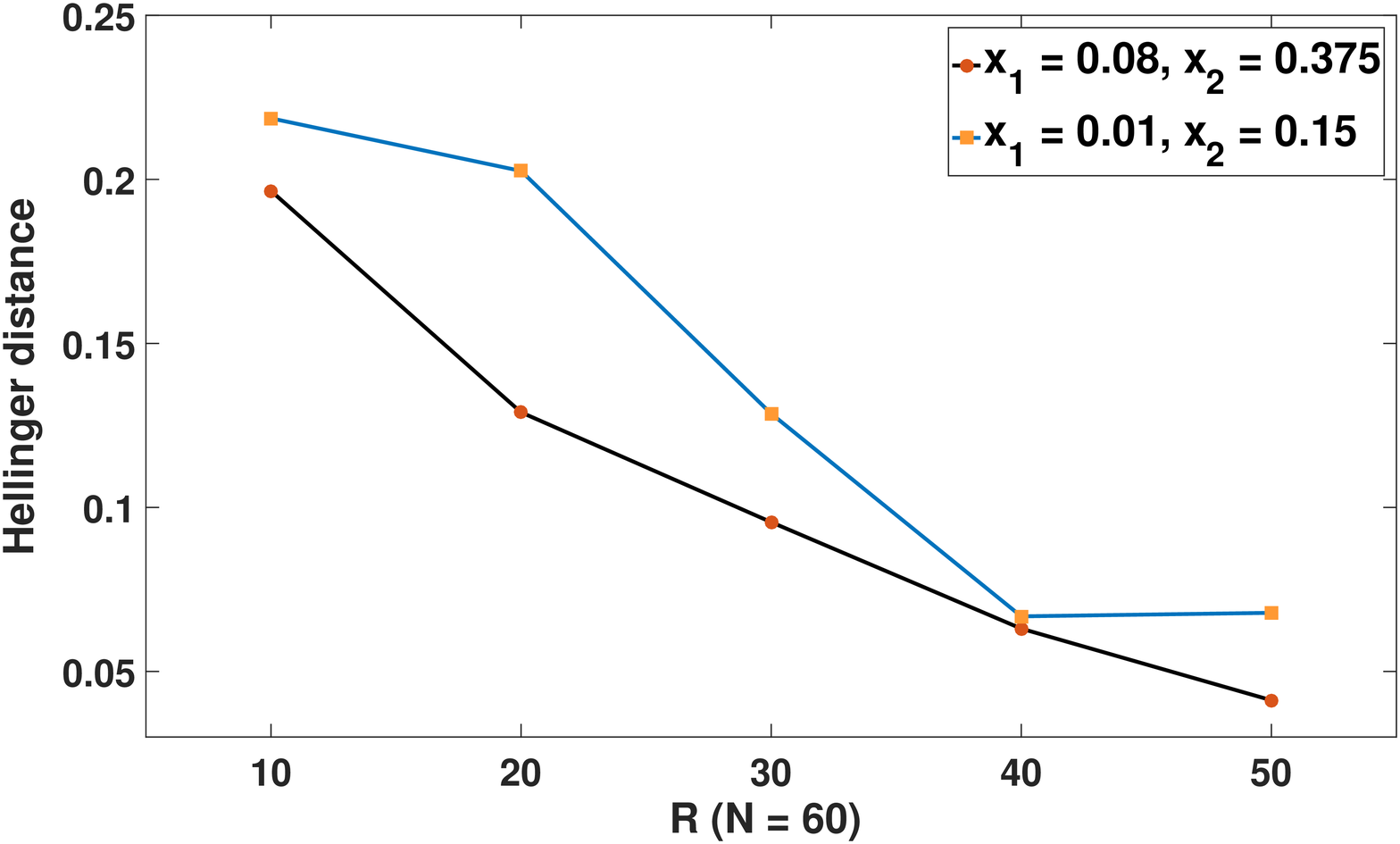}}
    \caption{Hellinger distance between predicted PDF and true response PDF for  $x_{1} = 0.08$ and $x_{2} = 0.375$, $x_{1} = 0.01$ and $x_{2} = 0.15$ with \textbf{(a)} variation in $N$ and $R$ for $N_{z} = 18$  \textbf{(b)} variation in $N$ for $R = 30$ and $N_{z} = 18$ \textbf{(c)} variation in $R$ for $N = 60$ and $N_{z} = 18$.}
    \label{fig:5nn1}
\end{figure}

\begin{figure}[htbp!]
    \centering
    {\includegraphics[width=0.5\textwidth]{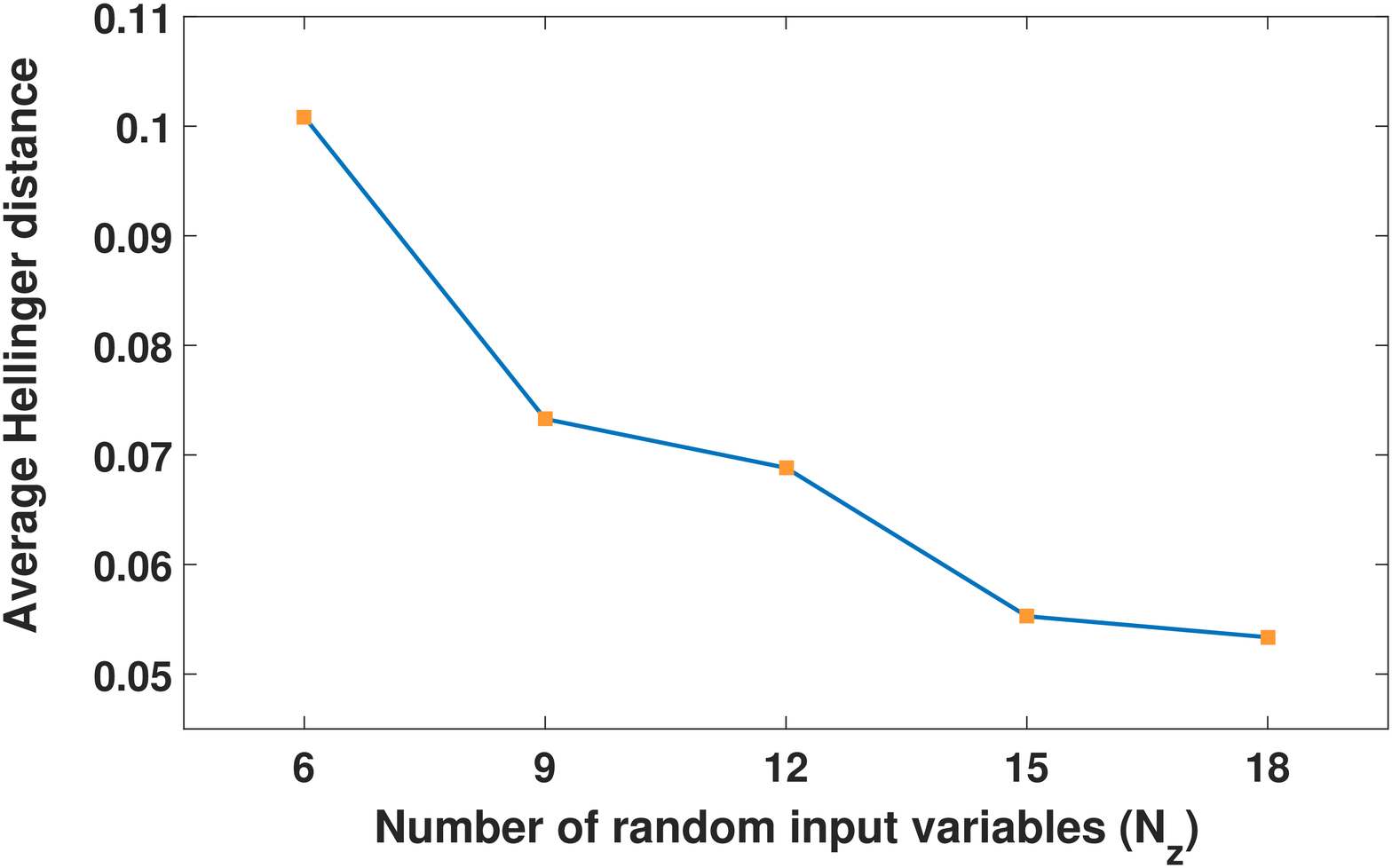}}
    \caption{Average of Hellinger distance between predicted PDF and true response PDF for  $x_{1} = 0.08$ and $x_{2} = 0.375$, $x_{1} = 0.05$ and $x_{2} = 0.23$ with variation in the number of standard normal random input variables, $N_{z}$, for $N = 60$ and $R = 50$.}
    \label{fig:5nn2}
\end{figure}

\begin{figure}[htbp!]
    \centering
    \subfigure[]{\includegraphics[width=0.45\textwidth]{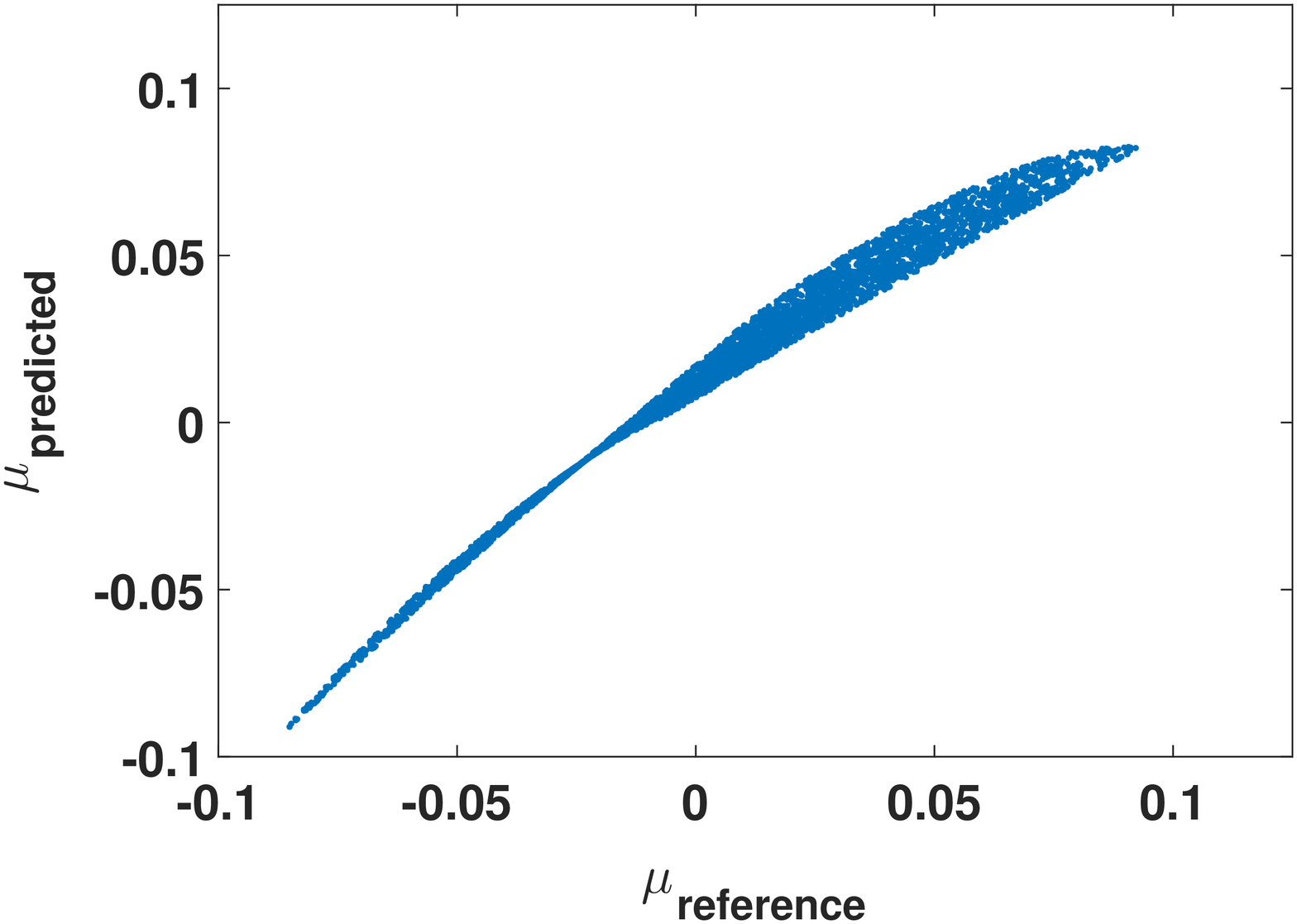}} 
    \subfigure[]{\includegraphics[width=0.45\textwidth]{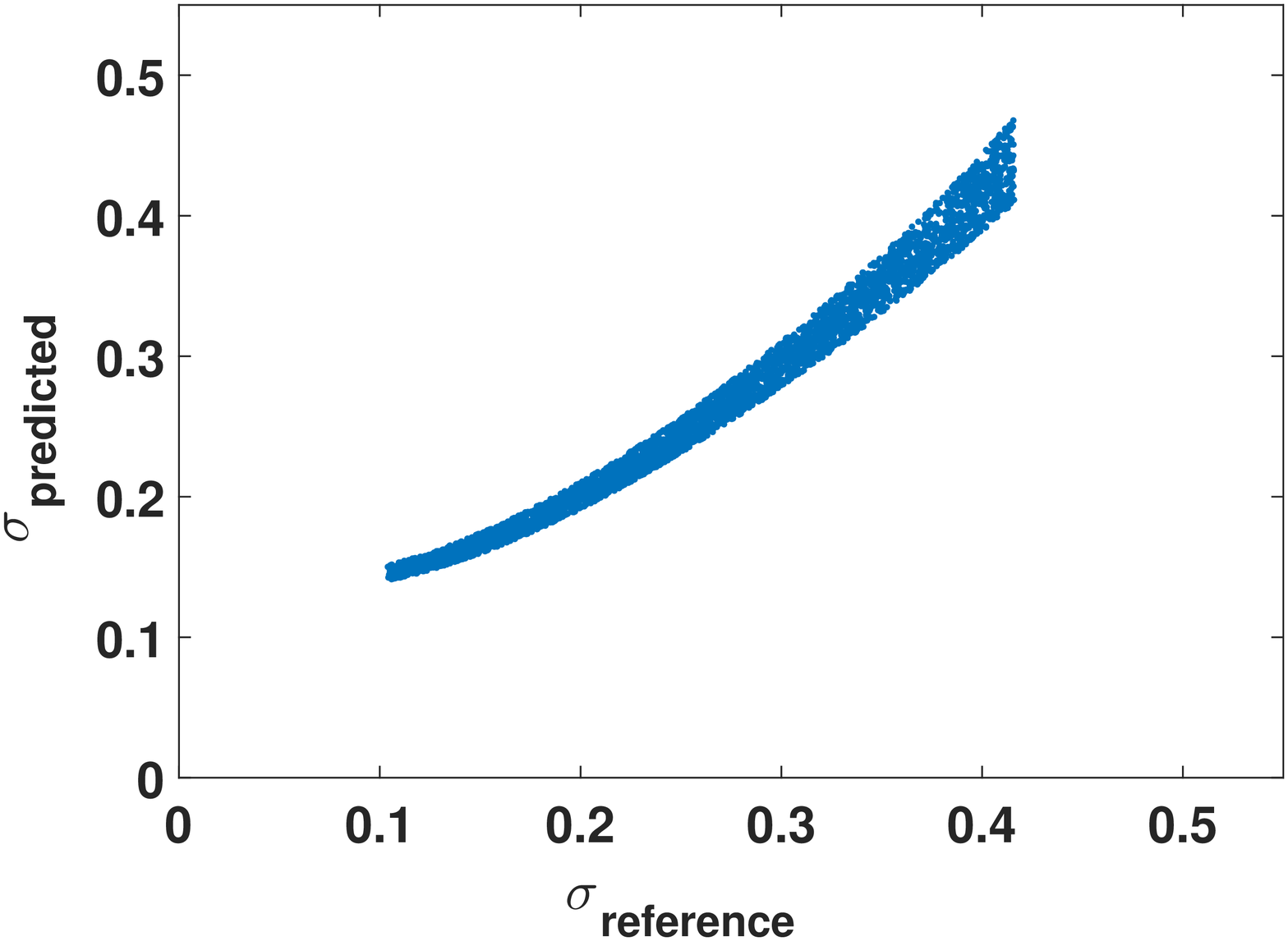}} 
    \subfigure[]{\includegraphics[width=0.45\textwidth]{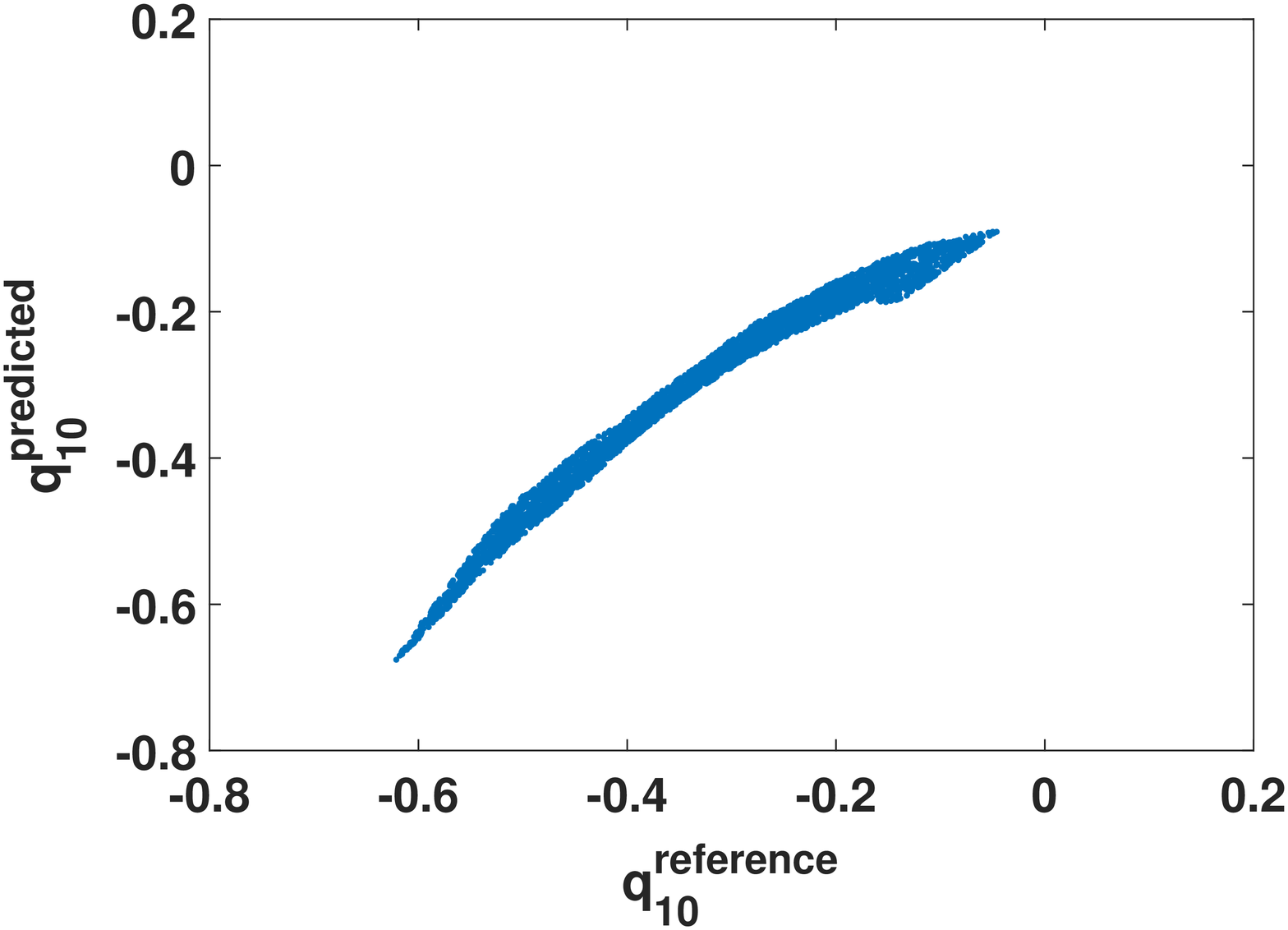}}
    \subfigure[]{\includegraphics[width=0.45\textwidth]{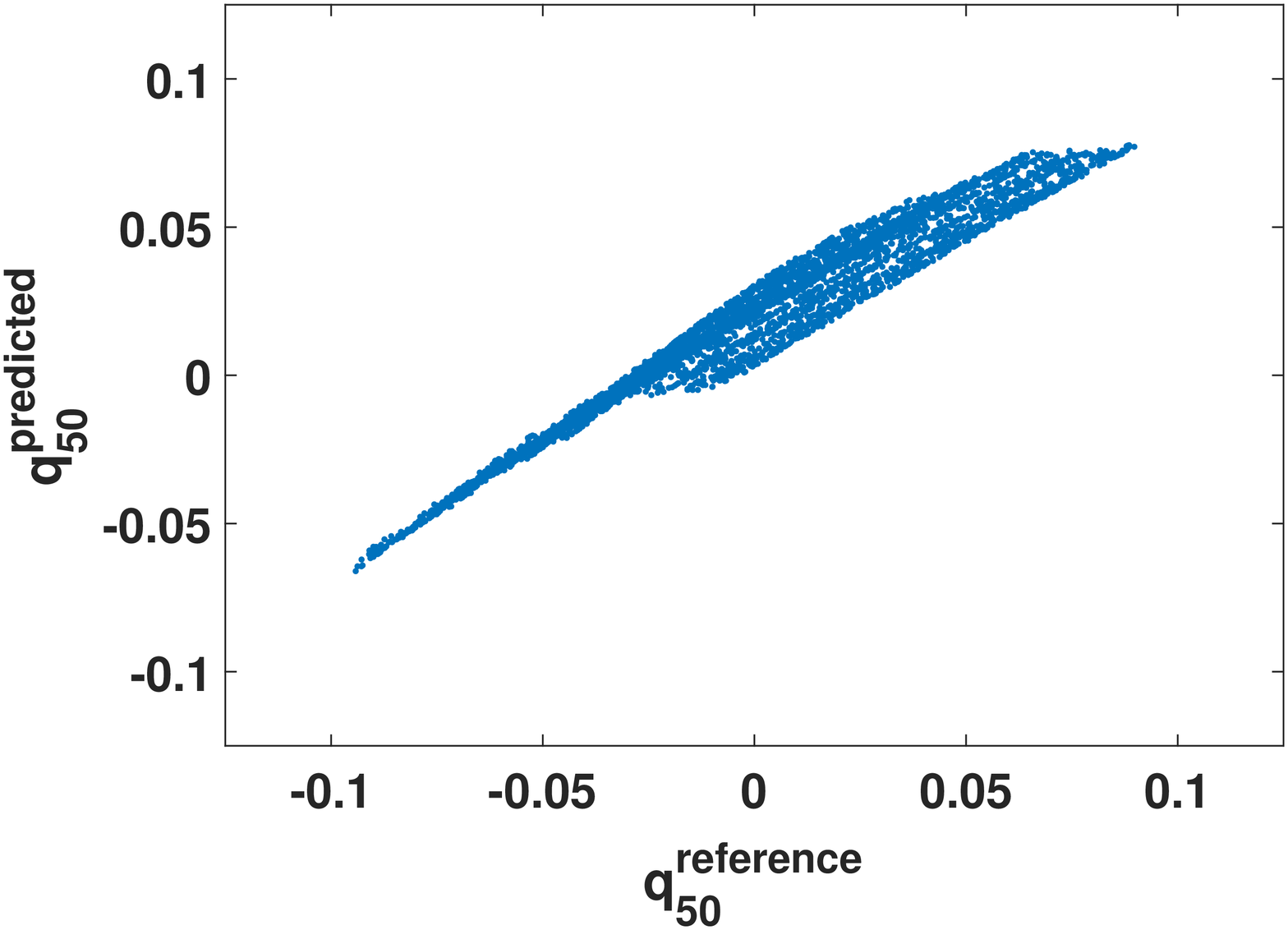}}
    \subfigure[]{\includegraphics[width=0.45\textwidth]{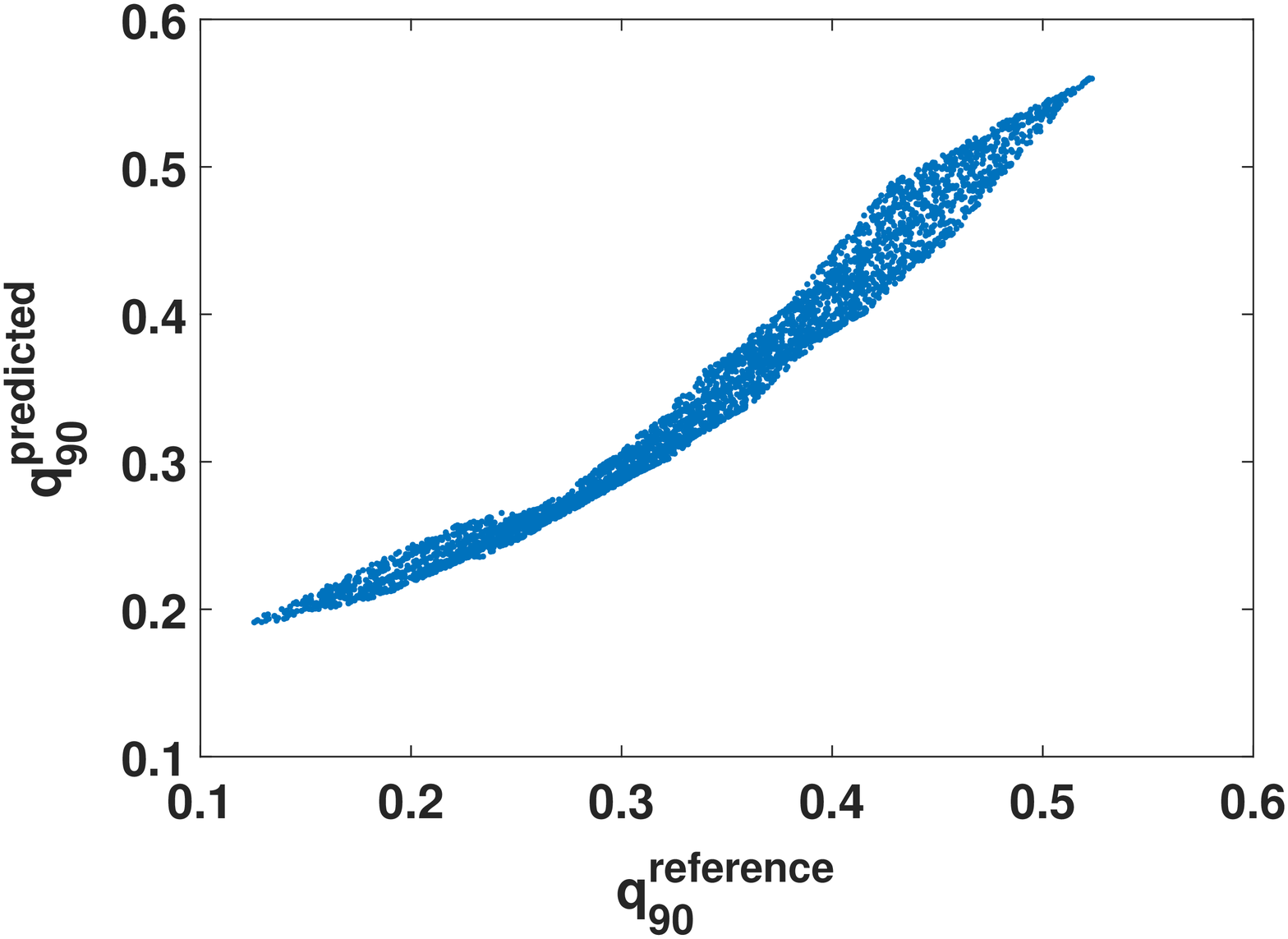}}
    \caption{ Comparison of predicted response distribution with true response distribution for 4000 values of input parameter $\vect{X}$ with  the help of \textbf{(a)} mean  \textbf{(b)} standard deviation \textbf{(c)} 10\% quantile   \textbf{(d)} 50\% quantile \textbf{(e)} 90\% quantile.}
    \label{fig:6}
\end{figure}

The PDF predictions by the model for $x_{1} = 0.08$, $x_{2} = 0.375$ and $x_{1} = 0.01$, $x_{2} = 0.15$ are shown in Fig. \ref{fig:4} with $N = 60$, $R = 50$, and $N_{z} = 18$. Results obtained indicate excellent performance of the proposed approach. The effect of $N$ and $R$ on the performance of proposed approach for current example is evaluated on two different sets of input parameter values. The results of the analysis are presented in Fig. \ref{fig:5nn1}. It is observed from Fig. \ref{subfig:lab5n1} that as the value of $N\times R$ is increased for a fixed $N_{z}$, the error metric follows a downward trend. Furthermore, the results presented in Fig. \ref{subfig:lab5n2} show that as $N$ is increased for fixed $R$ and $N_{z}$, the error again follows a downward trend. Likewise, for increase in $R$ with a fixed $N$ and $N_{z}$ decreasing error is observed and results are presented in Fig. \ref{subfig:lab5n3}. Finally, for increase in $N_{z}$ with fixed $N$ and $R$, again the mean Hellinger distances is computed. A decreasing trend is observed as shown in Fig. \ref{fig:5nn2}.
Again, the accuracy of the model predictions is again evaluated by comparing the statistics of predicted response distribution with that of the true response distribution for a set of 4000 values of input parameter $\vect{X}$. The value of $N$, $R$, and $N_{z}$ are 60, 50, and 18, respectively. The results are presented in Fig. \ref{fig:6}. We observe that the predicted results matches quite well with the actual results. For quantitative assessment, the Hellinger distance is computed between predicted and true response PDF for same values of input parameter $\vect{X}$, and the statistics are presented in Table \ref{tab:2}.
\begin{table}[htbp!]
    \centering
    \begin{tabular}{||c | c | c | c | c||} 
    \hline
    Mean & Standard deviation & 10\% quantile  & 50\% quantile  & 90\% quantile  \\ [0.5ex] 
    \hline\hline
    0.07168 & 0.02115 & 0.04888 & 0.06949 & 0.099391  \\ 
    \hline
    \end{tabular}
    \caption{Hellinger distance based error statistics for predicted and true response distributions at 4000 values of input parameter $\vect{X}$.}
    \label{tab:2}
\end{table}

\subsection{Example 3: Stochastic differential equation without a closed form solution}\label{ss:3}
Stochastic differential equations (SDEs) find widespread utility in fields ranging from epidemiology, biology to economics and physics, due to their ability to model complex systems. However, most of the times, the SDEs that come up in these fields do not have an analytical closed form solution (the previous example was an exception in that sense). Therefore, numerical methods such as the MC methods or the Euler-Maruyama method \cite{kloeden1992stochastic,tripura2021change} have to be used in order to solve such systems. This introduces a considerable number of latent variables in the response (solution of the SDE computed at a time t) as compared to response of systems which admit a closed form solution, and thus, makes the problem at hand much more complex. Therefore, in order to test the predictive capabilities of our deep learning based surrogate modeling and to display it's ability to model complex practical phenomena, for our third example, we consider the following problem as proposed by \citet{jimenez2017nonintrusive} and modified by \citet{zhu2020replication}
\begin{equation}
dY_{t} = (x_{1} - Y_{t})\,dt + (\nu Y_{t}+1)\,x_{2}\,dW_{t},
\label{eqn4}
\end{equation}
where $\vect{X} = (x_{1},x_{2})$ are the uncertain input parameters of the SDE, $W_{t}$ is the standard Wiener process, and the initial condition $Y_{0}$ is almost certainly equal to $0$. The solution of Eq.  \eqref{eqn4} at $\vect{X} = \vect{x}$ is denoted with $Y_{t}(\vect{x})$ and the QoI for this example is the value of solution at time, $t = 10$, denoted as $Y_{10}(\vect{x})$. The objective here is to use the proposed approach to develop a surrogate model for the stochastic differential equation. To that end, the vector, $(\vect{X},\vect{S})^{T}$, serves as the input to the neural network, where $\vect{X}$ is $(x_{1},x_{2})$ and $\vect{S}$ is a vector containing $M$ standard normal random variables, i.e., $(s_{1},s_{2},...,s_{M})$. Also, linear activation function is used for the output layer, the batch size is set to 300, and the number of epochs required for successful training is 262.

As a special case, if the value of $\nu = 0$, then $Y_{t}(\vect{x})$ becomes the Ornstein-Uhlenbeck process and also admits an analytic closed form solution. However, for $\nu \not= 0$, a multiplicative noise is introduced into the problem and it no longer admits a closed form solution like the former. For the current study, the value of $\nu = 0.2$. We use Euler-Maruyama method \cite{kloeden1992stochastic} to solve Eq.  \eqref{eqn4}. The value for the time-step $\Delta t$ is set to $0.01$. Subsequently, as one has to perform time-marching through the discretized version of Eq.  \eqref{eqn4} for arrival at the QoI (solution at $t = 10$), the QoI contains 1000 latent variables. Furthermore, we adopt the modified definition of the current problem provided by \cite{zhu2020replication}, and thus, to get different PDF shapes for the QoI, $x_{1}\sim\mathcal{U}(0.9,2)$ and $x_{2}\sim\mathcal{U}(0.1,1)$ are considered.

\begin{figure}[htbp!]
    \centering
    \subfigure[]{\label{subfig:lab1}
                  \includegraphics[width=0.475\textwidth]{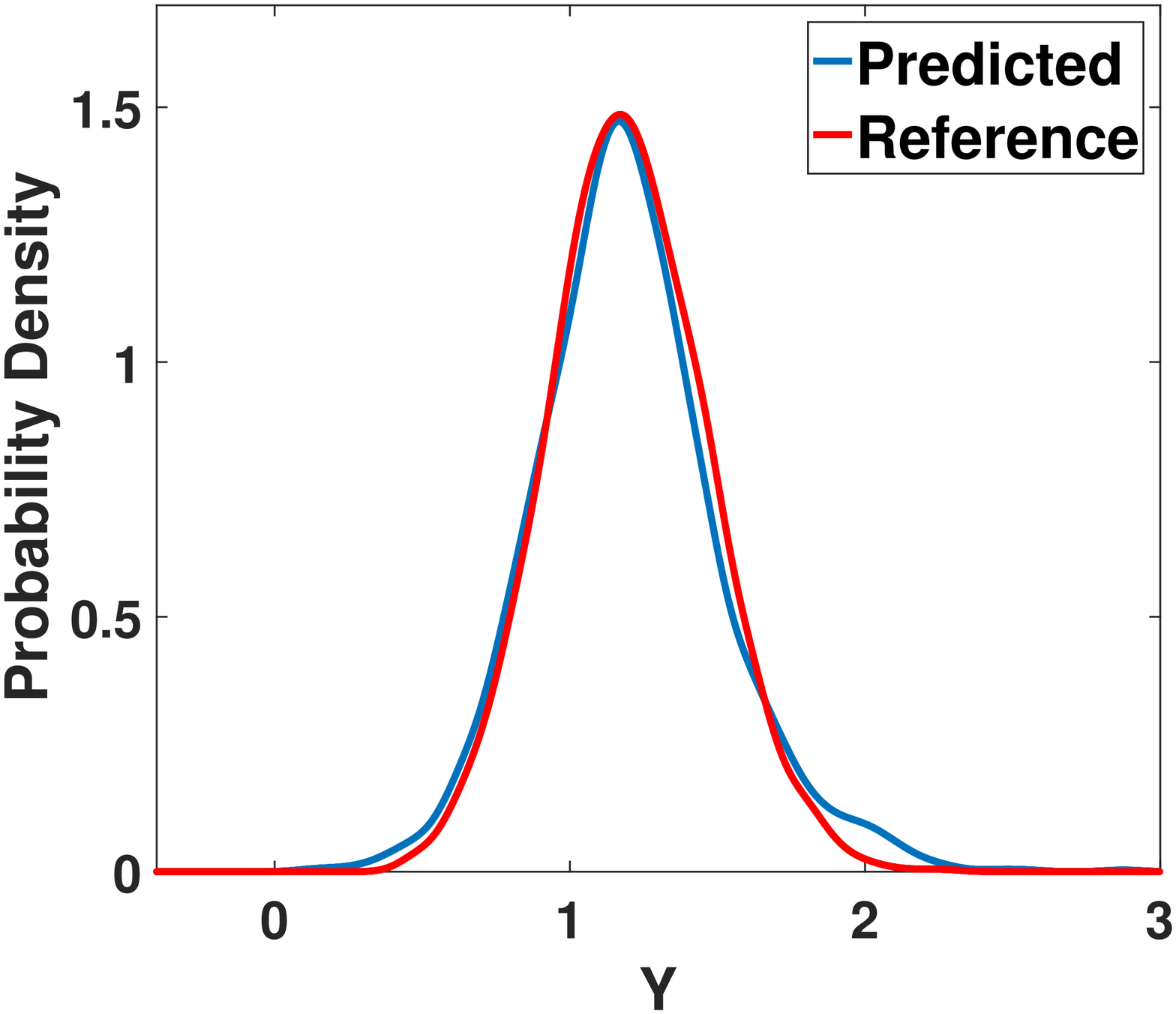}} 
    \subfigure[]{\label{subfig:lab2}
                 \includegraphics[width=0.475\textwidth]{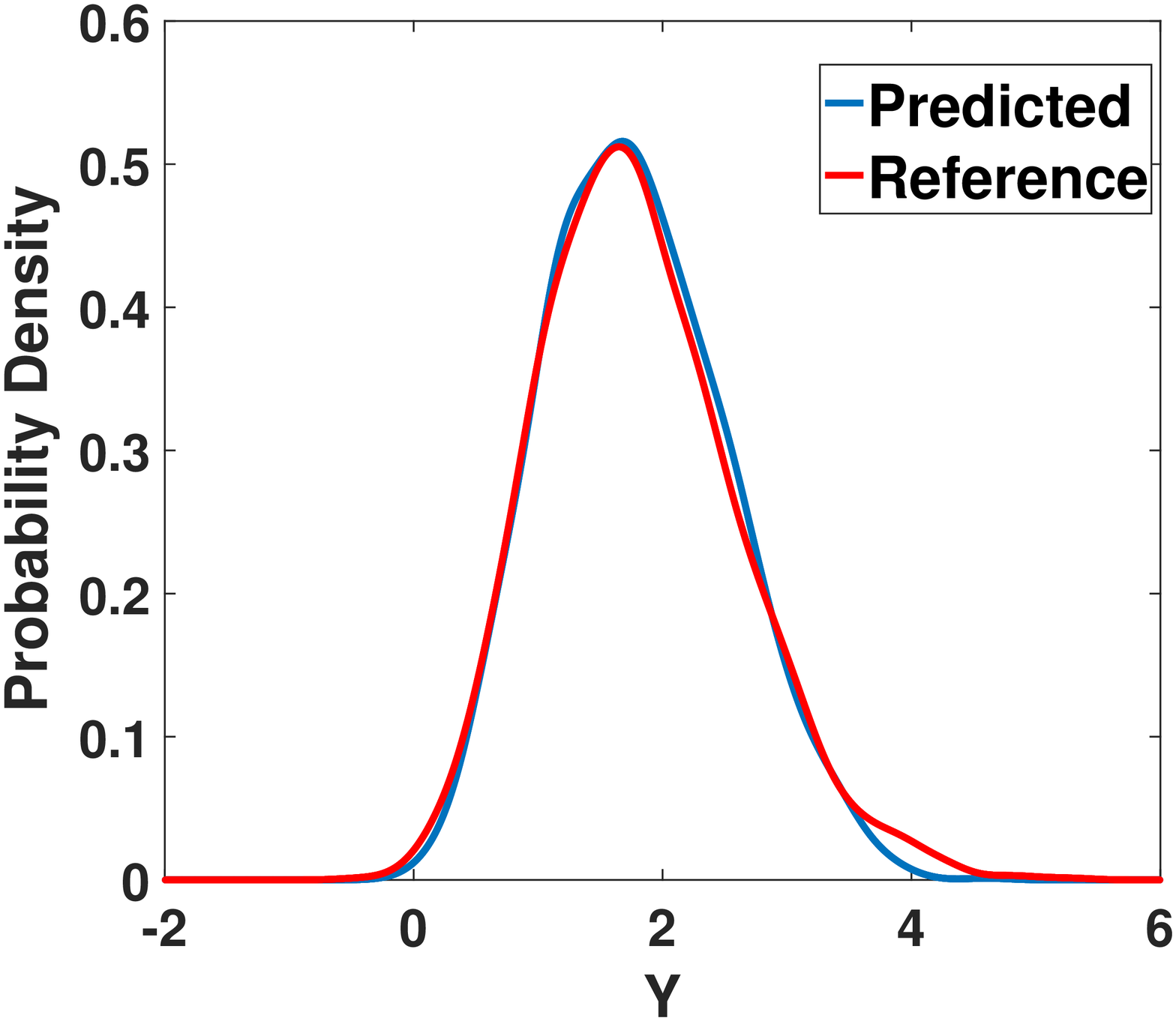}} 
    \caption{ \textbf{(a)} PDF prediction for $x_{1} = 1.2$ and $x_{2} = 0.3$   \textbf{(b)} PDF prediction for $x_{1} = 1.8$ and $x_{2} = 0.8$.}
    \label{fig:7}
\end{figure}

\begin{figure}[htbp!]
    \centering
    \subfigure[]{\label{subfig:lab8n1}\includegraphics[width=0.475\textwidth]{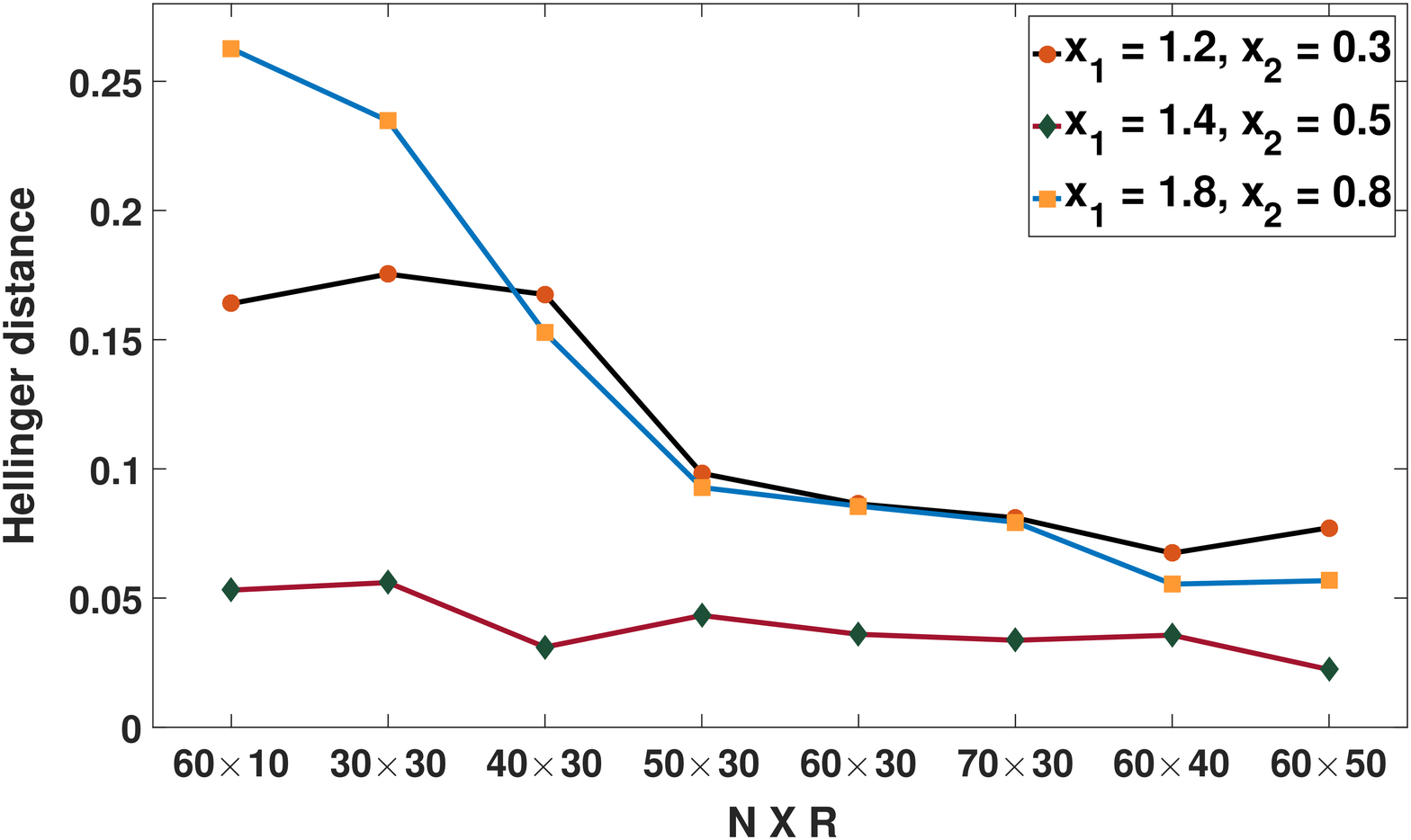}} 
    \subfigure[]{\label{subfig:lab8n2}\includegraphics[width=0.475\textwidth]{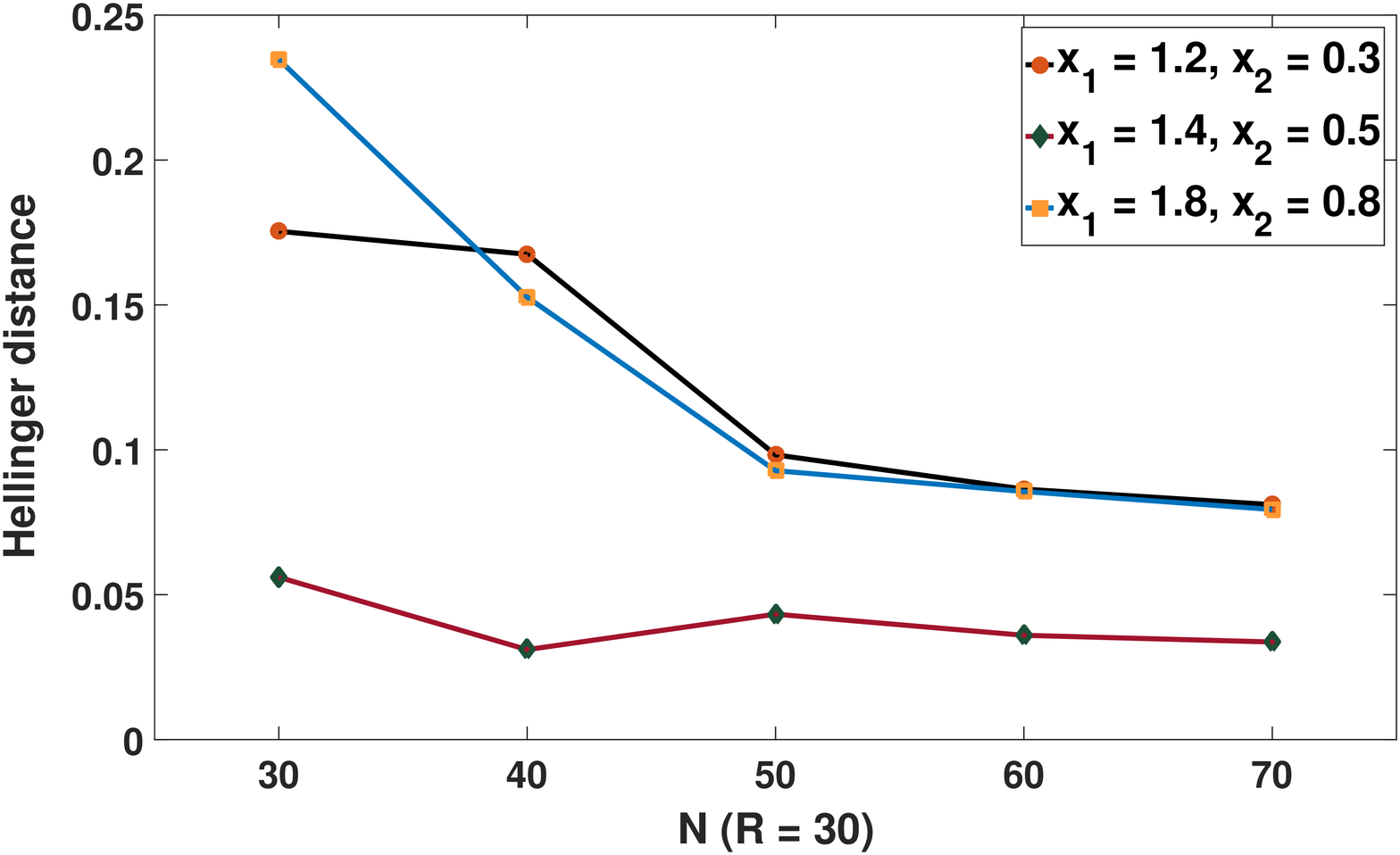}} 
    \subfigure[]{\label{subfig:lab8n3}\includegraphics[width=0.475\textwidth]{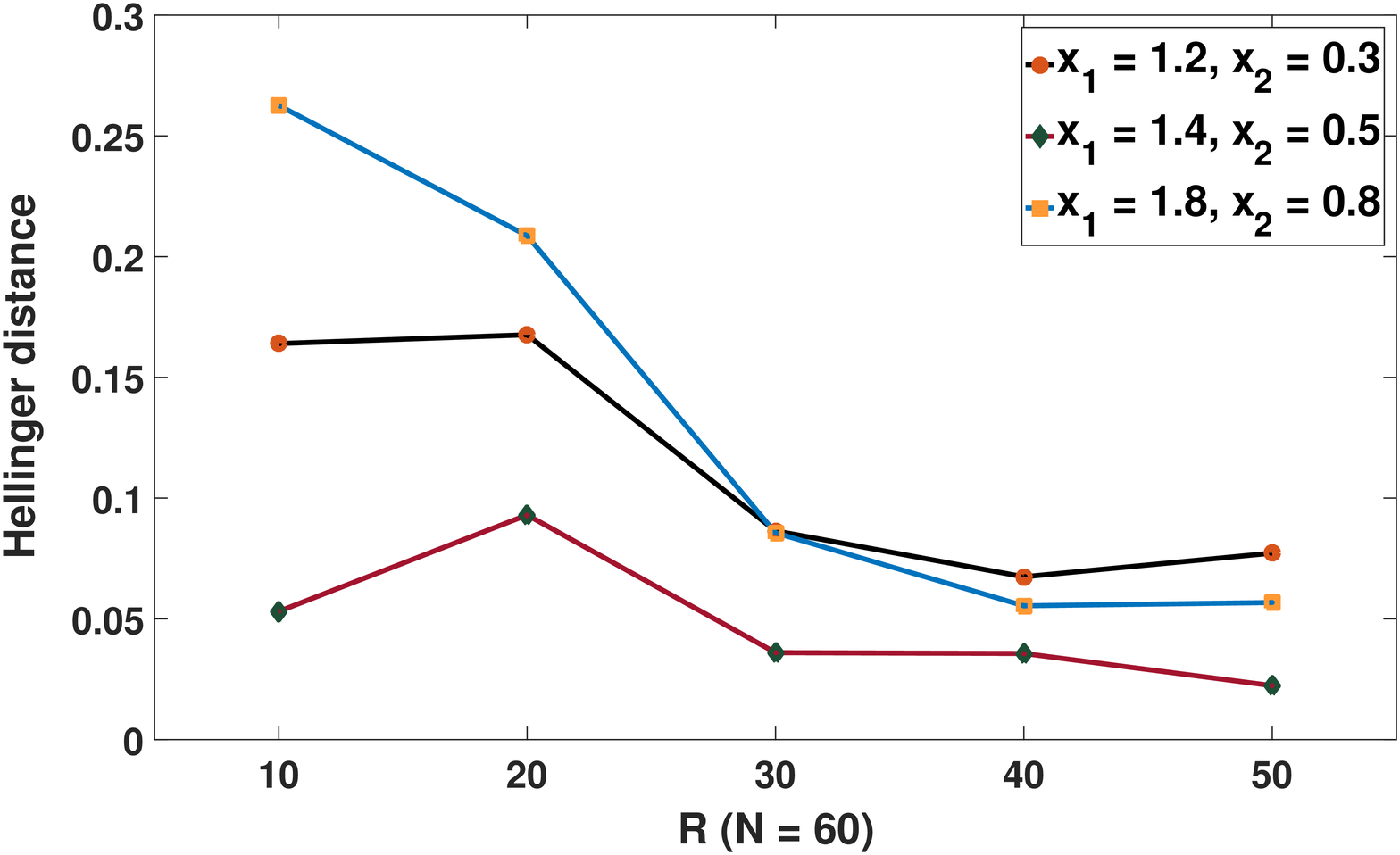}}
    \caption{Hellinger distance between predicted PDF and true response PDF for  $x_{1} = 1.2$ and $x_{2} = 0.3$, $x_{1} = 1.4$ and $x_{2} = 0.5$, and $x_{1} = 1.8$ and $x_{2} = 0.8$ with \textbf{(a)} variation in $N$ and $R$ for $N_{z} = 60$ \textbf{(b)} variation in $N$ for $R = 30$ and $N_{z} = 60$  \textbf{(c)} variation in $R$ for $N = 60$ and $N_{z} = 60$ .}
    \label{fig:8nn1}
\end{figure}

\begin{figure}[htbp!]
    \centering
    {\includegraphics[width=0.5\textwidth]{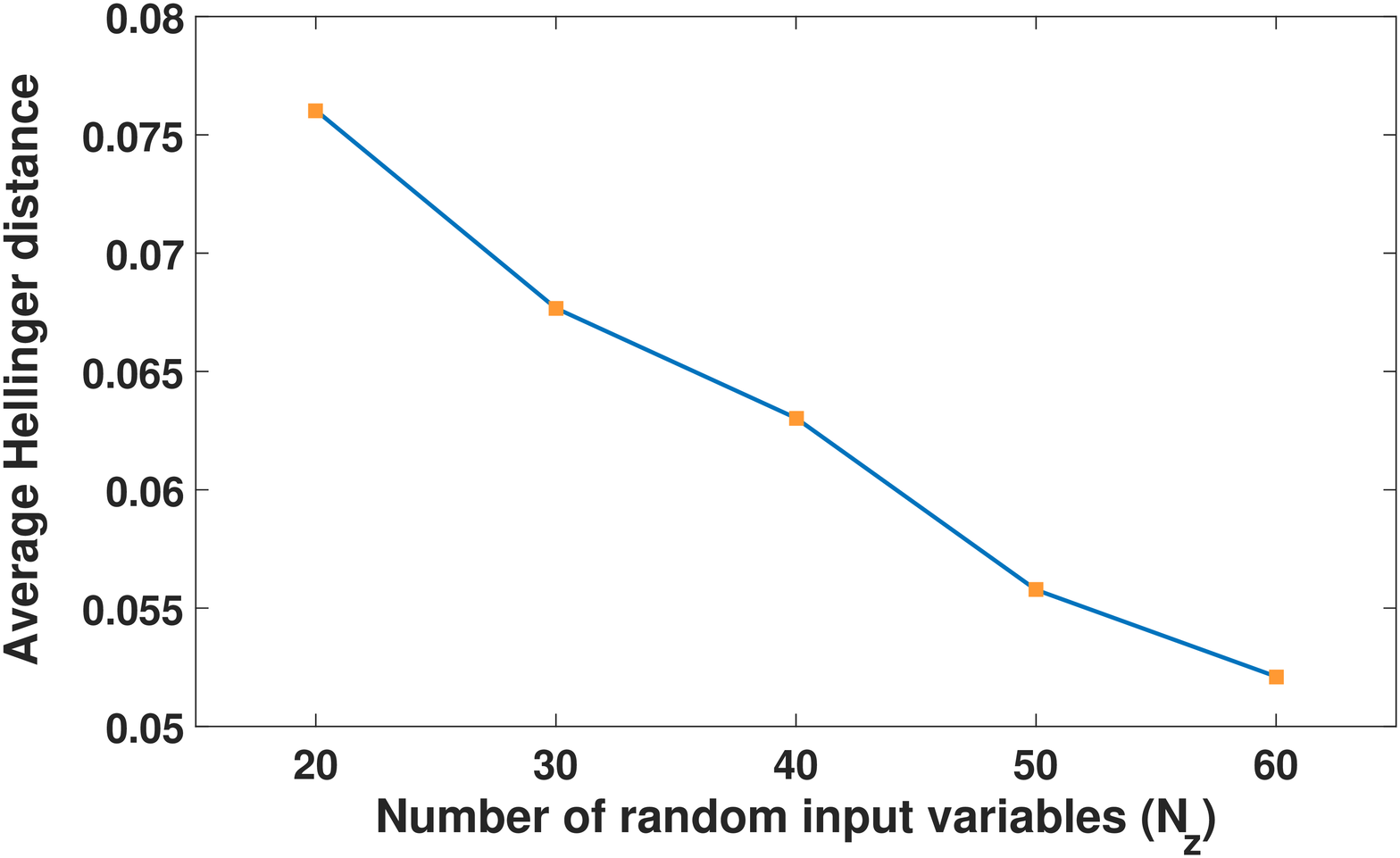}}
    \caption{Average of Hellinger distance between predicted PDF and true response PDF for  $x_{1} = 1.2$ and $x_{2} = 0.3$, $x_{1} = 1.4$ and $x_{2} = 0.5$, and $x_{1} = 1.8$ and $x_{2} = 0.8$ with variation in the number of standard normal random input variables, $N_{z}$, with $N = 60$ and $R = 50$.}
    \label{fig:8nn2}
\end{figure}

The PDF predictions by the model for $x_{1} = 1.2$ and $x_{2} = 0.3$, $x_{1} = 1.8$ and $x_{2} = 0.8$ are shown in Fig. \ref{fig:7} with $N = 60$, $R = 50$, and $N_{z} = 60$.
It can be observed that the PDF predictions from surrogate model produce an excellent match with the reference PDF, and the match is maintained whether the PDF at hand is close to a Normal distribution like \ref{subfig:lab1} or whether it is a positively skewed distribution like the one presented in \ref{subfig:lab2}.
$N$ and $R$ are varied for three different sets of input parameters to assess their influence on the performance of proposed approach for the current problem, and the results of the analysis are presented in Fig. \ref{fig:8nn1}. Observations from Fig. \ref{subfig:lab8n1} reveal that again, as the value of $N\times R$ is increased for a fixed $N_{z}$, the error metric follows a downward trend for all sets of input parameters . Furthermore, the results presented in Fig. \ref{subfig:lab8n2} show that as $N$ is increased for fixed $R$ and $N_{z}$, the error follows a downward trend for all cases. Similarly, it is observed, as shown in  Fig. \ref{subfig:lab8n3} that when $R$ is increased while keeping $N$ and $N_{z}$ fixed, the error decreases. Finally, for increase in $N_{z}$ with fixed $N$ and $R$, the average of the Hellinger distances and the results presented in Fig. \ref{fig:8nn2} reveal a decreasing trend in error.

\begin{figure}
    \centering
    \subfigure[]{\includegraphics[width=0.45\textwidth]{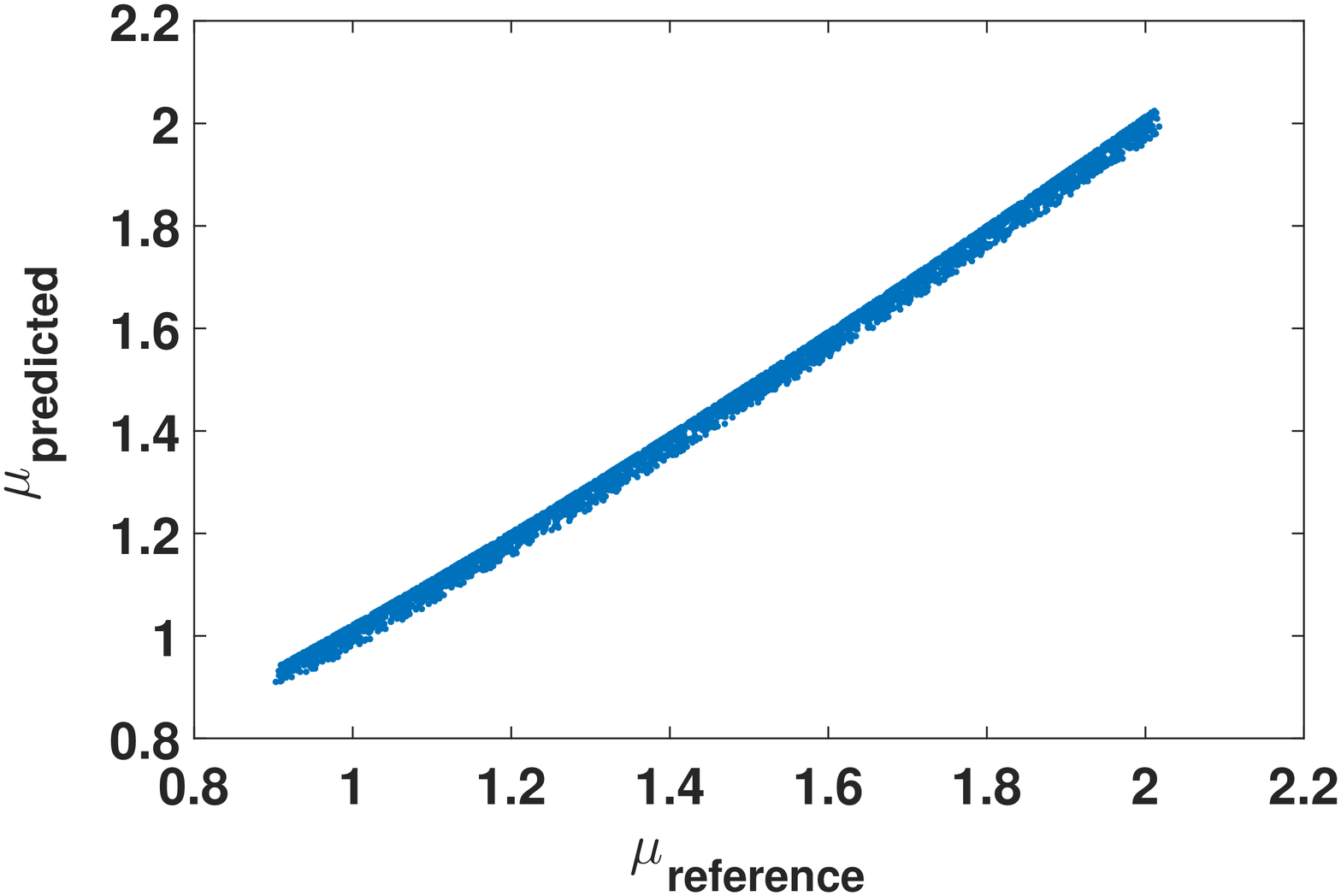}} 
    \subfigure[]{\includegraphics[width=0.45\textwidth]{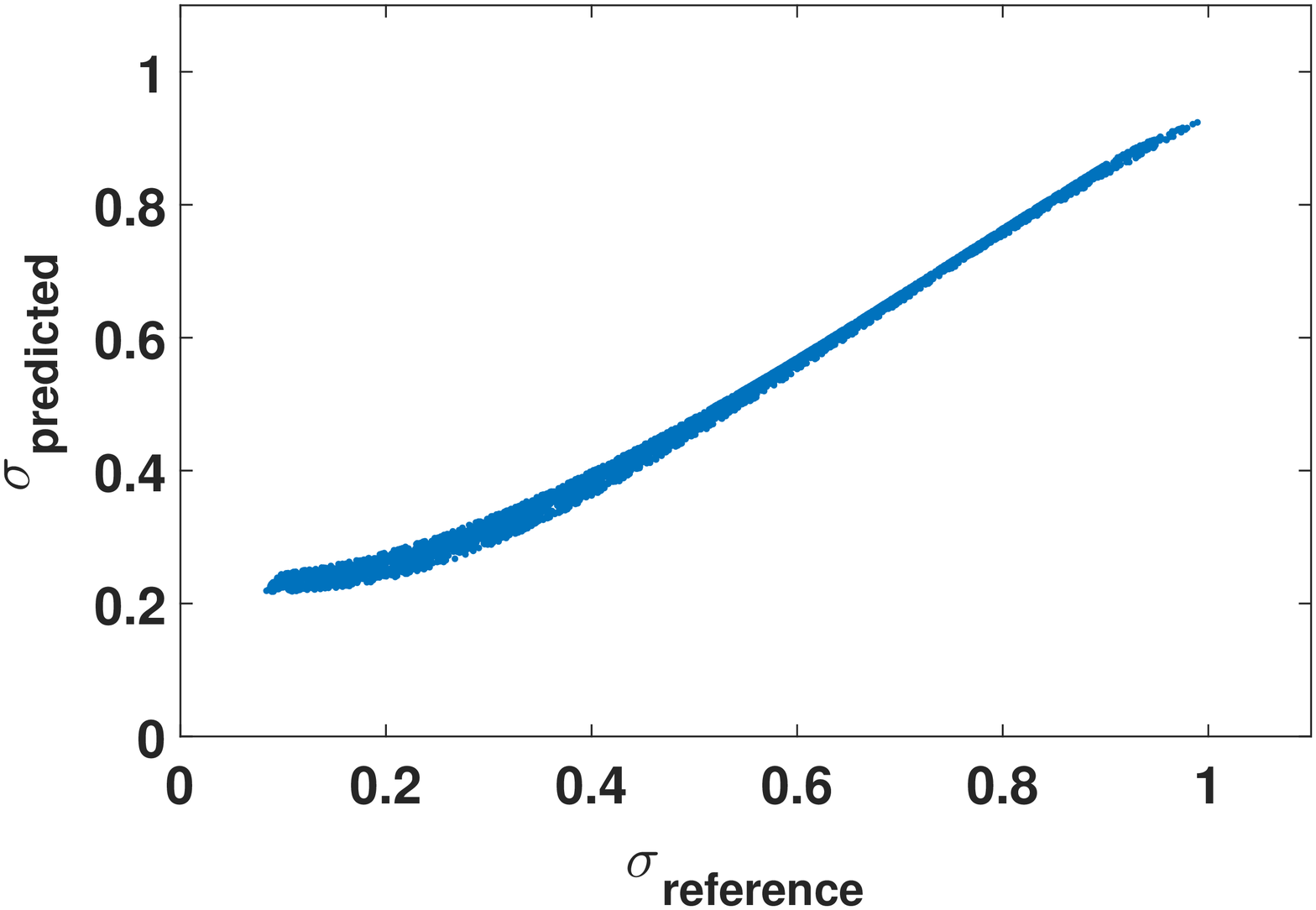}} 
    \subfigure[]{\includegraphics[width=0.45\textwidth]{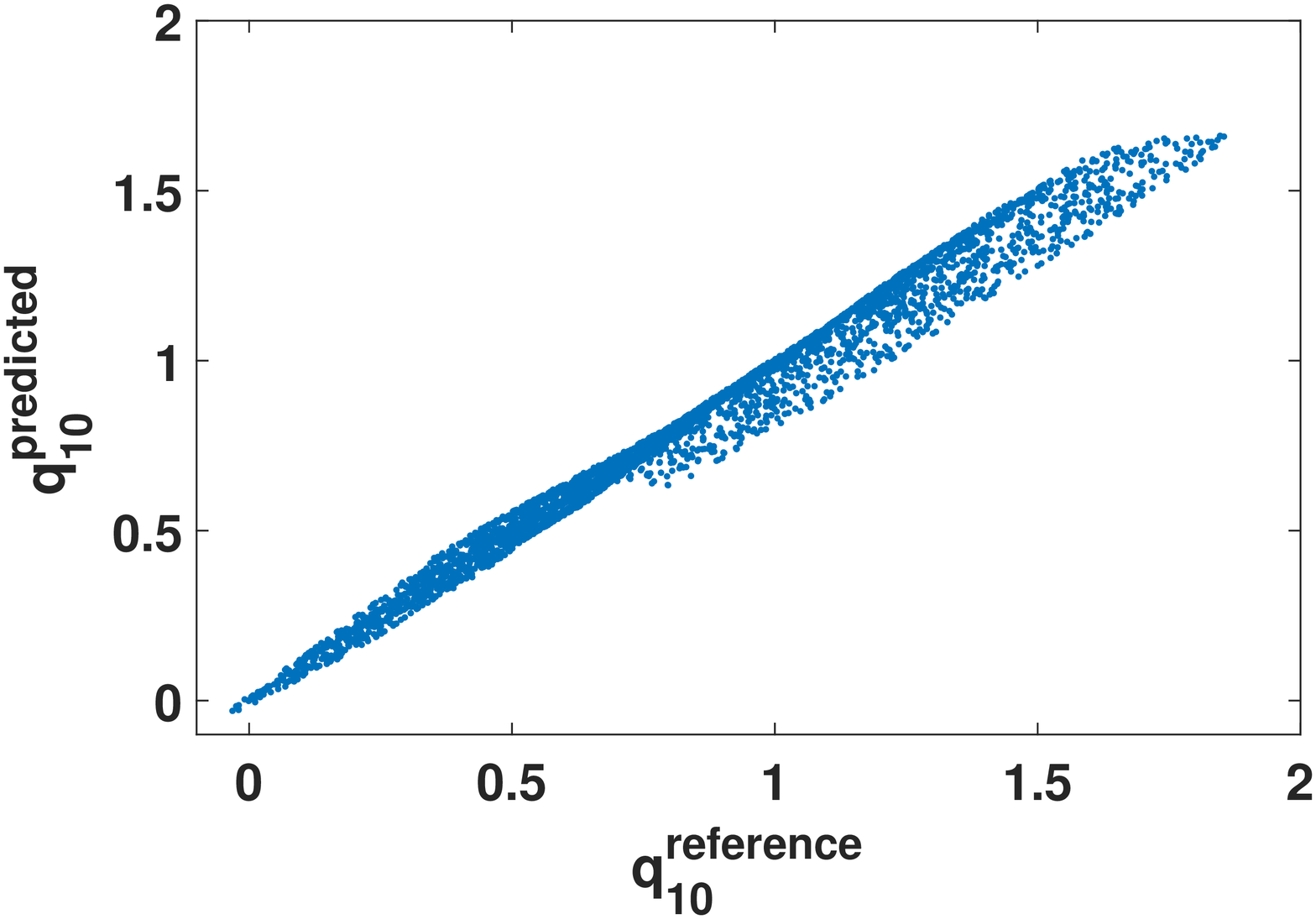}}
    \subfigure[]{\includegraphics[width=0.45\textwidth]{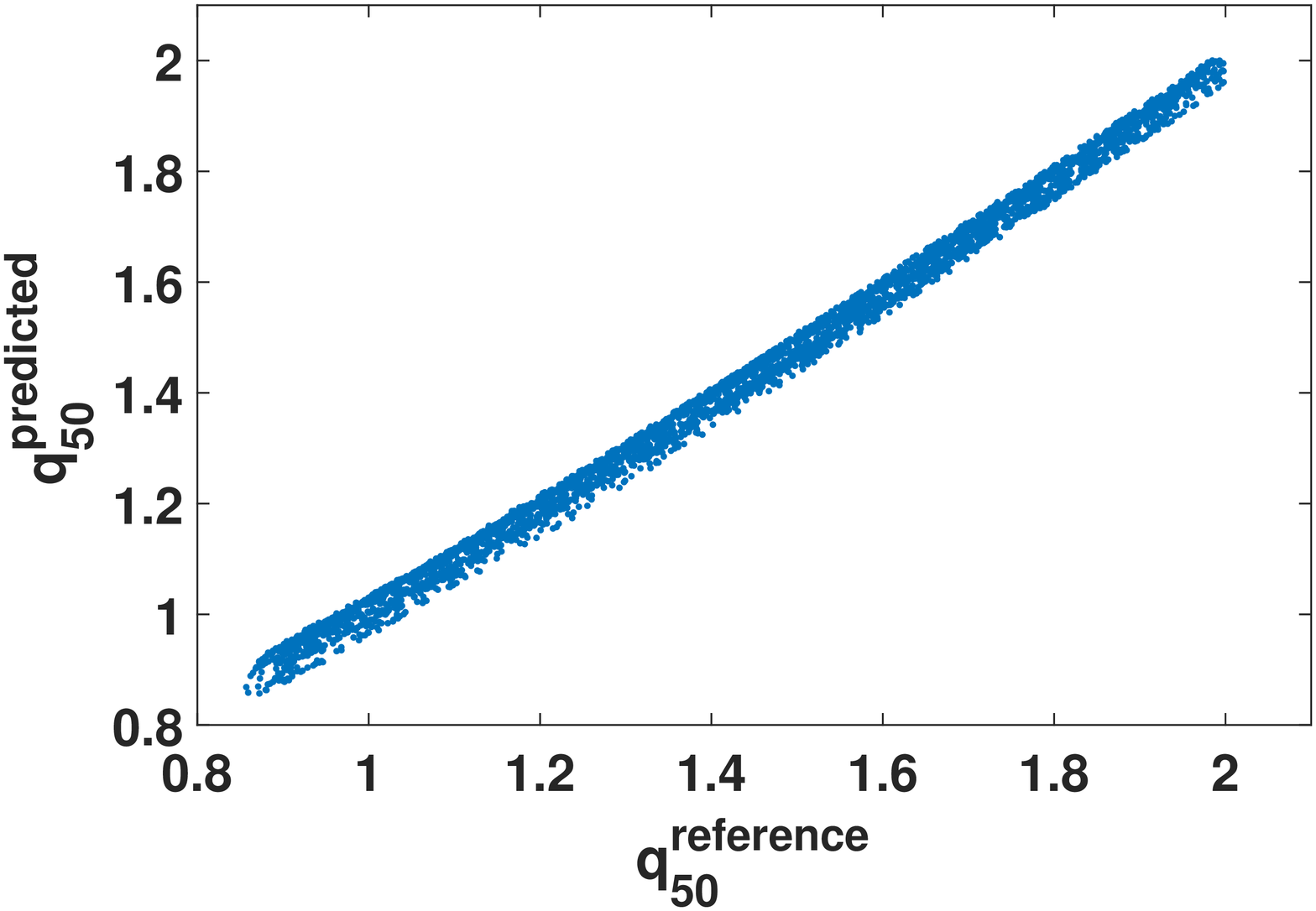}}
    \subfigure[]{\includegraphics[width=0.45\textwidth]{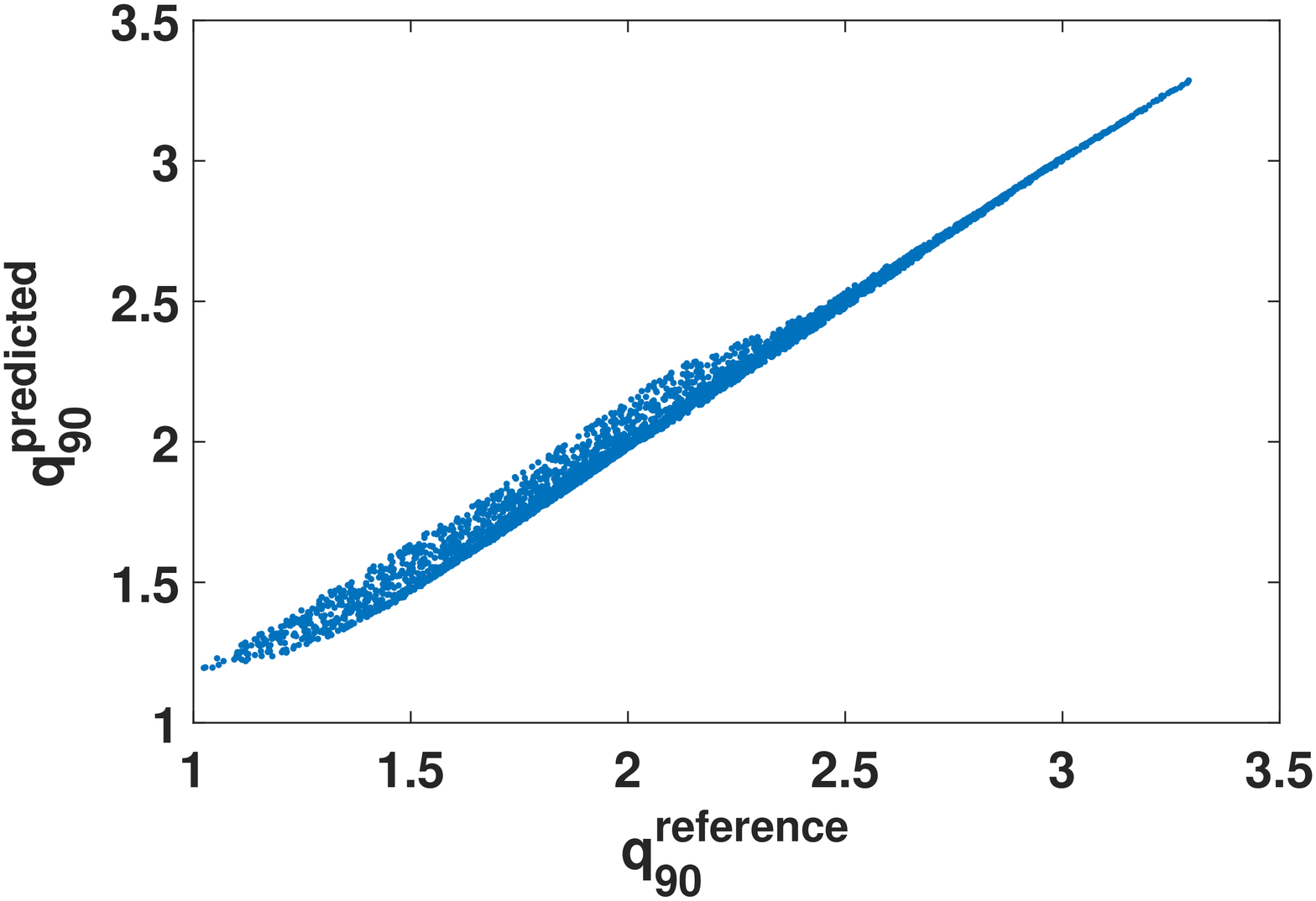}}
    \caption{ Comparison of predicted response distribution with true response distribution for 4000 values of input parameter $\vect{X}$ with  the help of \textbf{(a)} mean  \textbf{(b)} standard deviation \textbf{(c)} 10\% quantile   \textbf{(d)} 50\% quantile \textbf{(e)} 90\% quantile.}
    \label{fig:9}
\end{figure}

The accuracy of the model predictions is again evaluated by comparing the statistics of predicted response distribution with that of the true response distribution for a set of 4000 values of input parameter $\vect{X}$. The value of $N$, $R$, and $N_{z}$ are 60, 50, and 60, respectively. The results presented in Fig. \ref{fig:9} indicate excellent match between the results obtained using the proposed approach and the benchmark solutions. Finally, the Hellinger distance is computed between predicted and true response PDF for same values of input parameter $\vect{X}$, and the statistics are presented in Table \ref{tab:3}.
\begin{table}[htbp!]
    \centering
    \begin{tabular}{||c | c | c | c | c||} 
    \hline
    Mean & Standard deviation & 10\% quantile  & 50\% quantile  & 90\% quantile  \\ [0.5ex] 
    \hline\hline
    0.08927 & 0.07827 & 0.04606 & 0.06187 & 0.19155  \\ 
    \hline
    \end{tabular}
    \caption{Hellinger distance based error statistics for predicted and true response distributions at 4000 values of input parameter $\vect{X}$.}
    \label{tab:3}
\end{table}

\subsection{Example 4: Stochastic SIR model}
As evident from the current situation with SARS CoV-2 and with numerous infectious disease outbreaks in the past, infectious disease have been the cause of major mortality and illness time and again. The deterministic Susceptible-Infected-Recovered (SIR) epidemiological model introduced by \citet{kermack1927contribution} has proven to be a great success in epidemiological modeling. However, the applicability of deterministic disease transmission models is limited to the situations where the population is large. Also, it is simultaneously true that with most of the infectious disease transmission cases, there is stochasticity involved due to demographic factors, such as birth, death, recovery, or transmission, and environmental factors such as temperature, humidity, rainfall, or other conditions associated with ecological settings \cite{allen2017primer}, and therefore, it is seldom the case that deterministic laws are able to perfectly model these phenomena. Stochastic modeling of epidemics, on the other hand, allows to model the disease transmission even when the population number is small and also allows inclusion of stochasticity due to the factors discussed above. So, it is essential to study the stochastic SIR model as it allows for better simulation of the spread of the infection and in arriving at better strategies of intervention to control the outbreak. In our final example, we test the developed deep learning based framework  by using to create a surrogate for the stochastic SIR model.

In the deterministic SIR model for an epidemic in a large population, the entire population $N_{t}$ at any given time $t$ is essentially divided into three groups - Susceptible ($S_{t}$), Infected ($I_{t}$), and Recovered ($R_{t}$). A new infection can happen if an infected person comes into contact with a susceptible person or an infected person can recover as well. However, a recovered person cannot become infected again, as it is assumed that they become immune. The transmission and recovery rate are denoted with $\beta$ and $\gamma$, respectively. Also, the size of the population is assumed to be fixed with no new births or deaths, i.e. $N_{t}$ is equal to some constant $N$. Therefore, $S_{t}$, $R_{t}$, and $I_{t}$ are constrained by the following equation
\begin{equation}
S_{t} + I_{t} + R_{t} = N.
\label{eqn5}
\end{equation}
To model the evolution of the disease spread through time only $S_{t}$ and $I_{t}$ are required.
However, the stochastic SIR model, as mentioned above, helps in modeling the disease spread better by allowing for environmental and demo-graphical variability, and making random interactions among the population possible. This makes stochastic SIR model the primary choice for epidemiological modeling. The evolution of pair, $(S_{t},I_{t})$, is modeled as a Continuous Time Markov Chain (CTMC). For additional details and assumptions, the reader is referred to \cite{allen2017primer, allen2015stochastic}. Furthermore, the time $t$ at which $I_{t} = 0$, the epidemic stops, i.e., no further spread of infection, and we denote this time as $t_{a}$. Also, the simulation of CTMC SIR model is done with Gillespie algorithm \cite{Gillespie1977}.

We use the problem definition given by \citet{zhu2021emulation} for the current case study. The size of population, $N$, is taken to be equal to $2000$. The value of $\beta$ and $\gamma$ is set to $0.5$. The vector $\vect{X} = (S_{0},I_{0})$,  where $S_{0}$ is size of initial susceptible population and $I_{0}$ is the initial size of population infected, represents the initial configuration and also, serves as the uncertain input parameters leading to the inclusion of different scenarios to the CTMC model. In this example, $S_{0}\sim\mathcal{U}(1200,1800)$ and $I_{0}\sim\mathcal{U}(20,200)$ are considered. The number of individuals who were infected during the completed time span of an outbreak, $Y = S_{t_{a}} - S_{0}$, is considered as the QoI for this case study.

The vector $(\vect{X},\vect{S})^{T}$ serves as the input to the neural network, where $\vect{S}$ is a vector containing $M$ standard normal random variables, i.e., $(s_{1},s_{2},...,s_{M})$. Also, Leaky ReLu is used as the activation function for the output layer, the batch size is set to 300, and the number of epochs required for successful training is 226. \\
\begin{figure}[htbp!]
    \centering
    \subfigure[]{\label{subfig:lab3}
                  \includegraphics[width=0.48\textwidth]{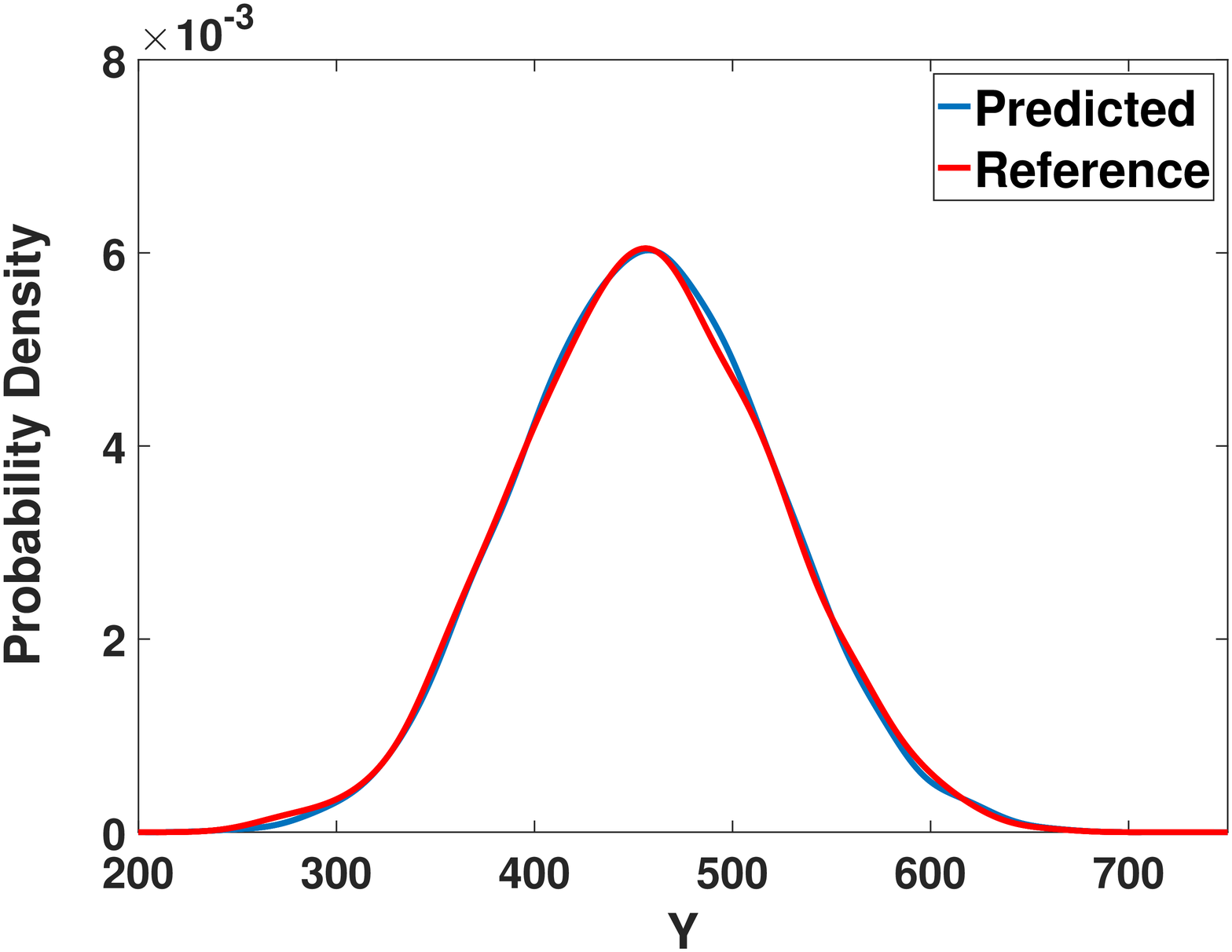}} 
    \subfigure[]{\label{subfig:lab4}
                 \includegraphics[width=0.48\textwidth]{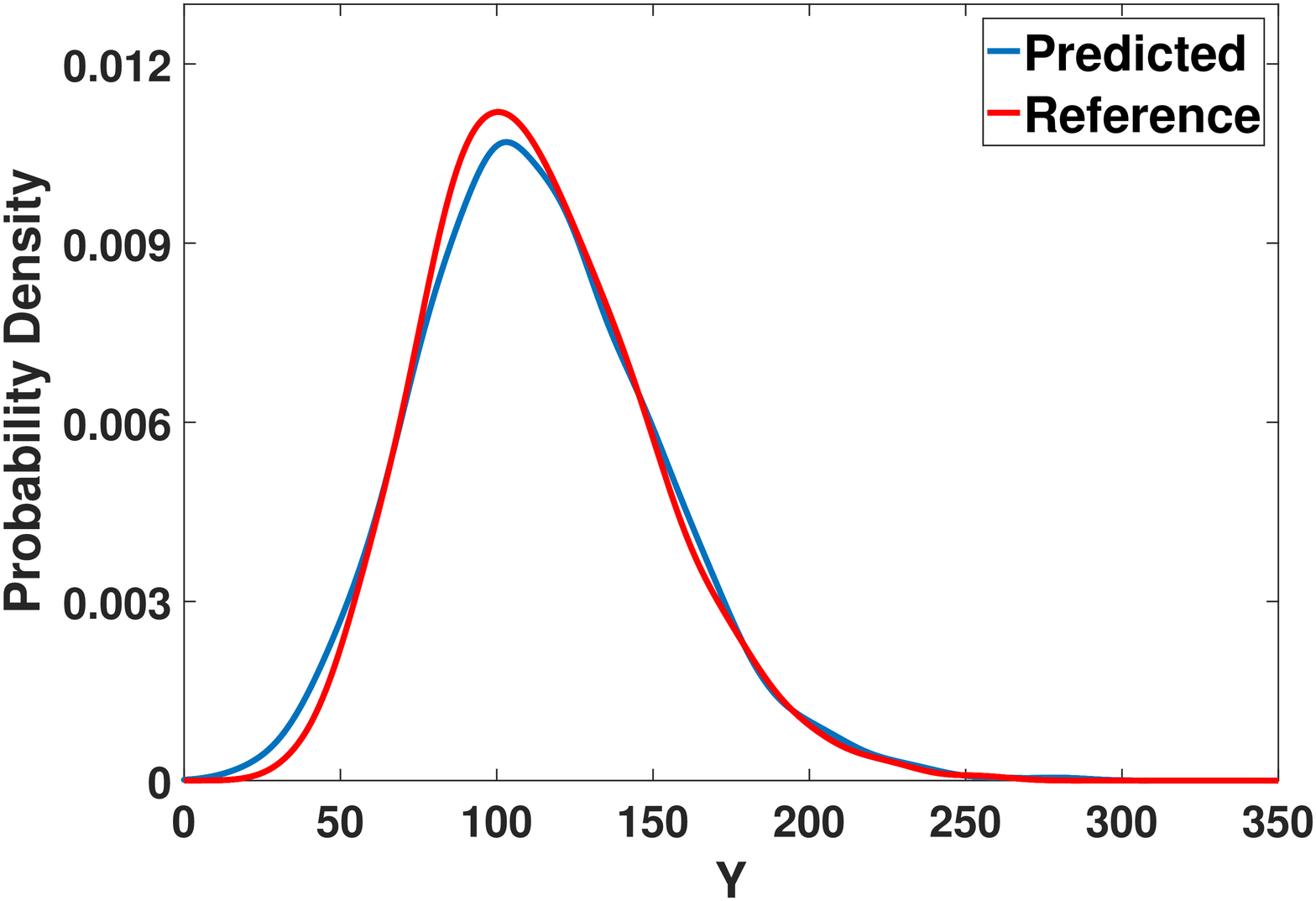}} 
    \caption{ \textbf{(a)} PDF prediction for $S_{0} = 1714$ and $I_{0} = 165$   \textbf{(b)} PDF prediction for $S_{0} = 1364$ and $I_{0} = 61$.}
    \label{fig:10}
\end{figure}

\begin{figure}[htbp!]
    \centering
    \subfigure[]{\label{subfig:lab5}
    \includegraphics[width=0.475\textwidth]{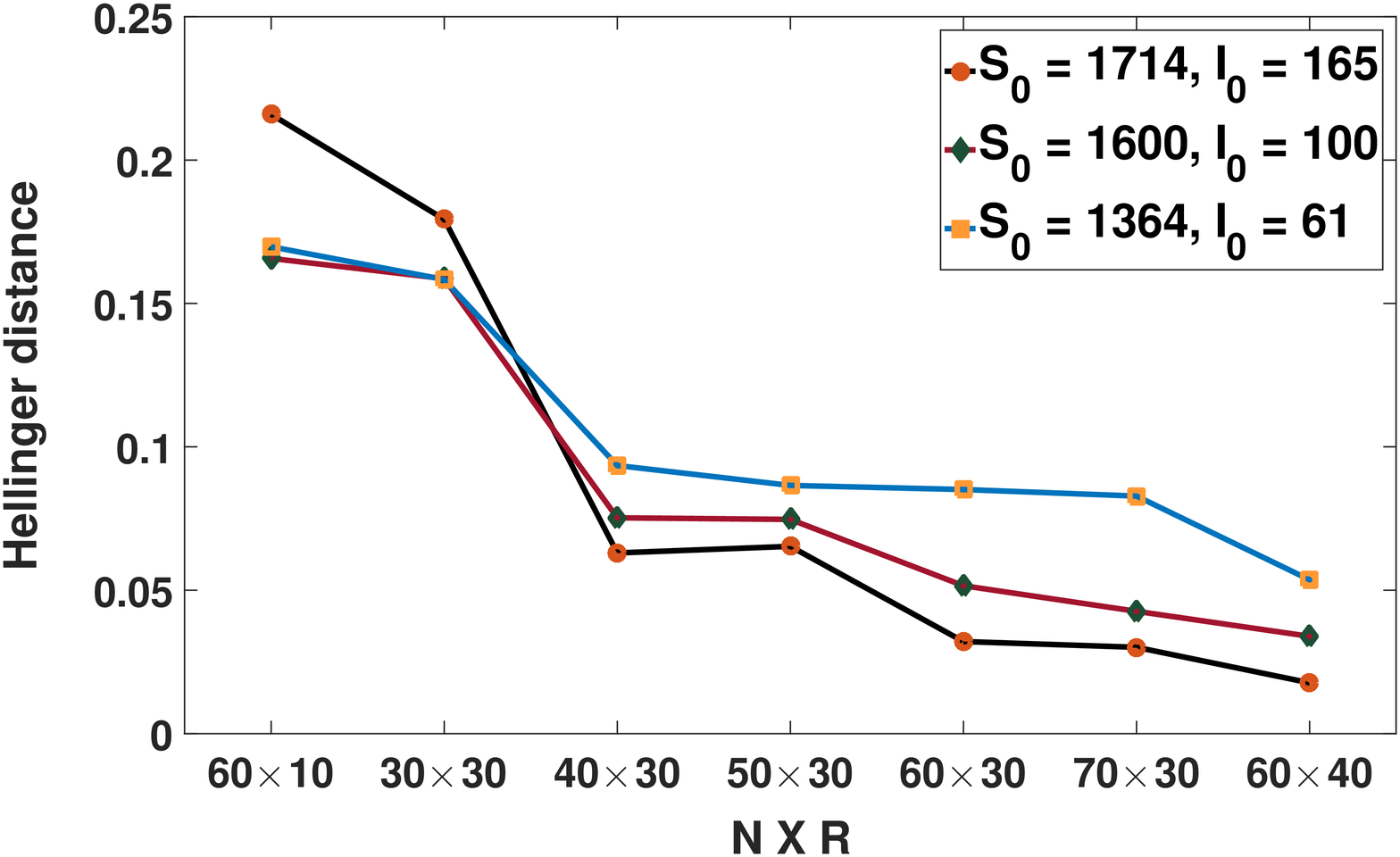}} 
    \subfigure[]{\label{subfig:lab6}
    \includegraphics[width=0.475\textwidth]{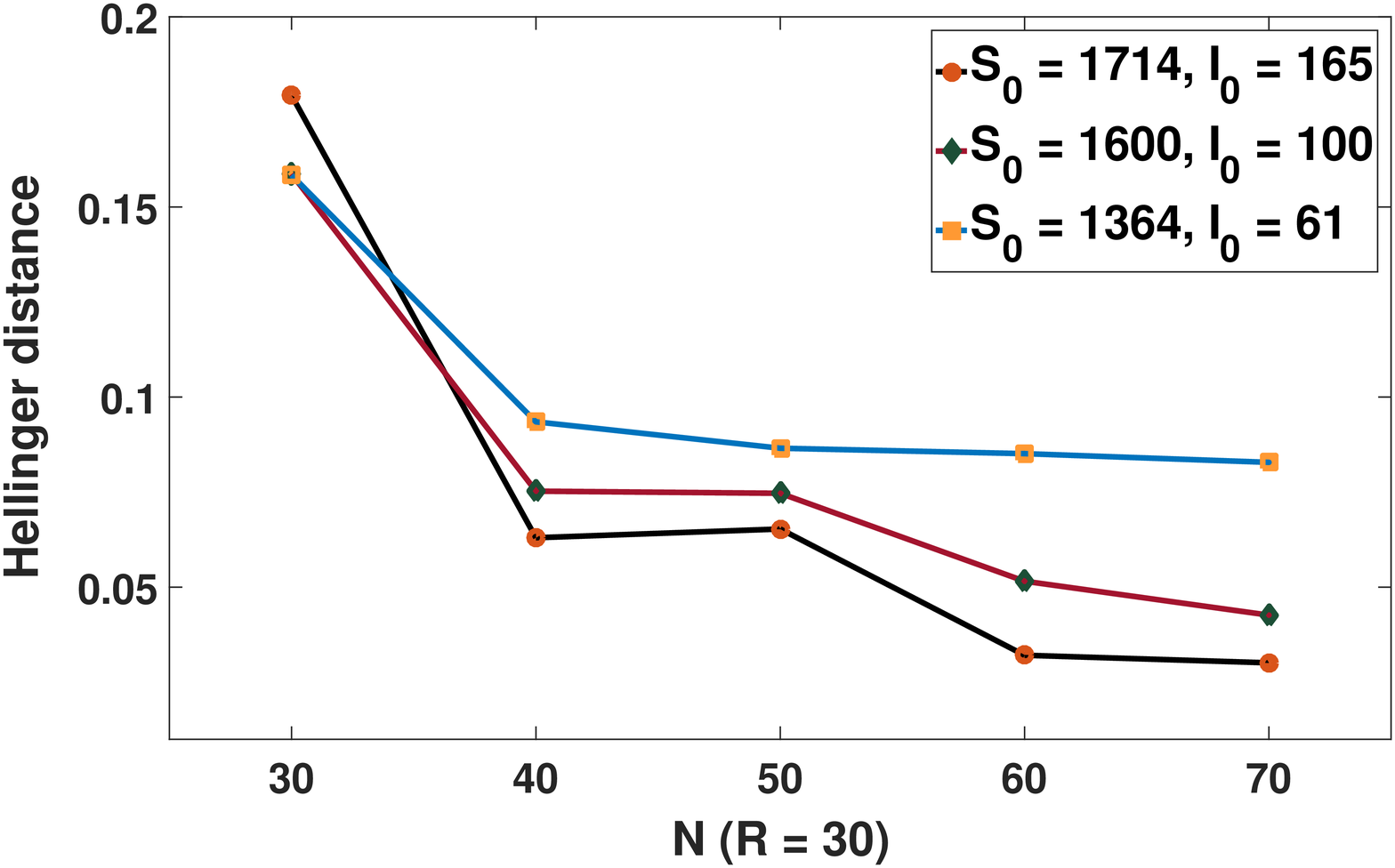}} 
    \subfigure[]{\label{subfig:lab7}
    \includegraphics[width=0.475\textwidth]{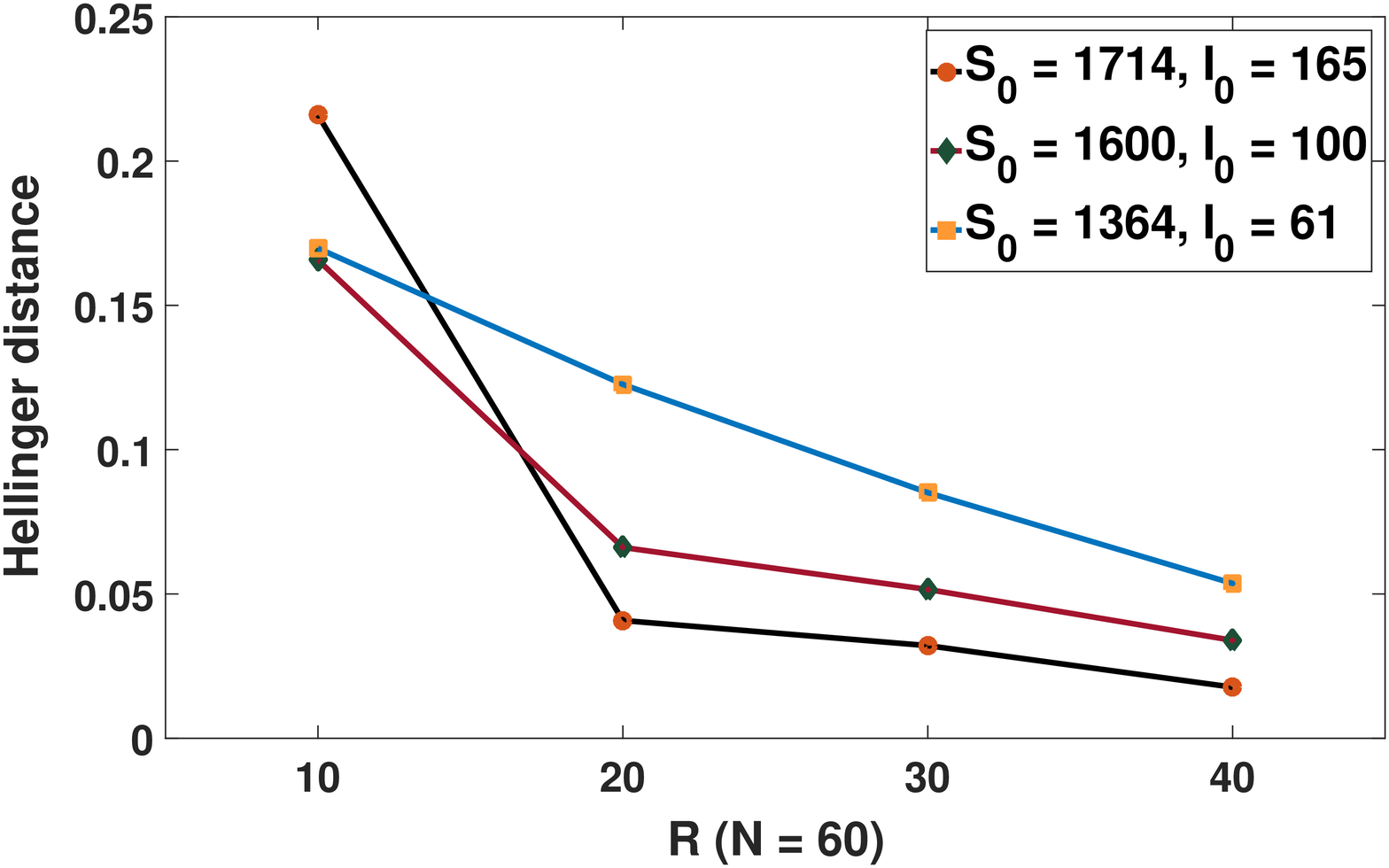}}
    \caption{Hellinger distance between predicted PDF and true response PDF for  $S_{0} = 1714$ and $I_{0} = 165$ , $S_{0} = 1600$ and $I_{0} = 100$, and $S_{0} = 1364$ and $I_{0} = 61$ with \textbf{(a)} variation in $N$ for $R$ and $N_{z} = 60$ \textbf{(b)} variation in $N$ for $R = 30$ and $N_{z} = 50$  \textbf{(c)} variation in $R$ for $N = 60$ and $N_{z} = 50$.}
    \label{fig:11n1}
\end{figure}

\begin{figure}[htbp!]
    \centering
    \subfigure[]{\label{subfig:lab8}
    \includegraphics[width=0.55\textwidth]{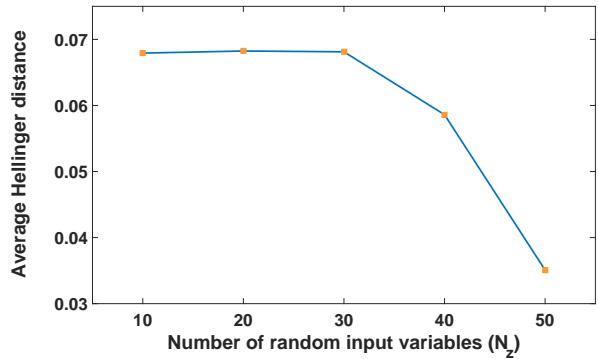}}
    \caption{Average of Hellinger distance between predicted PDF and true response PDF for  $S_{0} = 1714$ and $I_{0} = 165$ , $S_{0} = 1600$ and $I_{0} = 100$, and $S_{0} = 1364$ and $I_{0} = 61$ with variation in the number of standard normal random input variables, $N_{z}$, for $N = 60$ and $R = 40$.}
    \label{fig:11n2}
\end{figure}

The PDF predictions by the model for $S_{0} = 1714$ and $I_{0} = 165$, $S_{0} = 1364$ and $I_{0} = 61$ are shown in Fig. \ref{fig:10} with $N = 60$, $R = 40$, and $N_{z} = 50$. The shape of PDF varies with input parameter, in a manner very similar to Section \ref{ss:3}, from Normal to positively skewed. We can see from Fig. \ref{subfig:lab3} that, for the symmetric PDF, the surrogate model's predicted PDF is an excellent match for the reference PDF. It can also be seen from Fig. \ref{subfig:lab4} that, for a slightly right skewed PDF, an overall excellent match with reference PDF is achieved, with decent fit in the tail sections and most parts of the PDF; however, simultaneously, a very slight mismatch near the peak can also be observed. Again, the convergence behaviour is shown by computing the Hellinger distance between the predicted and the reference PDF for three experimental points taken, respectively, from the two extremes and the middle of the input parameter range. Then, the variation of the Hellinger distance is analysed with change in $N$ and $R$ while $N_{z}$ being fixed  and the results are presented in Fig. \ref{fig:11n1}. It is observed that the error for all three cases decreases with increase in parameter under variation or more generally, the size of training data. Similarly, in \ref{fig:11n2}, for the variation of mean Hellinger distance with $N_{z}$, it is observed that the average error decreases with increase in $N_{z}$.

\begin{figure}[htbp!]
    \centering
    \subfigure[]{\includegraphics[width=0.45\textwidth]{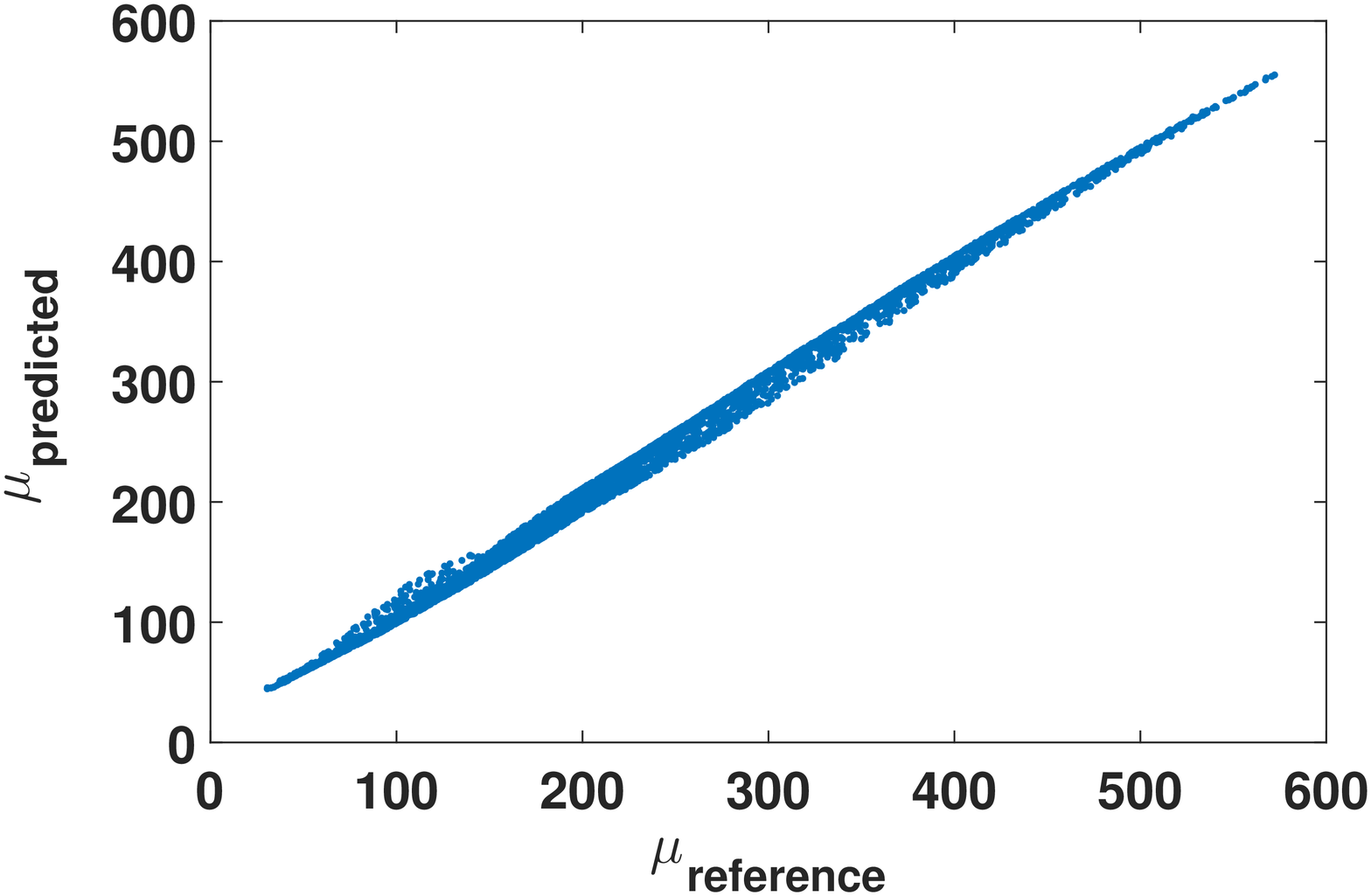}} 
    \subfigure[]{\includegraphics[width=0.45\textwidth]{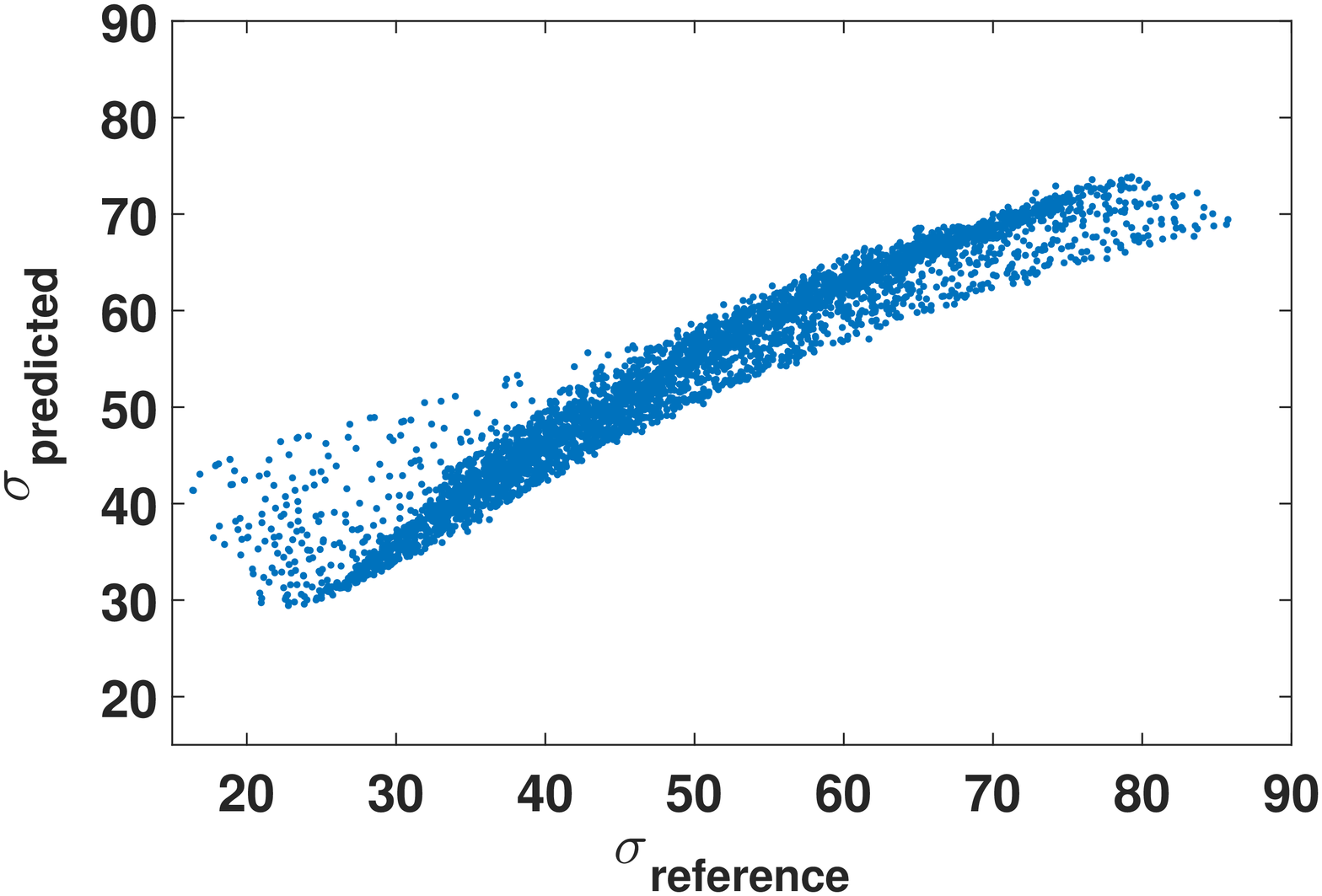}} 
    \subfigure[]{\includegraphics[width=0.45\textwidth]{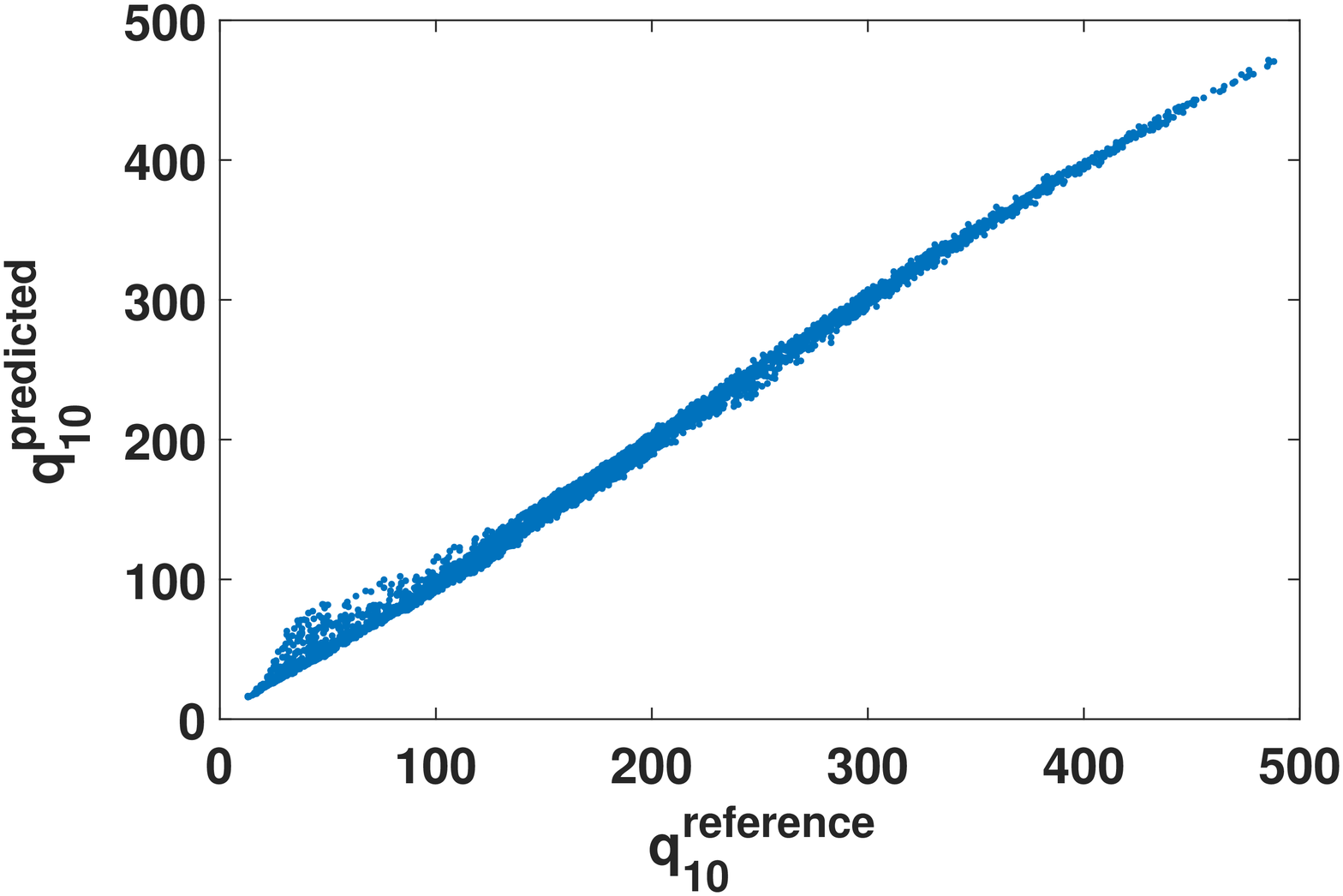}}
    \subfigure[]{\includegraphics[width=0.45\textwidth]{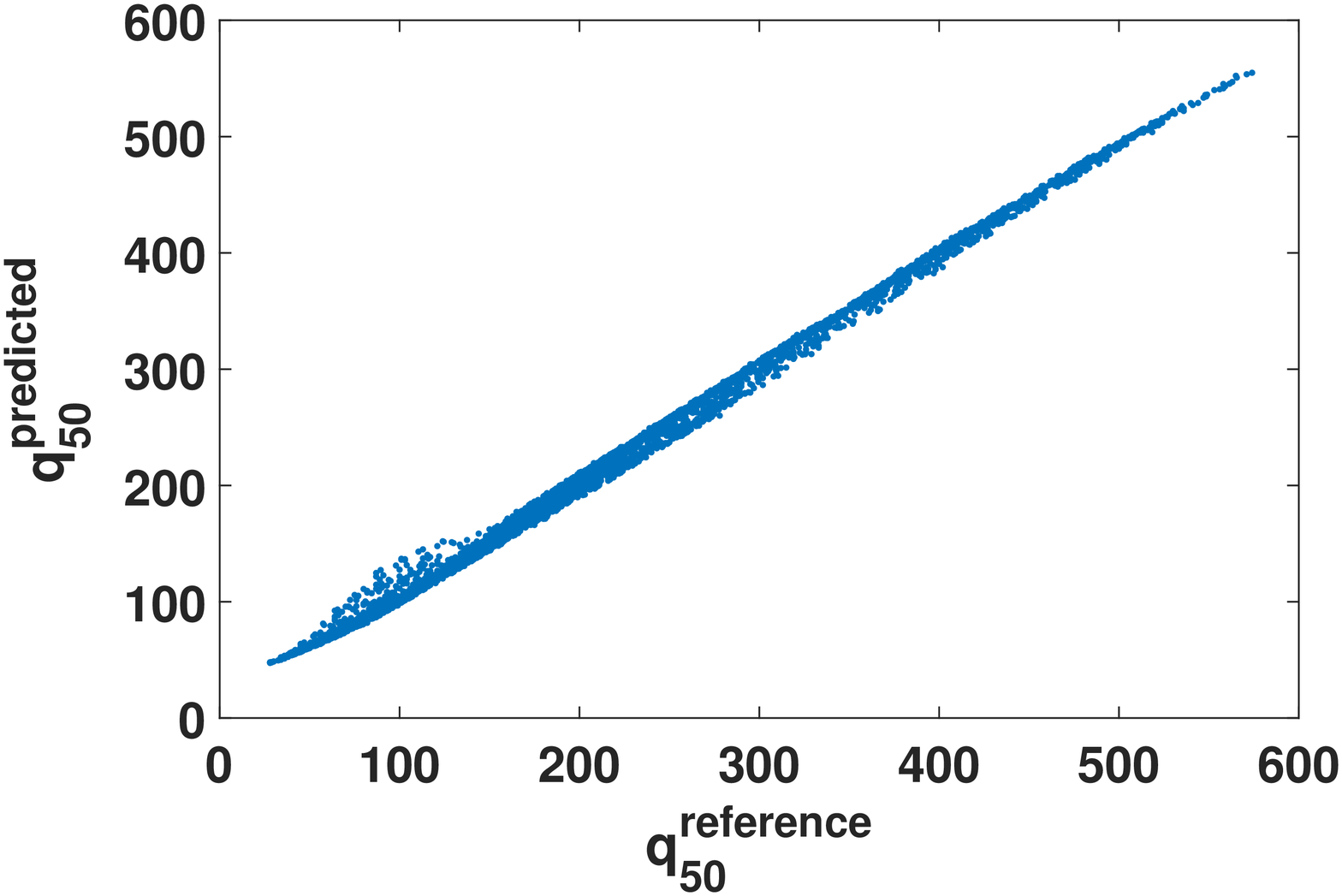}}
    \subfigure[]{\includegraphics[width=0.45\textwidth]{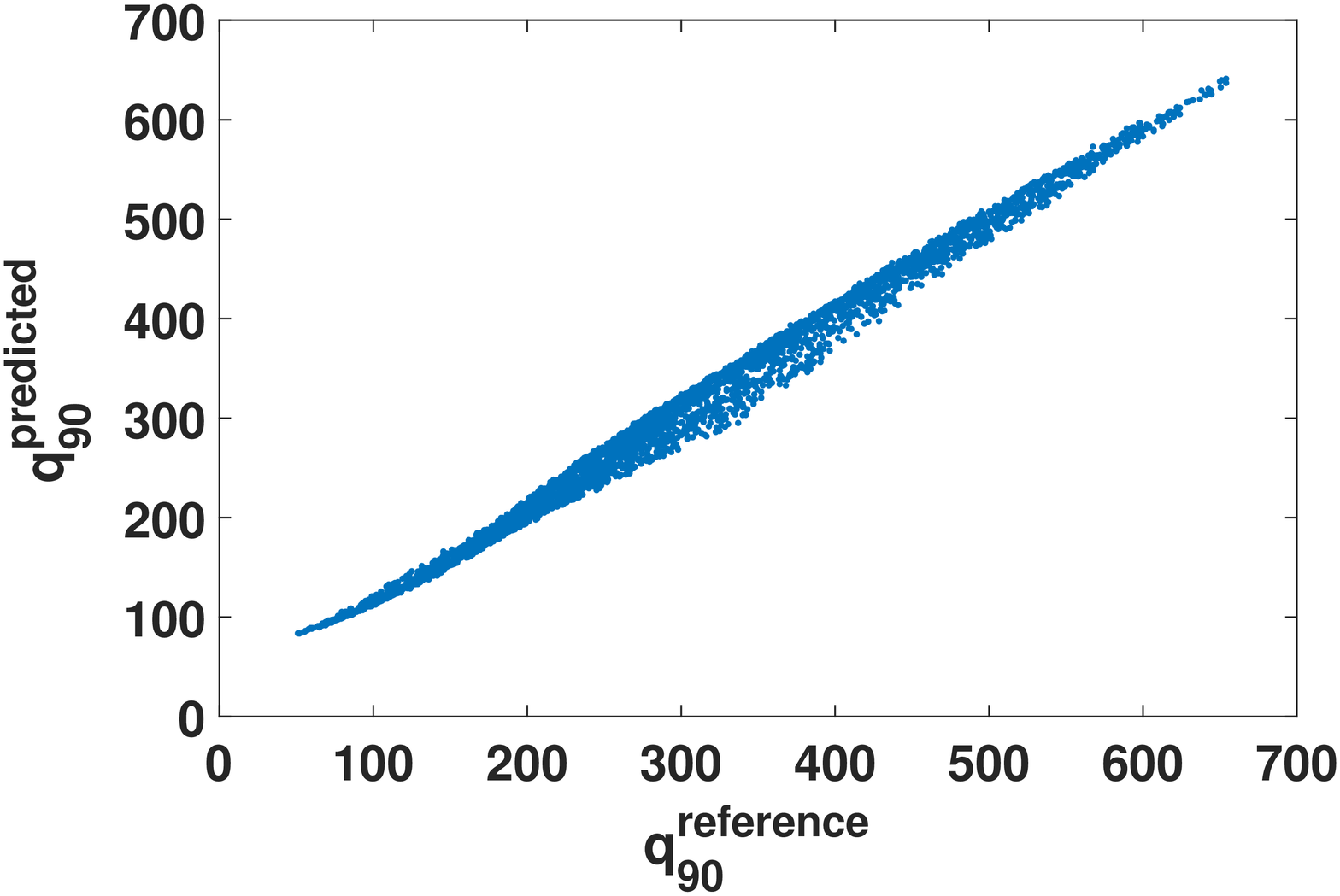}}
    \caption{ Comparison of predicted response distribution with true response distribution for 4000 values of input parameter $\vect{X}$ with  the help of \textbf{(a)} mean  \textbf{(b)} standard deviation \textbf{(c)} 10\% quantile   \textbf{(d)} 50\% quantile \textbf{(e)} 90\% quantile.}
    \label{fig:12}
\end{figure}

We evaluate the accuracy of the model predictions by comparing the statistics of predicted response distribution with that of the true response distribution for a set of 4000 values of input vector $\vect{X}$. The value of $N$, $R$, and $N_{z}$ are 60, 40, and 50, respectively. The results are presented in Fig. \ref{fig:12}. Good match between the surrogate predicted results and benchmark results is observed. For quantitative assessment, the Hellinger distance between predicted and true response PDF are presented in Table \ref{tab:4}.
\begin{table}[htbp!]
    \centering
    \begin{tabular}{||c | c | c | c | c||} 
    \hline
    Mean & Standard deviation & 10\% quantile  & 50\% quantile  & 90\% quantile  \\ [0.5ex] 
    \hline\hline
    0.08275 & 0.03741 & 0.046070 & 0.07555 & 0.12169  \\ 
    \hline
    \end{tabular}
    \caption{Hellinger distance based error statistics for predicted and true response distributions for stochastic SIR model at 4000 values of input parameter $\vect{X}$.}
    \label{tab:4}
\end{table}

\section{Conclusions}
\label{S:6}
In this paper, we proposed a CGMMN-based generative deep learning framework for construction of surrogate models for stochastic simulators. The proposed approach used an feed-forward deep neural network which estimated the response distribution by minimizing the conditional maximum mean discrepancy (CMMD) objective. The proposed approach is mathematically rigorous in the sense that it makes no assumption about the shape of the response distribution. Additionally, the kernel parameters present in the CMMD loss function are considered as unknowns and estimated along with the neural network parameters by using automatic differentiation and stochastic gradient descent. 
The most noteworthy point about the emulation method proposed in this study was its ability to match all orders of moments of a response distribution.

The performance of the proposed method was demonstrated on two analytical examples and two case studies --- the stochastic differential equation without a closed form solution and the stochastic SIR model. The results show that the propounded method is able to produce quite accurate approximations of the response distributions and the number of replications required were also less when compared to previous methods which did not impose a prior hypothesis on shape of the response. Another salient feature of the current framework is that it is not strictly dependent on replications as the values of error for smaller and larger number of replications are not very dissimilar.

Despite the success of the proposed approach in development of surrogate model for stochastic simulator, it is worthwhile to note that the proposed approach is not limited to feed-forward neural network; instead, one can easily accommodate more advanced neural network architectures including convolutional and recurrent neural networks. This will be useful in dealing with systems having large number of stochastic dimensions. To summarize, the proposed approach is capable of developing efficient surrogate models from very few replications and we envision the proposed approach to be of great use in reliability analysis and sensitivity analysis of systems governed by stochastic simulators. 

\section*{Acknowledgement}
SC acknowledges the financial support received from IIT Delhi through seed grant.

\end{document}